\documentclass[twoside,11pt]{article}

\usepackage{blindtext}
\usepackage[abbrvbib, preprint]{jmlr2e}

\newcommand{\indp}{\perp\!\!\!\!\perp} 
\newcommand{\calG}{\mathcal{G}}
\newcommand{\nbr}{\text{ne}}
\newcommand{\nc}{\text{nc}}
\newcommand{\ch}{\text{ch}}
\newcommand{\pa}{\text{pa}}
\newcommand{\PA}{\text{PA}}

\usepackage{lastpage}

\ShortHeadings{Causal Discovery by Sequential Edge Orientation}{Huang and Zhou}
\firstpageno{1}

\begin{document}

\title{Nonlinear Causal Discovery through a Sequential Edge Orientation Approach}

\author{\name Stella Huang \email stellahyh@ucla.edu \\
       \addr Department of Statistics and Data Science\\
       University of California, Los Angeles\\
       Los Angeles, CA 90095, USA
       \AND
       \name Qing Zhou \email zhou@stat.ucla.edu \\
       \addr Department of Statistics and Data Science\\
       University of California, Los Angeles\\
       Los Angeles, CA 90095, USA}

\editor{}

\maketitle

\begin{abstract}
Recent advances have established the identifiability of a directed acyclic graph (DAG) under additive noise models (ANMs), spurring the development of various causal discovery methods. However, most existing methods make restrictive model assumptions, rely heavily on general independence tests, or require substantial computation. To address these limitations, we propose a sequential procedure to orient undirected edges in a completed partial DAG (CPDAG), representing an equivalence class of DAGs, by leveraging a pairwise additive noise model (PANM) to identify their causal directions. We prove that this procedure can recover the true causal DAG assuming a restricted ANM. Building on this result, we develop a novel constraint-based algorithm for learning causal DAGs under nonlinear ANMs. 
Given an estimated CPDAG, we develop a ranking procedure that sorts undirected edges by their adherence to the PANM, which defines an evaluation order of the edges. To determine the edge direction, we devise a statistical test that compares the log-likelihood values, evaluated with respect to the competing directions, of a sub-graph comprising just the candidate nodes and their identified parents in the partial DAG. We further establish the structural learning consistency of our algorithm in the large-sample limit. Extensive experiments on synthetic and real-world data sets demonstrate that our method is computationally efficient, robust to model misspecification, and consistently outperforms many existing nonlinear DAG learning methods.
\end{abstract}


\section{Introduction}

Structural causal models \citep{pearl@2000} represent the causal relations amongst a set of variables using a directed acyclic graph (DAG), while the underlying data generating process is described by a set of structural equation models (SEMs). In practice, the true DAG is often unknown or difficult to construct due to limited domain knowledge. Consequently, a wide range of causal discovery methods have been developed to learn the underlying DAG or its equivalence class from observational data \citep{glymour2019review,vowels2022d}. 

In a general SEM, a variable is modeled as a deterministic function of other variables and an exogenous noise term. More precisely, for random variables $\{X_{i}\}_{i=1}^{p}$ with corresponding error terms $\{\varepsilon_{i}\}_{i=1}^{p}$, the general SEM takes the form
\begin{align}\label{eq:genSEM}
X_{i} = f_{i}(\PA_{i}, \varepsilon_{i}), \quad\quad i=1, \ldots, p, 
\end{align}
where $\PA_{i}$ denotes the parent set of $X_{i}$. A common assumption is that each $f_{i}(\cdot)$ is a linear SEM with additive Gaussian noise. Although analytically convenient, this assumption is not only overly simplistic, but also limits algorithms to learning a completed partially direct acyclic graph (CPDAG), which encodes a set of Markov equivalent DAGs sharing the same conditional independence relations, rather than the exact true DAG \citep{chickering2002learning}. Nevertheless, previous research has demonstrated conditions enabling the identifiability of the true DAG from observational data. Initial works by \citet{hoyer2008nonlinear} and \citet{shimizu2006linear} proved bivariate identifiability under nonlinear functions and/or non-Gaussian noises, respectively. These assumptions break the symmetry in the bivariate distribution of two nodes, enabling the identification of causal directions \citep{zhang2016nonlinear}.

\subsection{Relevant Work} 

In this paper, we focus on learning causal DAGs from nonlinear data. Prior works, such as the additive noise model (ANM) by \citet{hoyer2008nonlinear} and post-nonlinear model by \citet{zhang2009pnl_identifiability}, have established the identifiability of the true DAG under particular assumptions on the function class $f_{i}(\cdot)$ in the general SEM. \citet{peters2011IFMOC} proved identifiability for the general SEM~\eqref{eq:genSEM} by defining the concept of an identifiable functional model class. Specifically, the authors present a criterion on $\{X_{i}, f_{i}(\PA_{i}), \varepsilon_{i}\}$ for DAG identifiability, hence generalizing previous works focusing on specific models only. 
They also developed an algorithm to find all DAGs that satisfy their identifiability condition through iteratively testing for independence between residuals and parents.

New causal discovery algorithms have then ensued from these identifiability results. In the domain of constraint-based methods, kernel-based tests have garnered considerable attention. The kernel-based conditional independence (KCI) test, a notable early example proposed by \citet{kci_test}, captures nonlinear relationships by mapping random variables to reproducing kernel Hilbert spaces using kernel methods. Building on this, the algorithm RESIT \citep{resit_anm} utilizes a kernel-based statistic to recursively identify sink nodes and infer the topological ordering of a DAG. Another line of work employs regression-based methods to detect nonlinearity. For instance, the nonlinear invariant causal prediction (ICP) framework \citep{heinze2018invariant} fits regression models to data collected in different environments, and then tests for differences in residual distributions across environments after interventions to detect associations. 
\citet{monti2020causal} extend independence component analysis to the nonlinear setting and use the method to infer causal models. It utilizes non-stationary data and nonlinear independent component analysis to identify exogenous error variables, and then performs independence tests to determine causal directions. \citet{gretton2009nonlinear} consider the cases where the nonlinear function $f_{i}$ is invertible or where the noise is not additive, and demonstrate that only a PDAG is identifiable in these cases. Their method thereby focuses on identifying non-invertible ANMs in a Markov equivalence class (MEC) through iterative residual independence testing.
Improving upon an estimated CPDAG, \citet{wang_zhou_nonlinear} propose a statistical test that orients the undirected edges by comparing the goodness of fit of models corresponding to the two possible directions. With minimal assumptions on the regression model, this approach offers flexibility in modeling and greater applicability to real-world data.

However, these constraint-based methods also face practical limitations that hinder their widespread adoption. A significant drawback to the iterative testing approach is the computational cost, notably for kernel-based methods. As its runtime scales quadratically with sample size, the KCI test is computationally expensive to employ in constraint-based algorithms, and especially inefficient for learning larger DAGs \citep{strobl2019kci_rcot}. Additionally, the performance of the KCI test heavily depends on the chosen kernel function, which is often problem-specific and difficult to tune. The nonlinear ICP requires a sufficient number of observed environments to detect causal relations \citep{rosenfeld2021risks} and thus cannot be applied to data generated under a classical i.i.d. setting. Moreover, the accuracy of regression-based approaches is contingent on the quality of the estimated model, which requires strong domain knowledge for model selection and large sample sizes to accurately approximate the true SEM \citep{shah2020hardness}. Model misspecification can lead to violations of key assumptions, ultimately inflating false positive rates in independence tests \citep{li2020nonparametric}. More broadly, recent efforts tend to focus on developing independence tests -- without fully developing them into scalable, full-fledged causal discovery algorithms \citep{hasan2023survey} -- or approximating key statistical quantities to improve test efficiency, with less emphasis on using these metrics to infer causal relations.

Score-based methods for nonlinear learning have also seen significant development in recent years. The algorithm CAM \citep{cam}, for example, assumes an additive noise model with Gaussian errors and maximizes the joint log-likelihood function to learn a DAG. \citet{huang2018generalized} propose a score function by measuring independence in a reproducing kernel Hilbert space, thereby enabling a novel approach to learn the CPDAG under nonlinear relations and arbitrary distributions. \citet{ramsey2025scalable} obtain a set of embeddings for each variable using a truncated set of Legendre polynomials, allowing for fast approximation of nonlinear, additive SEMs. They propose a score function based on this method to estimate nonlinear CPDAGs. 
Several recent works have reformulated the structural learning task as a continuous-optimization problem by devising an algebraic characterization of DAGs \citep{notears}. Prominent examples include NOTEARS \citep{zheng2020learning}, DAG-GNN \citep{yu2019dag}, and DAGMA \citep{bello2023dagmalearningdagsmmatrices}, which incorporate deep learning techniques to enhance the flexibility of SEM estimation. Another line of work, exemplified by the SCORE algorithm \citep{rolland2022score}, iteratively identifies leaf nodes in a DAG using the Jacobian of the score function. However, score-based methods generally heavily depend on model assumptions, require intensive computational time, and are not guaranteed to find a globally optimal structure. Some of the above deep learning-based methods, in particular, perform poorly when the data is standardized, as standardization erases causal order information from the marginal variance when minimizing the least squares objective function \citep{cd_benchmarks_neurips2021}.

\subsection{Contribution of This Work} 

In this work, we propose a novel causal discovery algorithm SNOE (Sequential Nonlinear Orientation of Edges) for nonlinear DAG learning. SNOE builds upon the CPDAG learned by classical methods and sequentially determines the causal direction of undirected edges to learn the true DAG. 

In lieu of inferring the causal order of nodes, we introduce a local identifiability criterion based on a pairwise additive noise model (PANM). This criterion determines whether a given undirected edge in a partially directed acyclic graph (PDAG) can be correctly oriented. Edges that fulfill this criterion are ensured to be correctly oriented in the large-sample limit, without inducing errors in subsequent orientations. We prove that, at the population level, the algorithm consistently recovers the true DAG from its CPDAG.
Leveraging this result, we devise a sequential algorithm to identify undirected edges following the PANM criterion and infer their causal directions. As such, the algorithm effectively learns the DAG at a local scale by evaluating edges individually without necessitating evaluation of all nodes or the entire graph.
To determine the orientation of each edge, SNOE employs a likelihood-ratio test that compares the bivariate conditional probability distributions over the sub-DAG on both nodes and their learned parent sets under the competing directions. Contrary to general independence tests, the likelihood ratio test returns a definitive decision regarding the causal direction, thereby bridging the gap between the task of detecting nonlinear conditional independence relations and the structural learning problem. Empirical results further show that the test is robust to violations of model misspecification and yields accurate results across different functional settings.

The main contributions of this work are summarized below: 
\begin{itemize}

    \item A novel criterion derived from a pairwise additive noise model to determine the identifiability and correct orientation of undirected edges in a PDAG;
    \item An algorithm that is guaranteed at the population level to identify the true DAG from its Markov equivalence class by orienting undirected edges, according to their adherence to the PANM criterion, in a sequential manner;
    \item Theoretical results for the structural learning consistency of our algorithm in the large-sample limit;
    \item Higher accuracy and faster computation time compared to competing nonlinear DAG learning methods.
\end{itemize}
At a conceptual level, the sequential orientation algorithm is the most significant contribution of this work.
In essence, our method is rooted in this central idea: Starting from the CPDAG, there exists at least one undirected edge whose orientation can be determined by the PANM criterion at \textit{any} iteration in our sequential algorithm, leading to the correct recovery of the true DAG. In this process, we check whether the PANM is satisfied for an undirected edge only conditional on the identified parents of the two nodes.
This is made possible with a careful design of edge orientations by the PANM criterion and by a subset of the Meek's rules \citep{meek1995causal_background_knowledge}.
Although \citet{gretton2009nonlinear} also orients edges in a CPDAG, they use a greedy search to iteratively identify nodes satisfying a non-invertible ANM by examining all candidate parent sets of a node, i.e., by checking all subsets of its neighbors (connected to the node by an undirected edge).
This is clearly different from our sequential orientation procedure that only considers identified parents without the need to examine any subset of the neighbors. The method proposed by \citet{wang_zhou_nonlinear} ranks the undirected edges in a CPDAG by a goodness-of-fit metric, without establishing its rigorous property in terms of edge orientation or recovery of the true DAG. Moreover, they assume a specific piecewise linear SEM, while we consider a more general nonlinear ANM.
\citet{resit_anm} employs independence tests to identify a sink node in a sequential manner, which is distinct from the above sequential edge orientation methods.

The paper is structured as follows. Section~\ref{sec:prelim} introduces the fundamental concepts and model assumptions. In Section~\ref{sec:alg-overview}, we present the central notion of our work, the sequential edge orientation procedure and the finite-sample version of the algorithm. In Section~\ref{sec:edge-orientation}, we discuss the two key components of the edge orientation step: the ranking procedure and the orientation test. 
Then, we establish the structural learning consistency of the algorithm in Section~\ref{sec:theory}. Section~\ref{sec:experiments-simulated-data} presents the performance of our method against competing methods on simulated data, with a detailed analysis of some intermediate results. We further demonstrate its performance in causal discovery on real-world data in Section~\ref{sec:experiments-real-data}. Last, we summarize our work and outline directions for future research in Section~\ref{sec:conclusion}.
Proofs and supplemental numerical results are provided in the Appendix.

\section{Preliminaries} \label{sec:prelim}

\subsection{Directed Acyclic Graph}

A \textit{graph} $\calG=(V, E)$ consists of a set of vertices $V = [p]:=\{1, \ldots, p\}$ and a set of edges $E \subseteq V \times V$. For a pair of distinct nodes $i, j \in V$, a directed edge $i \rightarrow j \in E$ indicates that node $i$ is a parent of node $j$ and $j$ is a child of $i$. The parent set of node $i$ is denoted by $\pa_{\calG}(i)$ and the child set denoted by $\ch_{\calG}(i)$. In contrast, there may exist an undirected edge $i-j \in E$ in $\calG$, where $i$ is called a neighbor of $j$, i.e. $i \in \nbr_{\calG}(j)$, and vice versa. A \textit{directed acyclic graph} (DAG) $\calG$ consists only of directed edges and does not admit any directed cycles. A related type of graph is the \textit{partially directed acyclic graph} (PDAG), which contains both directed and undirected edges but does not contain any cycles in its directed subgraphs.

A \textit{causal DAG} is a structural causal model that employs a DAG $\calG$ to represent causal relations among random variables $X = \{X_1, \dots, X_p\}$. Specifically, these causal relations are described by SEMs of the form in~\eqref{eq:genSEM}, where $\PA_{i} = \{X_{j}: j \in \pa_\calG(i)\}$. The probability distribution over the noise variables $p(\varepsilon)=\prod_{i=1}^{p}p(\varepsilon_{i})$ induces a distribution $p(X)$ over $\{X_{i}\}_{i=1}^{p}$. In particular, the joint distribution $p(X_{1}, \ldots, X_{p})$ satisfies the \textit{Markov condition} as its density factorizes according to $\calG$:
\begin{equation}
    p(X_{1}, \ldots, X_{p}) = \prod_{i=1}^{p}p(X_{i}|\PA_{i}),
\end{equation}
where $p(X_{i}|\PA_{i})$ denotes the density of $X_{i}$ conditional on its parent set. The Markov condition implies that any $X_i$ is independent of its non-descendant nodes given its parents $\PA_i$. Hereafter, we identify the nodes $V$ and the random variables $X$.

To infer the structure of the graph $\calG$ from observed data, we require the \textit{causal sufficiency} and \textit{faithfulness} assumptions to hold \citep{pearl@2000}. \textit{Causal sufficiency} is satisfied when all common causes of any distinct pair of nodes are observed. In other words, there are no unobserved or latent confounders under this assumption. Suppose $A, B, C \subset [p]$ are disjoint subsets. A DAG $\calG$ and a probability distribution $P$ are \textit{faithful} to one another if the conditional independence relations in $P$ have a one-to-one correspondence with the d-separation relations in $\calG$: $A\perp\!\!\!\perp B | C \text{ in } P \Longleftrightarrow C$ d-separates $A, B$ in $\calG$. Faithfulness also implies the causal minimality condition. The pair $(\calG, P)$ satisfies the \textit{causal minimality} condition if $P$ is not Markov to any proper subgraph of $\calG$ over the vertex set $V$ \citep{causation_search_prediction}.

\subsection{Markov Equivalence Class}

Two DAGs are \textit{Markov equivalent} if and only if they have identical skeletons and v-structures, which are ordered triplets of nodes $i, j, k$ oriented as $i \rightarrow k \leftarrow j$ with no edge between $i, j$ \citep{verma1990}. Markov equivalent DAGs encode the same set of d-separations and form an equivalence class. Without further restrictions on the function classes in the SEM~\eqref{eq:genSEM}, Markov equivalent DAGs cannot be distinguished through observational data; hence we can only learn their equivalence class. The equivalence class is represented by a \textit{completed partially directed acyclic graph} (CPDAG), a PDAG with specific structural properties \citep{andersson1997characterization}. Every directed edge is compelled, or strongly protected, and every undirected edge is reversible in the CPDAG.

The CPDAG $\mathcal{E}$ of a DAG $\calG$ is typically obtained by first identifying v-structures in the skeleton, and then applying \textit{Meek's rules}, a set of four rules that orient edges based on graphical patterns \citep{meek1995causal_background_knowledge}. Meek's rules identify additional directed edges in the graph without introducing new v-structures. A \textit{maximally oriented PDAG} is a PDAG for which no edges can be further oriented by Meek's rules. As an example, a CPDAG $\mathcal{E}$ is a maximally oriented PDAG, or a maximal PDAG for short.

A \textit{consistent extension} of a PDAG $\calG$ is a DAG, $\widetilde{\calG}$, obtained by orienting all undirected edges in $\calG$ without introducing new v-structures. Therefore, $\widetilde{\calG}$ has the same skeleton, the same orientations of all directed edges in $\calG$, and the same v-structures as $\calG$ \citep{Dor1992ASA}. While not all PDAGs can be extended to a DAG, a CPDAG $\mathcal{E}$ is extendable since each DAG in the equivalence class represented by $\mathcal{E}$ is a consistent extension of $\mathcal{E}$. A DAG may be obtained from a CPDAG by iteratively making edge orientations without introducing new v-structures and applying Meek's rules, while preserving the compelled edges \citep{pmlr-v161-wienobst21a-extendability}.

\subsection{Additive Noise Models}
There are recent developments on learning causal DAGs from observational data, assuming the \textit{additive noise model} (ANM) \citep{hoyer2008nonlinear}. Under the ANM, each variable $X_{i}$ is a function of its parent nodes $\PA_{i}$ in DAG $\calG_{0}$ plus an independent additive noise $\varepsilon_{i}$, i.e.
\begin{align}
X_{i} = f_{i}(\PA_{i}) + \varepsilon_{i}, \qquad i = 1,\ldots,p,
\label{eq:anm}
\end{align}
where $f_{i}$ is an arbitrary function for each $i$ and the noise variables are jointly independent. Moreover, the parents and the noise term are independent of each other, i.e. $\PA_{i} \indp \varepsilon_{i}$.
A \textit{restricted additive noise model} is an ANM with restrictions on the functions $f_i$, conditional distributions of $X_i$, and noise variables \citep{resit_anm}. In particular, the functions must be three-times continuously differentiable. See Definition~\ref{eq:restricted-anm} in the Appendix for the full definition. Throughout this work, we assume causal minimality for the ANM, which is satisfied as long as the function $f_{i}$, for all $i$, is not constant in any of its arguments \citep{resit_anm}. Under the causal minimality assumption, a key result is that the true DAG $\calG_{0}$ can be identified from the joint distribution $p(X_1, \dots, X_p)$ when the SEM for $\{X_{i}\}_{i=1}^{p}$ satisfies a restricted additive noise model.

We also consider the \textit{causal additive model} (CAM), a special case of the ANM where the function $f_i$ is additive \citep{cam}. The model is defined as
\begin{align}
X_{i} = \sum_{j \in \pa(i)} f_{i, j}(X_{j}) + \varepsilon_{i}, \qquad i = 1,\ldots,p,
\label{eq:cam}
\end{align}
where $f_{i,j}$ are three times differentiable, nonlinear functions and the error terms $\varepsilon_i \sim N(0, \sigma_i^2)$ independently with $\sigma_i^2 > 0$. \cite{cam} have demonstrated that the true DAG $\calG_{0}$ can be identified from the joint distribution $p(X_1, \dots, X_p)$ when the SEM for $\{X_{i}\}_{i=1}^{p}$ satisfies a CAM. Furthermore, the source nodes are also allowed to have a non-Gaussian density. As the true SEM under a general ANM is difficult to recover in a practical setting, assuming a CAM assists in better recovering  the underlying causal relations, by utilizing regression models such as the generalized additive model.

Let $\mathcal{F}$ be a class of three times differentiable univariate functions that is closed with respect to the $\mathcal{L}_{2}$ norm of $P(X_{i}), i=1, \ldots, p$. We define a space of additive functions 
\begin{align}
    \mathcal{F}^{\oplus k} &:= \left\{f: \mathbb{R}^{k} \rightarrow \mathbb{R}, f(x) = \sum_{i=1}^{k}f_{i}(x_{i}), \quad f_{i} \in \mathcal{F}\right\}, \quad k\in[p].
\label{eq:additivefx}
\end{align}

\section{Algorithm Overview}\label{sec:alg-overview}

Classical causal discovery methods learn an equivalence class of the true DAG, represented by a CPDAG. Since nonlinear DAG models are identifiable \citep{resit_anm}, we aim to further infer the causal directions of undirected edges in a CPDAG or, more generally, a PDAG for the finite sample case. To this end,
we formulate a novel criterion, named the \textit{pairwise additive noise model} (PANM), to determine which undirected edges in a PDAG can be correctly oriented at the current stage in a sequential manner. 
Once an edge fulfilling this criterion is oriented, we further apply additional graphical rules to orient more undirected edges. 
We show that this sequential procedure is able to orient all undirected edges in a CPDAG into the true causal DAG, thus accomplishing
the goal of identifying all causal relations among the variables.
In this section, we introduce the pairwise additive noise model, the sequential edge orientation procedure, and the finite sample version of our full algorithm.

\subsection{Pairwise Additive Noise Model}

The pairwise additive noise model, a set of SEMs defined with respect to two nodes, encapsulates the conditions sufficient for edge orientation. For an undirected edge $X-Y$ in a PDAG $\calG$, given their parent sets $\pa_{\calG}(X)$ and $\pa_{\calG}(Y)$, its causal direction in true DAG $\calG_{0}$ can be identified when $(X,Y)$ follows a PANM. 
In certain cases, a pair of nodes may follow a PANM even if their parent sets in $\calG_0$ are not fully identified, i.e. $\pa_{\calG}(v) \subset \pa_{\calG_{0}}(v)$. 
We demonstrate how this criterion lays the foundation for our edge orientation procedure.
\begin{definition}[Pairwise Additive Noise Model]
Let $X,Y$ be two random variables and $Z_1,Z_2$ be two sets of random variables. We say that $[X,Y\mid Z_1,Z_2]$ follows a \textit{pairwise additive noise model} if either (i) or (ii) holds: 
\begin{enumerate}
    \item[(i)] $X=f_X(Z_1)+\varepsilon_X$, $\varepsilon_X\indp Z_1$ and $Y=f_Y(X,Z_2)+\varepsilon_Y$, $\varepsilon_Y\indp \{X,Z_2\}$,
    \item[(ii)] $X=f_X(Y,Z_1)+\varepsilon_X$, $\varepsilon_X\indp \{Y,Z_1\}$ and $Y=f_Y(Z_2)+\varepsilon_Y$, $\varepsilon_Y\indp Z_2$.
\label{eq:panm-def}
\end{enumerate}
Furthermore, we assume both SEMs above satisfy Condition~\ref{cond:anm-fx-ident} in Appendix~\ref{appendix:restricted-anm-def} for any value $(z_1,z_2)$ in the domain of $(Z_1,Z_2)$.
\end{definition}

Within the context of causal learning, the structural equation models in conditions (i) and (ii) correspond to the additive noise models under $X \rightarrow Y$ and $Y \rightarrow X$, respectively, where $Z_{1} = \pa_{\calG}(X)$ and $Z_{2} = \pa_{\calG}(Y)$ in PDAG $\calG$. In this regard, we may simply say that the undirected edge $X-Y$ or the pair of nodes $(X,Y)$ in the PDAG $\calG$ satisfies the PANM. To determine whether $(X,Y)$ follows a PANM, we can test if the independence relations hold. More specifically, only the PANM constructed under the correct orientation of $X-Y$ yields the independence relations between the noise term and the parent variables. The precise statement is summarized into the following lemma, which is an immediate consequence of the identifiability of bivariate ANMs~\citep{hoyer2008nonlinear}.

\begin{lemma}
Assume $\{X_{i}\}_{i=1}^{p}$ follows a restricted ANM with respect to a DAG $\calG_0$. Suppose two nodes $X, Y$ are connected by an undirected edge in a PDAG $\calG$ which has a consistent extension to $\calG_{0}$. 
If $[X, Y | \pa_{\calG}(X), \pa_{\calG}(Y)]$ follows the PANM, then the causal direction between $X$ and $Y$ is identifiable.
\label{lem:panm-identifiability}
\end{lemma}

The connection between the PANM and the identifiability of the causal direction is more concretely illustrated in Figure~\ref{fig:panm-example}, showing cases when the true causal direction can and cannot be recovered. Depicted in example (a), edge $X-Y$ can be correctly oriented because the causal relations depicted in the DAG (top) yield the correctly specified pairwise ANM and the true parent sets $\pa_{\calG}(X) = \pa_{\calG}(Y) = \{A, B\}$, except the relation between $X$ and $Y$, are identified in the PDAG (bottom). The independence relations hold for SEMs corresponding to $X \rightarrow Y$.  
Panel (b) shows an example in which an undirected edge satisfies the PANM even though the parent sets are not fully identified. The SEMs corresponding to $X \rightarrow Y$ are $Y = f_Y(X, B) + \varepsilon_{Y}$, which is the true SEM, and $X=\widetilde{\varepsilon}_{X}$, where $\widetilde{\varepsilon}_{X} = g(A, B)+\varepsilon_{X}$. While the parents $A, B$ for $X$ are not detected in the PDAG as $\pa_{\calG}(X) = \varnothing$, they are merged with the error term $\varepsilon_X$ to form the new error $\widetilde{\varepsilon}_{X}$.
Thus, $X-Y$ satisfies the PANM and its orientation is identifiable.

\begin{figure}[h]
\centering
\begin{tikzpicture}[node distance={12mm}, thick, main/.style = {draw, circle}] 
\node[main] (A) {$A$}; 
\node[right of=A] (temp1) { }; 
\node[main] (B) [right of=temp1] {$B$};
\node[main] (X) [below of=A] {$X$};
\node[right of=X] (temp2) { }; 
\node[main] (Y) [right of=temp2] {$Y$};
\node[main] (Z) [below of=temp2] {$Z$};
\draw[->] (A) -- (X); 
\draw[->] (A) -- (Y);
\draw[->] (A) to [out=300,in=110] (Z); 
\draw[->] (B) -- (X); 
\draw[->] (B) -- (Y);
\draw[->] (B) to [out=240,in=70] (Z); 
\draw[->] (X) -- (Y); 
\draw[->] (X) -- (Z);
\draw[->] (Y) -- (Z); 
\end{tikzpicture} 
\hspace{0.5cm}
\begin{tikzpicture}[node distance={12mm}, thick, main/.style = {draw, circle}] 
\node(temp1) { }; 
\node[main] (A) [above of=temp1] {$A$}; 
\node[main] (B) [left of=temp1] {$B$};
\node[main] (X) [right of=temp1] {$X$};
\node[main] (Y) [below of=temp1] {$Y$};
\draw[->] (A) -- (B); 
\draw[->] (A) -- (X);
\draw[->] (B) -- (X);
\draw[->] (A) -- (B); 
\draw[->] (X) -- (Y);
\draw[->] (B) -- (Y);
\end{tikzpicture} 
\hspace{0.5cm}
\begin{tikzpicture}[node distance={12mm}, thick, main/.style = {draw, circle}] 
\node[main] (A) {$A$}; 
\node[right of=A] (temp1) { }; 
\node[main] (B) [right of=temp1] {$B$};
\node[main] (Z) [below of=A] {$Z$};
\node[right of=Z] (temp2) { }; 
\node[main] (X) [right of=temp2] {$X$};
\node[main] (Y) [below of=temp2] {$Y$};
\draw[->] (A) -- (Z); 
\draw[->] (A) -- (X);
\draw[->] (A) to [out=300,in=110] (Y); 
\draw[->] (B) -- (Z); 
\draw[->] (B) -- (X);
\draw[->] (B) to [out=240,in=70] (Y); 
\draw[->] (Z) -- (X); 
\draw[->] (Z) -- (Y); 
\draw[->] (X) -- (Y);
\end{tikzpicture} 
\hspace{0.5cm}
\begin{tikzpicture}[node distance={12mm}, thick, main/.style = {draw, circle}] 
\node(temp1) { }; 
\node[main] (X) [above of=temp1, node distance=15mm] {$X$}; 
\node[main] (A) [below left of=temp1] {$A$};
\node[main] (Y) [below right of=temp1] {$Y$};
\draw[->] (X) -- (A); 
\draw[->] (X) -- (Y);
\draw[->] (A) -- (Y); 
\end{tikzpicture}

\begin{tikzpicture}[node distance={12mm}, thick, main/.style = {draw, circle}] 
\node[main] (A) {$A$}; 
\node[right of=A] (temp1) { }; 
\node[main] (B) [right of=temp1] {$B$};
\node[main] (X) [below of=A] {$X$};
\node[right of=X] (temp2) { }; 
\node[main] (Y) [right of=temp2] {$Y$};
\node[main] (Z) [below of=temp2] {$Z$};
\node[below of=Z, node distance=10mm] {(a)};
\draw[->] (A) -- (X); 
\draw[->] (A) -- (Y);
\draw[->] (A) to [out=300,in=110] (Z); 
\draw[->] (B) -- (X); 
\draw[->] (B) -- (Y);
\draw[->] (B) to [out=240,in=70] (Z); 
\draw [color=red, very thick] (X) -- (Y); 
\draw (X) -- (Z);
\draw (Y) -- (Z); 
\end{tikzpicture} 
\hspace{0.5cm}
\begin{tikzpicture}[node distance={12mm}, thick, main/.style = {draw, circle}] 
\node(temp1) { }; 
\node[main] (A) [above of=temp1] {$A$}; 
\node[main] (B) [left of=temp1] {$B$};
\node[main] (X) [right of=temp1] {$X$};
\node[main] (Y) [below of=temp1] {$Y$};
\node[below of=Y, node distance=10mm] {(b)};
\draw (A) -- (B); 
\draw (A) -- (X);
\draw (B) -- (X);
\draw[color=red, very thick]  (X) -- (Y);
\draw[->] (B) -- (Y);
\end{tikzpicture} 
\hspace{0.5cm}
\begin{tikzpicture}[node distance={12mm}, thick, main/.style = {draw, circle}] 
\node[main] (A) {$A$}; 
\node[right of=A] (temp1) { }; 
\node[main] (B) [right of=temp1] {$B$};
\node[main] (Z) [below of=A] {$Z$};
\node[right of=Z] (temp2) { }; 
\node[main] (X) [right of=temp2] {$X$};
\node[main] (Y) [below of=temp2] {$Y$};
\node[below of=Y, node distance=10mm] {(c)};
\draw[->] (A) -- (Z); 
\draw[->] (A) -- (X);
\draw[->] (A) to [out=300,in=110](Y); 
\draw[->] (B) -- (Z); 
\draw[->] (B) -- (X);
\draw[->] (B) to [out=240,in=70] (Y); 
\draw (Z) -- (X); 
\draw (Z) -- (Y); 
\draw[color=red, very thick] (X) -- (Y);
\end{tikzpicture} 
\hspace{0.5cm}
\begin{tikzpicture}[node distance={12mm}, thick, main/.style = {draw, circle}] 
\node(temp1) { }; 
\node[main] (X) [above of=temp1, node distance=15mm] {$X$}; 
\node[main] (A) [below left of=temp1] {$A$};
\node[main] (Y) [below right of=temp1] {$Y$};
\node[below of=temp1, node distance=18mm] {(d)};
\draw (X) -- (A); 
\draw[color=red, very thick] (X) -- (Y);
\draw (A) -- (Y); 
\end{tikzpicture} 
\caption{Examples to illustrate the PANM. The top row shows the true DAG, while the bottom row features a PDAG extendable to the true DAG with the evaluated edge $X-Y$. (a) $[X, Y\mid A,B]$ satisfies the PANM because both parent sets are fully identified. (b) $[X, Y\mid\varnothing, B]$ satisfies the PANM, despite $A, B$ missing from $\pa_{\calG}(X)=\varnothing$, since we can write $X = \widetilde{\varepsilon}_{X} = g(A, B) + \varepsilon_{X}$. (c) $[X, Y \mid A, B]$ does not form a PANM, as common parent $Z$ is not identified and becomes a latent confounder in the model. (d) $[X, Y]$ does not satisfy the PANM since node $Y$ is missing parent $A$, which does not guarantee $\varepsilon_{Y} \indp X$.} 
\label{fig:panm-example}
\end{figure}

In example (c), however, the common parent node $Z$ is not identified in the PDAG. The SEMs constructed over $[X, Y \mid A, B]$ would not satisfy either set of independence relations in Definition~\ref{eq:panm-def} due to the presence of a hidden confounder $Z$, which would result in incorrect or inconclusive inference on the causal direction. In some cases, the independence relations entailed by the PANM may not hold if a parent node, say of $Y$, on a directed path from $X$ to $Y$ is not identified. Consider the undirected edge $X-Y$ in example (d), where $\pa_\calG(X)=\pa_\calG(Y)=\varnothing$ in the PDAG. Node $A$ is a parent of $Y$ in the DAG $\calG_0$ on the directed path $X\to A\to Y$, but it is not identified as a parent of $Y$ in the PDAG. We now examine whether the model $[X,Y]$ satisfies PANM under $X \rightarrow Y$. Node $X$ has no parents in the true DAG, hence its SEM $X = \varepsilon_{X}$ matches the true form. Yet, node $Y$ can only be expressed as $Y = f_Y(X, A) + \varepsilon_{Y} = f_Y(X, g(X) + \varepsilon_{A}) + \varepsilon_{Y}$, where $A$ is substituted as $A = g(X) + \varepsilon_A$ and marginalized out. This shows that the SEM for $Y$ is not an additive noise model. Suppose $\widehat{Y}$ is the best approximation of $Y$ by functions of $X$ assuming an additive noise. Then, the residual $Y - \widehat{Y}$ in general depends on $X$, so the independence relation $(Y - \widehat{Y}) \indp X$ does not hold.

\subsection{Key Idea: Sequential Edge Orientation} \label{sec:alg-key-idea}

Central to our algorithm is a sequential edge orientation procedure. Given a PDAG $\calG$, the procedure aims to determine the true causal direction of undirected edges in the graph. The core idea is to identify an undirected edge that satisfies the pairwise additive noise model, and then conduct a statistical test to determine its exact direction. We show that this procedure can be performed sequentially until all edges are oriented.

We present the high-level, population version of the sequential edge orientation procedure in Algorithm~\ref{alg:sneo-population-version} to demonstrate this core idea. Given a PDAG, the algorithm identifies an edge $(i, j)$ that satisfies the PANM on Line~\ref{lst:line:snoe-population-edge-to-orient} and then orients the edge into its true causal direction, which can be identified (Lemma~\ref{lem:panm-identifiability}). Subsequently, the algorithm leverages information read from $\calG$ to further identify common children of $i$ and $j$ on Line~\ref{lst:line:snoe-population-orient-nc}, where we denote the set of neighbors and children of $i$ as $\nc_{\calG}(i) := \nbr_{\calG}(i) \cup \ch_{\calG}(i)$. For each node $k \in \nc_{\calG}(i) \cap \nc_{\calG}(j)$, we orient the edges $i \rightarrow k$ and $j \rightarrow k$. 
Figure~\ref{fig:snoe-orient-nc} features the three scenarios in which Line~\ref{lst:line:snoe-population-orient-nc} is applicable: 
(1) $k\in\nbr_{\calG}(i) \cap \nbr_{\calG}(j)$, (2) $k\in\nbr_{\calG}(i) \cap \ch_{\calG}(j)$, and (3) $k \in \ch_{\calG}(i) \cap \nbr_{\calG}(j)$. Case 2 and case 3 respectively correspond to when $j \rightarrow k$ and $i \rightarrow k$ have been oriented by prior actions before evaluating $i-j$. For all three cases, we orient $k$ as a common child of $i$ and $j$ as shown in the bottom panel of Figure~\ref{fig:snoe-orient-nc}. This is because of the following reasoning: If $k$ were a common parent node of $i, j$, then $[i, j\mid \pa_\calG(i),\pa_\calG(j)]$ would not satisfy the independence relations entailed by the PANM since $k$ would be a hidden confounder, as discussed in Figure~\ref{fig:panm-example}c. If the true orientation were $i\to k\to j$ or $j\to k \to i$, this would be the case of Figure~\ref{fig:panm-example}d and again would violate the PANM assumptions.
Therefore, $k$ must be a child node of both $i$ and $j$. 
On the following line, the procedure applies rule 1 of Meek's orientation rules, where the configuration $a \rightarrow b - c$ is oriented as $a \rightarrow b \rightarrow c$ given there is no edge between $a$ and $c$, to identify descendant nodes of $a$.
Finally, if no undirected edges satisfy the condition on Line~\ref{lst:line:snoe-population-edge-to-orient}, we apply the Meek's rules to maximally orient the PDAG on Line~\ref{lst:line:sneo-population-all-meeks-rules}.

\begin{algorithm}[t]{
    \KwIn{PDAG $\calG=(V, E)$ and its undirected edges $U = E_{U}(\calG)$} 

    \While{$|U| > 0$}{\label{lst:line:snoe-while-loop-start}
        Search for $(i, j)\in U$ such that $[i, j\mid \pa_\calG(i),\pa_\calG(j)]$ satisfies the PANM\; \label{lst:line:snoe-population-edge-to-orient}
        \If{such $(i, j)$ is found}{
            Identify the causal direction between $i,j$ and orient $(i,j)$ in $\calG$ accordingly\; \label{lst:line:snoe-population-orient-found-edge}
            (suppose $i\to j$ hereafter)\;
            \If{$\nc_{\calG}(i) \cap \nc_{\calG}(j) \neq \varnothing$}{ \label{lst:line:snoe-population-orient-nc-condition}
                Orient $i\to k$ and $j\to k$ in $\calG$ for all $k \in \nc_{\calG}(i) \cap \nc_{\calG}(j)$\; \label{lst:line:snoe-population-orient-nc}
            }  
            Apply Meek's orientation rule 1 to $\calG$\ repeatedly until it cannot be applied\; \label{lst:line:snoe-population-meeks-rules}
            Update $U \gets E_{U}(\calG)$;
        }
        \Else{
            \textbf{break}
        }
    }
    Apply all of Meek's rules to $\calG$ repeatedly until none of them can be further applied. \label{lst:line:sneo-population-all-meeks-rules}
}
\caption{Sequential Orientation of Edges \textit{(SequentialOrientation)}}
\label{alg:sneo-population-version}
\end{algorithm}

\begin{figure}
\centering
\begin{tikzpicture}[node distance={14mm}, thick, main/.style = {draw, circle}] 
\node(temp1) { }; 
\node[main] (i) [below left of =temp1]{$i$}; 
\node[main] (j) [below right of=temp1] {$j$};
\node[main] (k) [below right of=i] {$k$};
\draw[->] (i) -- (j);
\draw (i) -- (k); 
\draw (j) -- (k); 
\end{tikzpicture} 
\hspace{1.75cm}
\begin{tikzpicture}[node distance={14mm}, thick, main/.style = {draw, circle}] 
\node(temp1) { }; 
\node[main] (i) [below left of =temp1]{$i$}; 
\node[main] (j) [below right of=temp1] {$j$};
\node[main] (k) [below right of=i] {$k$};
\draw[->] (i) -- (j);
\draw (i) -- (k); 
\draw[->] (j) -- (k); 
\end{tikzpicture} 
\hspace{1.75cm}
\begin{tikzpicture}[node distance={14mm}, thick, main/.style = {draw, circle}] 
\node(temp1) { }; 
\node[main] (i) [below left of =temp1]{$i$}; 
\node[main] (j) [below right of=temp1] {$j$};
\node[main] (k) [below right of=i] {$k$};
\draw[->] (i) -- (j);
\draw[->] (i) -- (k); 
\draw (j) -- (k); 
\end{tikzpicture} 
\begin{tikzpicture}[node distance={14mm}, thick, main/.style = {draw, circle}] 
\node(temp1) { }; 
\node[main] (i) [below left of =temp1]{$i$}; 
\node[main] (j) [below right of=temp1] {$j$};
\node[main] (k) [below right of=i] {$k$};
\node[below of=k, node distance = 10mm] {(1) $k \in \nbr_{\calG}(i) \cap \nbr_{\calG}(j)$};
\draw[->] (i) -- (j);
\draw[->, color=red] (i) -- (k); 
\draw[->, color=red] (j) -- (k); 
\end{tikzpicture} 
\hspace{0.5cm}
\begin{tikzpicture}[node distance={14mm}, thick, main/.style = {draw, circle}] 
\node(temp1) { }; 
\node[main] (i) [below left of =temp1]{$i$}; 
\node[main] (j) [below right of=temp1] {$j$};
\node[main] (k) [below right of=i] {$k$};
\node[below of=k, node distance = 10mm] {(2) $k \in \nbr_{\calG}(i) \cap \ch_{\calG}(j)$};
\draw[->] (i) -- (j);
\draw[->, color=red] (i) -- (k); 
\draw[->] (j) -- (k); 
\end{tikzpicture} 
\hspace{0.5cm}
\begin{tikzpicture}[node distance={14mm}, thick, main/.style = {draw, circle}] 
\node(temp1) { }; 
\node[main] (i) [below left of =temp1]{$i$}; 
\node[main] (j) [below right of=temp1] {$j$};
\node[main] (k) [below right of=i] {$k$};
\node[below of=k, node distance = 10mm] {(3) $k \in \ch_{\calG}(i) \cap \nbr_{\calG}(j)$};
\draw[->] (i) -- (j);
\draw[->] (i) -- (k); 
\draw[->, color=red] (j) -- (k); 
\end{tikzpicture} 

\caption{Orientation rules of Line~\ref{lst:line:snoe-population-orient-nc} in Algorithm~\ref{alg:sneo-population-version}. For each of the three cases in which $k \in \nc_{\calG}(i) \cap \nc_{\calG}(j)$ (top panel), 
we show the corresponding orientation of the red edge(s) in the bottom panel.}
\label{fig:snoe-orient-nc}
\end{figure}

Now we present a main result on Algorithm~\ref{alg:sneo-population-version}:
\begin{theorem}
Suppose that $(X_1,\ldots,X_p)$ follows a restricted additive noise model with respect to a DAG $\calG_0$. If the input $\calG$ is the CPDAG of $\calG_0$, then the sequential orientation procedure in Algorithm~\ref{alg:sneo-population-version} orients $\calG$ into the DAG $\calG_0$.
\label{theorem:population-eo}
\end{theorem}
Theorem~\ref{theorem:population-eo} shows that the edge orientation procedure in Algorithm~\ref{alg:sneo-population-version} can recover the true DAG from its CPDAG. A proof is provided in Appendix~\ref{app:proof-theorem-1}. The key of the proof is to show that there always exists an undirected edge $(i, j)$ in $\calG$ that meets the condition in Line~\ref{lst:line:snoe-population-edge-to-orient} as long as there are still undirected edges in $\calG$. This is achieved by the careful design of the orientation rules from Line~\ref{lst:line:snoe-population-orient-found-edge} to Line~\ref{lst:line:snoe-population-meeks-rules}. To illustrate this point, suppose we did not include the orientation rule on Line~\ref{lst:line:snoe-population-orient-nc} after orienting $i \rightarrow j$. As exemplified in case 1 of Figure~\ref{fig:snoe-orient-nc}, neither remaining undirected edges $i-k$ nor $j-k$ would satisfy the PANM.
When evaluating $j-k$, node $i$ is now a latent parent of $k$ that would yield an error term dependent on $X_j$ even under the true orientation $j \rightarrow k$.
The case of $i-k$ corresponds to that in Figure~\ref{fig:panm-example}d, which does not satisfy the PANM as we discussed. 
Therefore, the orientation rule on Line~\ref{lst:line:snoe-population-orient-nc} not only identifies additional causal relations, but also ensures that there exists an undirected edge satisfying the PANM in the next iteration. Details on the existence of such an edge are expounded on in the proof. 

In essence, our algorithm recovers the true DAG through two key steps: (1) to identify an edge $(i,j)$ that satisfies the PANM model; (2) to infer the causal direction of $(i, j)$ after it is identified. These key steps are achieved by our edge ranking and edge orientation procedures, which are introduced in Section~\ref{sec:edge-orientation}. Precisely, we propose a criterion based on the pairwise additive noise model to identify an undirected edge for orientation and develop a likelihood-ratio test to infer its causal direction.

\subsection{Algorithm Outline}

The full SNOE algorithm is formally described in Algorithm~\ref{alg:sneo-practical-version}, which implements the key idea of Algorithm~\ref{alg:sneo-population-version} through three main steps: first to learn the initial CPDAG structure, then to orient the undirected edges in the CPDAG, and lastly to remove extraneous edges. In the most general case, the final output is a PDAG. However, practitioners may choose to output a DAG if they assume it is identifiable. See Remark~\ref{remark:alg-step-4} in Section~\ref{sec:edge-orientation-lrt} for details.

\begin{algorithm}[h]{
    \KwIn{Observed data $X=(X_{1}, ..., X_{p})$, complete undirected graph $\calG=(V, E)$, sig. levels $\alpha_{1}, \alpha_{2}, \text{where } \alpha_{2} >\alpha_{1}$}
    \KwOut{PDAG $\calG=(V, E)$}

    \For{$(i, j) \in E$}{\label{lst:line:alpha2skelstart}
        Search for separating set $S_{ij} \subseteq V$ such that $\textit{p-val}(X_{i} \indp X_{j} | S_{ij}) > \alpha_{2}$\; \label{lst:line:alpha2sepset}
        Update $E \gets E \setminus \{(i, j), (j, i)\}$ and store $S_{ij}$ \text{ if found}\; \label{lst:line:alpha2citest} 
    }
    $E_{\alpha_2} \gets E$\; \label{lst:line:alpha2skelfinal}
    \For{$(i, j) \in E$}{ \label{lst:line:alpha1skelstart}
        Search for separating set $S_{ij} \subseteq V$ such that $\textit{p-val}(X_{i} \indp X_{j} | S_{ij}) > \alpha_{1}$\; \label{lst:line:alpha1sepset}
        Update $E \gets E \setminus \{(i, j), (j, i)\}$ and store $S_{ij}$ \text{ if found}\; \label{lst:line:alpha1citest}
    }
    Detect v-structures given $E$ and $\{S_{ij}\}$\; \label{lst:line:alpha2vstruct}
    Orient remaining undirected edges by Meek's rules\; \label{lst:line:alpha2meeks}
    
    Obtain candidate edge set $U_{\alpha_{2}} \gets E_{\alpha_2} \setminus E$\; \label{lst:line:getalpha1edges}
    Merge edge sets to obtain $\calG = (V, E \cup U_{\alpha_{2}})$\; \label{lst:line:stage1pdag}
    
    Orient undirected edges in $\calG: \textit{OrientEdges}(X, \calG, \alpha_{1})$\; \label{lst:line:stage2pdag}

    \For{$i=1,\ldots,p$}{ \label{lst:line:startprune}
        GAM regression $X_i\sim \{f_{i,k}(X_k):k\in \pa_{\calG}(i) \cup \nbr_{\calG}(i)\}$ and obtain $\textit{p-val}(f_{i,k}(X_k))$ from significance testing for each $k$\; \label{lst:line:prunecovartest}
        \If{$\textit{p-val}(f_{i,k}(X_k)) > \alpha_{1}$}{ \label{lst:line:removeinsigpa}
            Update $E \gets E \setminus \{(k,i)\}$\; \label{lst:line:removeinsigedge}
        }
    }
}
\caption{Causal Discovery by SNOE \label{alg:sneo-practical-version}}
\end{algorithm}

First, we apply a modified version of the PC algorithm \citep{spirtes1991algorithm} to learn the initial structure (Lines \ref{lst:line:alpha2skelstart}$\textendash$\ref{lst:line:alpha2meeks}). Specifically, we employ two significance levels: a stringent threshold $\alpha_{1}$ to learn the CPDAG and a relaxed threshold $\alpha_{2}$ to obtain a set of candidate edges. In our implementation, we use the partial correlation test to detect conditional independence relations. Starting with a complete, undirected graph, the PC algorithm removes edge $(i, j)$ if nodes $(X_{i}, X_{j})$ are independent given a subset of their neighbors $S_{ij}$, tested at significance level $\alpha_{1}$ (Lines~\ref{lst:line:alpha2skelstart}$\textendash$~\ref{lst:line:alpha1citest}). To obtain the candidate edges, we in fact first learn the skeleton using the relaxed significance level $\alpha_{2}$ in the conditional independence tests, resulting in a denser skeleton as described on Lines~\ref{lst:line:alpha2skelstart}$\textendash$~\ref{lst:line:alpha2skelfinal}. The candidate edges $U_{\alpha_{2}}$ (Line~\ref{lst:line:getalpha1edges}) are the edges removed when continuing the skeleton learning phase with $\alpha_{1}$, and then are reintroduced to form the graph $\calG=(V, E \cup U_{\alpha_{2}})$. The procedure is practically equivalent to learning a CPDAG under a strict significance level, then adding undirected edges between pairs of nodes with moderate association for consideration. We essentially separate skeleton learning and edge orientation into two tasks to obtain v-structures and directed edges with higher confidence, while preserving candidate edges to reduce the number of missing edges in the graphical structure. Although our work utilizes the PC algorithm, any causal discovery algorithm that learns the equivalence class of DAG $\calG_{0}$, with multiple sparsity levels, would be compatible with our method.

The second stage aims to determine the true causal direction of undirected edges in the CPDAG. This is accomplished through our orientation procedure \textit{OrientEdges}, which finds an evaluation order for undirected edges and then identifies their causal directions. To ensure that the undirected edges are correctly oriented in a sequential manner, we develop a measure to recursively rank undirected edges by their likelihood of satisfying the independence relations implied by the PANM. Then to orient an undirected edge $X-Y$, the edge orientation test \textit{LikelihoodTest}, described in Algorithm~\ref{alg:Likelihood Test}, computes a likelihood ratio to compare the competing directions $X \rightarrow Y$ and $Y \rightarrow X$ given their learned parent sets $\PA_{X}, \PA_{Y}$ in the current PDAG $\calG$. The test provides a definitive decision to either orient the edge in the preferred direction, if statistically significant, or leave it as undirected. The full details of the edge orientation procedure are presented in Algorithm~\ref{alg:OrientEdges}.

In the third and last step, the algorithm removes extraneous edges in the graphical structure by covariate selection (Lines \ref{lst:line:startprune} $\textendash$ \ref{lst:line:removeinsigedge}). Since the graph may contain undirected edges, the algorithm also considers neighbors when performing covariate selection. Recall that a neighbor of $X$ is a node that shares an undirected edge with $X$, excluding the parents and children of $X$ in the graph. For a node $X_{i}$, the algorithm regresses $X_{i}$ on its parents $\pa_{\calG}(X_{i})$ and neighbors $\nbr_{\calG}(X_{i})$ using a generalized additive model (GAM). We perform significance testing and remove incoming edges from statistically insignificant nodes. For a neighbor $X_{j} \in \nbr_{\calG}(X_{i})$, the edge is oriented as $X_{j} \rightarrow X_{i}$ if $f_{i, j}(X_{j})$ is statistically significant in the model for $X_{i}$ and $f_{j,i}(X_{i})$ is not significant in the model for $X_{j}$. If both terms are insignificant, the undirected edge is removed from the PDAG; otherwise, it remains intact.

\begin{remark}
In the implementation of Algorithm~\ref{alg:sneo-practical-version}, we assume a causal additive model~\eqref{eq:cam} for each node $i$. Accordingly, we use GAMs to complete all regression analysis in the algorithm. Since Theorem~\ref{theorem:population-eo} applies to any identifiable additive noise model, one may replace GAM with other nonlinear regression techniques for a more general functional form of $f_i(\PA_i)$.
\end{remark}

\section{Nonlinear Edge Orientation} \label{sec:edge-orientation}

To recover the true DAG from the learned CPDAG, we address two overarching questions: (1) how to determine the true causal direction of an undirected edge and (2) how to determine the evaluation order of edges. As discussed in Section~\ref{sec:alg-key-idea}, the core idea of SNOE is to identify and orient an undirected edge that, given the current parent sets in the PDAG, satisfies the pairwise additive noise model. In this section, we present how the two key components of SNOE, the edge ranking procedure and edge orientation test, resolve these challenges.

To determine the true causal direction, our method employs a likelihood ratio test, also referred to as the edge orientation test. Given an undirected edge $X-Y$ and the parent sets $\PA_{X}, \PA_{Y}$ in the PDAG, the test compares the bivariate conditional densities $p(X, Y \mid \PA_{X}, \PA_{Y})$ factorized according to the directions $X \rightarrow Y $ and $Y \rightarrow X$. The likelihood ratio test can correctly identify the causal direction when $[X, Y \mid \PA_{X}, \PA_{Y}]$ satisfies the PANM. Furthermore, the test statistic exhibits a desirable asymptotic property that renders the result easy to obtain and interpret.

As previously shown in Figure~\ref{fig:panm-example}, not every pair of nodes connected by an undirected edge satisfies the PANM. A violation of this assumption may cause incorrect conclusions in the orientation test. Thus, we develop an inference procedure to sort the undirected edges. We define a measure to quantify the adherence of an edge to the PANM, which is then utilized to determine edges eligible for orientation at a given stage. At every iteration in our sequential orientation procedure, there exists at least one edge following the PANM (Theorem~\ref{theorem:population-eo}), which is expected to be ranked and evaluated before all other undirected edges. By sequentially orienting the undirected edges in a correct order, the algorithm can ultimately learn the true DAG from the CPDAG.

\subsection{Edge Orientation Algorithm}

The full edge orientation procedure is presented in Algorithm~\ref{alg:OrientEdges}. Lines~\ref{lst:line:rankedges} to~\ref{lst:line:meeksrule} correspond to the procedure detailed in Algorithm~\ref{alg:sneo-population-version}. Before ordering the edges, the algorithm partitions undirected edges $\{U_{k}\}_{k=1}^{m}$ based on the number of neighbors $|\nbr(U_{k})|$ shared between the two nodes on Line~\ref{lst:line:filterbynbr}. The undirected edges are then evaluated in subsets, starting with pairs of nodes sharing $|\nbr(U_{k})|=0$ neighbors. Within each subset, the edges are ranked by an independence measure, which we utilize to determine their adherence to the PANM. This approach allows the algorithm to identify edges eligible for orientation more readily and reduce computation in practice, as nodes with fewer shared neighbors are more likely to satisfy the PANM. We also apply all four of Meek's rules to further orient edges in $\calG$, since the input PDAG may not be the true CPDAG in practice. This assists in reducing the number of undirected edges to sort.

\begin{algorithm}{
    \KwIn{Observed data $X = \{X_{i}\}_{i=1}^{p}$, PDAG $\calG=(V, E)$,  sig. level $\alpha$}
    \KwOut{PDAG $\calG$}
    Let $U=\{U_{1},...,U_{m}\}$ be the set of all undirected edges\;
    Calculate the number of common neighbors $|\nbr(U_{k})|$ in $\calG$ for each edge $U_{k} \in U$\; \label{lst:line:calculate-nbr}
    \For{$i = 0, \ldots,\max\{|\textup{\nbr}(U_{k})|\}$}{
        $\widetilde{U} \gets \{U_{k} \in U: |\nbr(U_{k})| = i \}$\; \label{lst:line:filterbynbr}
        Order edges in $\widetilde{U}$ by the edge-wise independence measure~\eqref{eq:indms}\;\label{lst:line:rankedges} 

        \For{$j = 1, \ldots, |\widetilde{U}|$}{
            $\widetilde{U}_{j}= (a, b) \gets \textit{LikelihoodTest}(\calG, \widetilde{U}_{j}, X, \alpha) $\; \label{lst:line:orientation-procedure-lrt}
            \If{$\widetilde{U}_{j}$ \text{is oriented} }{\label{lst:line:orientation-procedure-post-lrt} 
                Orient $a \rightarrow k \text{ and } b \rightarrow k, \forall k \in \nc_{\calG}(a) \cap \nc_{\calG}(b)$\; \label{lst:line:orientation-procedure-orient-shared-nbr} 
            }
            Apply Meek's rules to $\calG$ and update $U$ accordingly\;\label{lst:line:meeksrule}
        } 
    }
}
\caption{Edge Orientation Procedure \textit{(OrientEdges)} \label{alg:OrientEdges}}
\end{algorithm}

Moreover, we may utilize the undirected components of a PDAG, defined below, to facilitate parallel orientation of undirected edges.

\begin{definition}[Undirected Component in PDAG]
    Let $\calG = (V, E)$ be a PDAG, and $\calG' = (V, E\setminus E_d)$ be the undirected graph obtained after removing all directed edges $E_{d}$ of $\calG$. We call a connected component of $\calG'$ an \textit{undirected component} of $\calG$. 
\end{definition}

The undirected components provide practical significance in the edge orientation procedure. They not only isolate the set of undirected edges from directed edges, but also further partition the undirected edges into disjoint sets. Since the orientation of an undirected edge only affects the structure of its undirected component,  edges in different undirected components can be evaluated and oriented in parallel. An efficient implementation is to apply Algorithm~\ref{alg:OrientEdges} separately to each undirected component of the input PDAG.

\subsection{Ranking Undirected Edges by the PANM Criterion}

As demonstrated in Lemma~\ref{lem:panm-identifiability}, the true orientation of an undirected edge $X-Y$ is identifiable when $[X, Y \mid \pa_{\calG}(X), \pa_{\calG}(Y)]$ follows the pairwise additive noise model. 
Given all the undirected edges, our ranking procedure positions such edges first for orientation by utilizing an independence measure derived from the independent noise property of the PANM. 

Let us employ a pairwise dependence measure $I(X, Y)$ such that $I(X,Y)=0$ if $X \indp Y$ and $I(X,Y)>0$ otherwise.
Let $\widehat{X}$ and $ \widehat{Y}$ denote the regression approximations of $\mathbb{E}[X \mid \pa_{\calG}(X)]$ and $\mathbb{E}[Y \mid \pa_{\calG}(Y), X]$ under $X \rightarrow Y$, where $\calG$ is a PDAG as in Algorithm~\ref{alg:OrientEdges}. Then for each undirected edge $X-Y$, we first calculate the maximum pairwise dependence between parents and residual of a node assuming the orientation $X \rightarrow Y$,
\begin{equation}\label{eq:IXtoY}
    I(X \rightarrow Y) = \max_{Z, W}\{I(X-\widehat{X}, Z), I(Y-\widehat{Y}, W)\}
\end{equation}
over all $Z \in \pa_{\calG}(X)$ and $W \in \pa_{\calG}(Y)\cup\{X\}$. For the opposite orientation, $I(Y\to X)$ is calculated similarly. The maximum pairwise dependence $I(X \rightarrow Y)=0$ if the edge follows a PANM and the true orientation is $X\rightarrow Y$.  
Otherwise, we have $I(X \rightarrow Y) > 0$. 
The edge-wise independence measure for $X-Y$, accounting for both possible directions, is the minimum of the two measures: 
\begin{align}\label{eq:indms}
\widetilde{I}(X,Y) = \min[I(X \rightarrow Y), I(Y \rightarrow X)].
\end{align}
Note that if edge $X-Y$ satisfies the PANM, then $\widetilde{I}(X, Y)=0$. In our work, we use normalized mutual information as the pairwise dependence measure between two random variables $Y_1$ and $Y_2$, 
\begin{align}
I(Y_1, Y_2) = \frac{MI(Y_1, Y_2)}{\min[H(Y_1), H(Y_2)]}, 
\label{eq:normalized-mi}
\end{align}
where $MI(\cdot, \cdot)$ is the mutual information and $H(\cdot)$ is the entropy measure. 
This dependence measure is bounded within $[0, 1]$ and more comparable across different pairs of random variables, as they may have quite different or extreme entropy measures. To simplify computation, we discretize all continuous variables for the calculation of mutual information and entropy. We employ sample splitting on the data to ensure the accuracy of this metric, where the data is split into training and test sets. To calculate the quantity $I(X-\widehat{X}, Z)$ in~\eqref{eq:IXtoY}, for instance, we first fit a (nonlinear) regression model $\widehat{f}(\PA_X)$ for $X$ using training data. Then we obtain the fitted value $\widehat{X}=\widehat{f}(\PA_X)$ from test data. Consequently, the residual $X-\widehat{X}$ and normalized mutual information $I(X-\widehat{X}, Z)$ are both calculated from test data, independent of training data, thus avoiding bias from model overfitting or reuse of the same data.

The purpose of this procedure is to distinguish edges that satisfy the PANM from those that do not. This is achieved simply by calculating $\widetilde{I}(\cdot, \cdot)$ for individual edges and sorting edges in ascending order of $\widetilde{I}(\cdot, \cdot)$. Naturally, this ranking produces an evaluation order for undirected edges, which is different from the common notion of a topological ordering of nodes. 

\begin{figure}
\centering

\begin{tikzpicture}[node distance={15mm}, thick, main/.style = {draw, circle}] 
\node[main] (2) {$X_2$}; 
\node[main] (1) [above of =2] {$X_1$};
\node[main] (3) [below left of=2] {$X_3$}; 
\node[main] (4) [below right of=3] {$X_4$}; 
\node[main] (5) [below left of=4] {$X_5$}; 
\node[main] (6) [below of=4] {$X_6$}; 
\node[main] (7) [below right of=4] {$X_7$}; 
\node[below of=6, node distance=10mm] {(a)};
\draw[->] (1) -- (2); 
\draw[->] (2) -- (3); 
\draw[->] (2) -- (4); 
\draw[->] (3) -- (4); 
\draw[->] (4) -- (5); 
\draw[->] (4) -- (6); 
\draw[->] (4) -- (7); 
\draw[->] (5) -- (6); 
\draw[->] (6) -- (7); 
\end{tikzpicture}
\hspace{0.25cm}%
\begin{tikzpicture}[node distance={15mm}, thick, main/.style = {draw, circle}] 
\node[main] (2) {$X_2$}; 
\node[main] (1) [above of =2] {$X_1$};
\node[main] (3) [below left of=2] {$X_3$}; 
\node[main] (4) [below right of=3] {$X_4$}; 
\node[main] (5) [below left of=4] {$X_5$}; 
\node[main] (6) [below of=4] {$X_6$}; 
\node[main] (7) [below right of=4] {$X_7$}; 
\node[below of=6, node distance=10mm] {(b)};
\draw [color=red, very thick](1) -- (2); 
\draw (2) -- (3); 
\draw (2) -- (4); 
\draw (3) -- (4); 
\draw (4) -- (5); 
\draw (4) -- (6); 
\draw (4) -- (7); 
\draw (5) -- (6); 
\draw (6) -- (7); 
\end{tikzpicture} 
\hspace{0.25cm}%
\begin{tikzpicture}[node distance={15mm}, thick, main/.style = {draw, circle}] 
\node[main] (2) {$X_2$}; 
\node[main] (1) [above of =2] {$X_1$};
\node[main] (3) [below left of=2] {$X_3$}; 
\node[main] (4) [below right of=3] {$X_4$}; 
\node[main] (5) [below left of=4] {$X_5$}; 
\node[main] (6) [below of=4] {$X_6$}; 
\node[main] (7) [below right of=4] {$X_7$}; 
\node[below of=6, node distance=10mm] {(c)};
\draw[->] (1) -- (2); 
\draw[->] (2) -- (3); 
\draw[->] (2) -- (4); 
\draw[color=red, very thick] (3) -- (4); 
\draw[->] (4) -- (5); 
\draw[->] (4) -- (6); 
\draw[->] (4) -- (7); 
\draw[color=red, very thick] (5) -- (6); 
\draw (6) -- (7); 
\end{tikzpicture} 
\hspace{0.25cm}%
\begin{tikzpicture}[node distance={15mm}, thick, main/.style = {draw, circle}] 
\node[main] (2) {$X_2$}; 
\node[main] (1) [above of =2] {$X_1$};
\node[main] (3) [below left of=2] {$X_3$}; 
\node[main] (4) [below right of=3] {$X_4$}; 
\node[main] (5) [below left of=4] {$X_5$}; 
\node[main] (6) [below of=4] {$X_6$}; 
\node[main] (7) [below right of=4] {$X_7$}; 
\node[below of=6, node distance=10mm] {(d)};
\draw[->] (1) -- (2); 
\draw[->] (2) -- (3); 
\draw[->] (2) -- (4); 
\draw[->] (3) -- (4); 
\draw[->] (4) -- (5); 
\draw[->] (4) -- (6); 
\draw[->] (4) -- (7); 
\draw[->] (5) -- (6); 
\draw[color=red, very thick] (6) -- (7); 
\end{tikzpicture} 
\caption{An illustration of the edge orientation procedure. (a) The true DAG. (b) The CPDAG, with $X_{1}-X_{2}$ highlighted to orient first since it satisfies the pairwise ANM. (c) The resulting PDAG after orienting $X_{1} \rightarrow X_{2}$ and employing Meek's rules. Edges $X_{3}-X_{4}$ and $X_{5}-X_{6}$ follow the pairwise ANM and can be oriented. (d) The true DAG is correctly recovered after orienting edge $X_{6}-X_{7}$, which is ranked last due to missing the parent node $X_{5}$ for $X_6$ in (c).}
\label{fig:edge-ordering-example}
\end{figure}

The edge orientation procedure is illustrated through an example in Figure~\ref{fig:edge-ordering-example}. In the CPDAG in Figure~\ref{fig:edge-ordering-example}b, edges $\{X_{1}- X_{2}, X_{2}-X_{3}, X_{4}-X_{5}\}$ follow the PANM and result in $\widetilde{I}(\cdot, \cdot) = 0$. Yet since Algorithm~\ref{alg:OrientEdges} considers pairs of nodes sharing no neighbors first, it first orients $X_{1} - X_{2}$. 
As a result of orienting $X_{1} \rightarrow X_{2}$ and applying Meek's rules, we obtain the maximally oriented PDAG shown in Figure~\ref{fig:edge-ordering-example}c with several more directed edges uncovered. 
Two undirected components, $\{X_{3}, X_{4}\}$ and $\{X_{5}, X_{6}, X_{7}\}$, of the PDAG can now be oriented in parallel.
The algorithm would then find $\widetilde{I}(X_{3}, X_{4}) = 0$ and $\widetilde{I}(X_{5}, X_{6}) = 0$ because both edges satisfy the PANM, and $\widetilde{I}(X_{6}, X_{7}) > 0$ due to $X_{5}$ missing from $\pa_{\calG}(X_{6})=\{X_{4}\}$. Therefore, our method would rank and evaluate $X_3-X_4$ and $X_{5} - X_{6}$ before $X_{6} - X_{7}$. After applying the orientation test and rules again, the algorithm would orient the last undirected edge $X_{6} - X_{7}$ to recover the true DAG, as seen in Figure~\ref{fig:edge-ordering-example}d.

\subsection{Likelihood Ratio Test for Edge Orientation}\label{sec:edge-orientation-lrt}

Our method adopts a comprehensive approach to edge orientation by considering the subgraph formed by both nodes and their learned parent sets. The likelihood ratio test returns a clear decision for edge orientation, whereas the causal relation is difficult to interpret when separate independence tests for opposite edge directions both return statistically significant outcomes \citep{shah2020hardness}. It is also more robust against violations of the model assumptions, e.g. the noise distribution. Given that an undirected edge meets the PANM criterion, the algorithm applies the test to determine its causal direction. We first introduce the formulation of the test statistic and then describe the testing procedure.

Our orientation test takes inspiration from Vuong's test, a series of likelihood ratio tests for model selection and testing non-nested hypotheses \citep{vuong_test}. Consider two nodes $X, Y$ connected by an undirected edge in a PDAG, where $Z_{1} = \PA_{X}$ and $Z_{2} = \PA_{Y}$ have been  identified. As indicated in Lemma~\ref{lem:panm-identifiability}, the PANM is only fulfilled under one causal direction. If the true direction is $X \rightarrow Y$, a reverse causal model $Y \rightarrow X$ would not satisfy the independence relations in Definition~\ref{eq:panm-def} nor adequately fit to the joint distribution. Building on this insight, we compare two sets of conditional models that factorize the joint conditional density $p(x, y \mid z_{1}, z_{2})$ according to the opposing directions: 
\begin{align}
    \text{Under } X \rightarrow Y: F_{\theta^{*}}(x,y\mid z_{1}, z_{2}) &= p(y \mid z_{2}, x; \theta_{1}^{*})p(x \mid z_{1}; \theta_{2}^{*})\label{eq:lrt-x-to-y}, \\
    \text{Under } Y \rightarrow X: G_{\gamma^{*}}(x,y\mid z_{1}, z_{2}) &= p(x \mid z_{1}, y; \gamma_{1}^{*})p(y \mid z_{2}; \gamma_{2}^{*})\label{eq:lrt-y-to-x}.
\end{align}
The two conditional densities are parameterized respectively by $\theta^{*}=(\theta_{1}^{*}, \theta_{2}^{*})$ and $\gamma^{*}=(\gamma_{1}^{*}, \gamma_{2}^{*})$.
We perform two-fold sample splitting on the observed data set, where the training data is used to estimate model parameters $(\widehat{\theta}, \widehat{\gamma})$ and the test data is used to evaluate the log-likelihood. This ensures that $(\widehat{\theta}, \widehat{\gamma})$ are independent of the test data $(X_i, Y_i)$ and allows us to establish the asymptotic distribution of the log-likelihood ratio. 

To determine the edge orientation, we consider three hypotheses in the likelihood-ratio test. The null hypothesis is given as 
\begin{equation}\label{eq:lrt-null-hypothesis}
    H_{0}: \mathbb{E}\left[\log \frac{F_{\theta^{*}}(X, Y\mid Z_{1}, Z_{2})}{G_{\gamma^{*}}{(X, Y\mid Z_{1}, Z_{2})}}\right] \\
    = \mathbb{E}\left[\log \frac{p(Y \mid Z_{2}, X; \theta_{1}^{*}) p(X \mid Z_{1}; \theta_{2}^{*}) }{p(X \mid Z_{1}, Y; \gamma_{1}^{*}) p(Y \mid Z_{2}; \gamma_{2}^{*})}\right] 
    = 0
\end{equation}
and the two alternative hypotheses are formulated as 
\begin{equation}
    H_{f}: \mathbb{E}\left[\log \frac{F_{\theta^{*}}(X, Y\mid Z_{1}, Z_{2})}{G_{\gamma^{*}}{(X,Y\mid Z_{1}, Z_{2})}}\right] > 0 \quad\text{and}\quad
    H_{g}: \mathbb{E}\left[\log \frac{F_{\theta^{*}}(X, Y\mid Z_{1}, Z_{2})}{G_{\gamma^{*}}{(X,Y\mid Z_{1}, Z_{2})}}\right] < 0.
\end{equation}

The test uses the likelihood ratio statistic to select the model closest to the true conditional distribution. The null hypothesis $H_{0}$ indicates that the likelihood is comparable between the two candidate models, hence we cannot identify the causal direction from the observed data. The alternative hypotheses, $H_{f}$ and $H_{g}$, are accepted when a particular edge direction is more probable. The variance of the log-likelihood ratio  with respect to the joint distribution $[X, Y , Z_{1}, Z_{2}]$ is denoted as
\begin{equation}
    \omega_{*}^{2} = \Var \left[\log \frac{F_{\theta^{*}}(X, Y \mid Z_{1}, Z_{2})}{G_{\gamma^{*}}{(X, Y \mid Z_{1}, Z_{2})}} \right].
\end{equation}
When the conditional models are equivalent, i.e. $F_{\theta^{*}} = G_{\gamma^{*}}$, we have $\omega_{*}^{2} = 0$.

We first establish a general result for the log-likelihood ratio with sample splitting. Let $(X,Z)$ denote a generic observation. Suppose two models $F(\cdot\mid Z)$ and $G(\cdot\mid Z)$ are estimated using an independent training sample of size $n$,
yielding estimators $\widehat F(\cdot\mid Z)$ and $\widehat G(\cdot\mid Z)$. Let $\{(X_i,Z_i)\}_{i=1}^n$ be an independent test sample.

 Define the test-sample log-likelihood ratio
\[
R_i
=
\log\frac{\widehat F(X_i\mid Z_i)}{\widehat G(X_i\mid Z_i)},
\qquad i=1,\ldots,n,
\]
and let
\[
\bar R = \frac1n\sum_{i=1}^n R_i,
\qquad
s_R^2 = \frac{1}{n-1}\sum_{i=1}^n (R_i-\bar R)^2.
\]
The population-level parameters are
\[
\mu_0=\E\!\left[\log\frac{F(X\mid Z)}{G(X\mid Z)}\right], \qquad
\sigma_0^2=\Var\!\left[\log\frac{F(X\mid Z)}{G(X\mid Z)}\right].
\]

Let $D(p\,\|\,q):=\E_{(X,Z)}[\log(p(X\mid Z)/q(X\mid Z))]$ for two conditional densities $p(x\mid z)$ and $q(x\mid z)$,
where the expectation is taken with respect to the true joint distribution of $(X,Z)$. If $p$ is the true distribution,
then $D(p\,\|\,q)=\E_Z\left[\text{KL}(p(\cdot\mid Z)\,\|\,q(\cdot\mid Z))\right]$, where $\text{KL}$ denotes the Kullback--Leibler (KL) divergence.

\begin{theorem}
\label{theorem:np_lrt_clt}
Suppose the estimators $\widehat F$ and $\widehat G$ are measurable with respect to the
training sample of size $n$ and independent of the test sample $\{(X_i,Z_i)\}_{i=1}^n$.
Assume the following conditions:
\begin{enumerate}
\item[(i)]
There exists $\eta>0$ such that 
\[
\sup_{n\geq 1}
\E\!\left[
\left|
\log\frac{\widehat F(X\mid Z)}{\widehat G(X\mid Z)}
\right|^{2+\eta}
\right]
<\infty.
\]

\item[(ii)] 
As $n\to\infty$, the conditional variance
\[
v_n=\Var\!\left[
\log\frac{\widehat F(X\mid Z)}{\widehat G(X\mid Z)}
\,\left|\,\widehat F,\widehat G \right.\right]
\overset{p}{\longrightarrow}
\sigma_0^2,
\qquad
0<\sigma_0^2<\infty.
\]

\item[(iii)]
With respect to the distribution of the training sample,
\[
D\bigl(F\,\|\,\widehat F\bigr)=o_p(n^{-b}), \qquad
D\bigl(G\,\|\,\widehat G\bigr)=o_p(n^{-b}), \qquad b\ge 0.
\]
\end{enumerate}
Then, as $n\to\infty$, $\bar R\xrightarrow{p}\mu_0$ if $b=0$ and 
\[
\frac{\sqrt{n}\,(\bar R-\mu_0)}{s_R}\overset{D}{\longrightarrow}N(0,1)
\]
if $b=1/2$.
\end{theorem}

Theorem~\ref{theorem:np_lrt_clt} shows that the log-likelihood ratio converges to the standard normal distribution in the large sample limit, under assumptions (i) to (iii) if the convergence rates in (iii) are $o(n^{-1/2})$. 
Assumption (i) requires the $(2 + \eta)$-th moment of the log-likelihood ratio to be finite such that $\bar{R}$ and $s^2_R$ are well-behaved and converge to their respective true values. Under (ii), the conditional variance stabilizes across training samples. Assumption (iii) states that the loss of the conditional density estimators, measured by $D(\cdot\|\cdot)$, converges at rate $o(n^{-1/2})$. This rate can be achieved by many common nonparametric methods under certain smoothness assumption for the densities. Moreover, the rate can be relaxed to $o(1)$ with $b=0$ if we only need to establish convergence in probability for $\bar{R}$, which is relevant for $H_f$ and $H_g$.
Note that neither $F$ nor $G$ may be the true conditional distribution of $X$ given $Z$. They are usually defined as the minimizer of the expectation of an empirical loss.

Given samples $\{X_{i}, Y_{i}, Z_{1,i}, Z_{2,i}\}_{i=1}^{n}$, we define the likelihood ratio statistic and the estimated variance as
\begin{align}
\begin{split}
    LR_{n}(\widehat{F}, \widehat{G}) &= \sum_{i=1}^{n} \log \frac{\widehat{F}(X_i, Y_i \mid Z_{1, i}, Z_{2, i})}{\widehat{G}{(X_i, Y_i \mid Z_{1, i}, Z_{2, i})}},
\label{eq:likelihood-ratio-stat}
\end{split}
\\[2ex]
\begin{split}
    \widehat{\omega}_{n}^{2} &= \Var \left\{\log \frac{\widehat{F}(X_i, Y_i \mid Z_{1, i}, Z_{2, i})}{\widehat{G}{(X_i, Y_i \mid Z_{1, i}, Z_{2, i})}} \,\left|\, \widehat{F},\widehat{G}\right.\right\}_{i=1}^{n}.
\label{eq:likelihood-ratio-var}
\end{split}
\end{align}

We now establish the following asymptotic result for the likelihood ratio test, as an immediate consequence of Theorem~\ref{theorem:np_lrt_clt}.

\begin{prop} 
Suppose the estimators $\widehat{F}, \widehat{G}$ are independent of $\{(X_i,Y_i, Z_{1, i}, Z_{2, i}),i\in[n]\}$ and satisfy conditions (i) and (ii) in Theorem~\ref{theorem:np_lrt_clt}. Let $\omega_*^2>0$. If $H_0$~\eqref{eq:lrt-null-hypothesis} is true and condition (iii) holds for $b=1/2$, then
    \begin{equation}\label{eq:lrt_null_distr}
    \frac{LR_{n}(\widehat{F}, \widehat{G})}{\sqrt{n}\widehat{\omega}_{n}} \overset{D}{\longrightarrow} N(0, 1), \quad\quad n\to\infty.
    \end{equation}
If $H_{f}$ is true and condition (iii) holds for $b=0$, then
    \begin{equation}
        \frac{LR_{n}(\widehat{F}, \widehat{G})}{\sqrt{n}\widehat{\omega}_{n}} \overset{p}{\longrightarrow} +\infty, \quad\quad n\to\infty.
    \end{equation}
\label{prop:likelihood-ratio-test}
\end{prop}

By symmetry, the above test statistic converges to $-\infty$ under $H_g$.
The test is applicable to $\widehat{F}$ and $\widehat{G}$ obtained by either parametric or nonparametric methods. To interpret the final statistic, the standard decision rule in hypothesis testing applies according to the normal distribution in~\eqref{eq:lrt_null_distr}. For a chosen significance level, 
if the null hypothesis is not rejected, we leave the edge as undirected. Otherwise,
the test has detected a probable causal direction and the edge is oriented accordingly. In contrast to kernel-based tests, the asymptotic property of our likelihood ratio test statistic makes the orientation test computationally tractable, thereby enabling efficient and reliable inference of the causal direction. This test also exhibits an advantage over a score-based approach. Rather than choosing the edge direction with a higher likelihood value to improve the score, the p-value quantifies the statistical significance and uncertainty for the magnitude of the likelihood ratio.

An outline of the edge orientation test is given in Algorithm~\ref{alg:Likelihood Test}. 
When the conditional models $F_{\theta^{*}}$ and $G_{\gamma^{*}}$ are equivalent, conducting the likelihood ratio test is unnecessary as the edge direction is practically indistinguishable. To test for model equivalence, we devise a variance test to assess whether $\omega_{*}^{2} = 0$, as seen on Line~\ref{lst:line:vartest}. The test compares the sample variance $\widehat{\omega}_{n}^{2}$ to 
\[v^{2} \triangleq \min\left\{\Var(\log \widehat{F}_{n}(X, Y \mid \PA_{X}, \PA_{Y})), \Var(\log \widehat{G}_{n}(X, Y \mid \PA_{X}, \PA_{Y}))\right\},\]
which is the smaller variance of the log-likelihood estimates computed under one direction. If $\widehat{\omega}_{n}^{2} / v^{2} < \delta$, for some small threshold $\delta$, then we bypass the likelihood test and declare the edge direction as indistinguishable.

\begin{algorithm}{
    \KwIn{PDAG $\calG$, undirected edge $X-Y$, observed data $\{X, Y, \PA_{\calG}(X), \PA_{\calG}(Y)\}$, sig. level $\alpha$}
    \KwOut{Edge (X,Y) $\in \{X\rightarrow Y, Y \rightarrow X, X-Y\}$}
    Perform train-test split on observed data\;
    Estimate models $\widehat{F}$ and $\widehat{G}$ using GAMs with training data\;   
    Conduct variance test on $\widehat{\omega}_{n}^{2}$~\eqref{eq:likelihood-ratio-var} for model equivalence\;\label{lst:line:vartest}
    \If{$\widehat{\omega}_{n}^{2} / v^{2} > \delta$}{
        Compute the likelihood ratio test statistic $LR_{n}(\widehat{F}, \widehat{G})$~\eqref{eq:likelihood-ratio-stat}\;
        Obtain p-value $p_{E_{X, Y}}$ and preferred edge direction $E_{X, Y}:$
            \[E_{X, Y} = 
            \begin{cases}
              X \rightarrow Y,\ \text{if } LR_{n}(\widehat{F}, \widehat{G}) > 0 \\
              Y \rightarrow X,\ \text{otherwise}
            \end{cases}.\]
        \If{$p_{E_{X, Y}} < \alpha$}{
            Orient edge $(X, Y)$ as $E_{X, Y}$.
        }
    }
}
\caption{Test for Edge Orientation \textit{(LikelihoodTest)} \label{alg:Likelihood Test}}
\end{algorithm}

\begin{remark}\label{remark:alg-step-4}
    While the final output of Algorithm~\ref{alg:sneo-practical-version} is a PDAG, users may specify to return a DAG if they believe that the nonlinear ANM assumption holds. Then, in an additional fourth and final stage, the edge orientation procedure in Algorithm~\ref{alg:OrientEdges} is applied again to extend the PDAG to a DAG. If any undirected edge still remains, the algorithm chooses the orientation with a higher log-likelihood value as the inferred causal direction of the edge.
\end{remark}

\begin{remark}
    Our full algorithm involves constructing regression models in several tasks. This includes estimating residuals for computing $\widetilde{I}(X, Y)$ to rank edges, fitting models to compute the log-likelihood values under possible configurations in subgraphs, and performing covariate testing in the last stage. In the software implementation, the algorithm utilizes generalized additive models from the \textbf{mgcv} package \citep{mgcv_package} to construct regression models, with the thin plate spline selected as the basis function.
\end{remark}

\section{Structural Learning Consistency} \label{sec:theory}

In this section, we establish the correctness of our algorithm in the large-sample limit based on the validity of the sequential orientation procedure at the population level stated in Theorem~\ref{theorem:population-eo}. There are two key elements for demonstrating the consistency of the algorithm. The first key element is to establish the consistency of the nonlinear regression methods. The second is to establish the consistency of the tests utilized in various steps in our algorithm, namely the initial CPDAG learning stage, the ranking of undirected edges for evaluation, and the orientation of undirected edges. 

\begin{remark}
To ease the exposition of technical details, we consider a simplified version of Algorithm~\ref{alg:sneo-practical-version}, in which we do not partition undirected edges based on the number of neighbors in Algorithm~\ref{alg:OrientEdges}. Instead, we simply rank all undirected edges by the independence measure $\widetilde{I}$ after each round of edge orientation. We only consider the initial learning and edge orientation phases, as the edge pruning phase is not needed in the large-sample limit. Moreover, we apply Meek's rules according to Algorithm~\ref{alg:sneo-population-version}. Our consistency results in this section are established for this simplified Algorithm~\ref{alg:sneo-practical-version}.
\end{remark}

First, we define population regression functions and the associated residual variables. The population regression function for $X_{i}$ given subset $S\subseteq [p]\setminus\{i\}$ and the associated residual variable are defined as
\begin{align}
\begin{split} \label{eq:population-regression-form}
    g_{i,S} &:= \argmin_{h \in \mathcal{F}^{\oplus k}}\ \mathbb{E}\left[X_{i}-h(X_S)\right]^2,
\end{split}
\\[2ex]
\begin{split} \label{eq:population-error-form}
    \varepsilon_{i,S}&:=X_i-g_{i,S}(X_S),
\end{split}
\end{align}
where $k=|S|$ and $\mathcal{F}^{\oplus k}$ is the space of additive functions defined in~\eqref{eq:additivefx}. 
Let $\sigma^2_{i,S} := \Var(\varepsilon_{i, S})>0$ be the variance of $\varepsilon_{i, S}$.
 We make the following assumptions on the error variables and the estimation of $g_{i,S}$ and $\sigma^2_{i,S}$.
\begin{assumption}\label{asp:approx-reg-fx}
For all $i\in[p]$ and $S \subseteq [p]\setminus \{i\}$, $\mathbb{E}|\varepsilon_{i,S}|^{4+2\eta} <\infty$ for some $\eta>0$ and the estimators ($\widehat{g}_{i,S}$, $\widehat{\sigma}^2_{i,S}$)
constructed with a sample of size $n$ satisfy:
\begin{align}
\label{eq:gaussian-anm-finite-est-variance}
&c \leq \widehat{\sigma}_{i,S}^{2} \le M \text{ for all } n, \\
\label{eq:gaussian-anm-regression-moment-sup}
&\sup_{n}\E\!\left[\bigl|\widehat{g}_{i, S}(X_{S}))-{g}_{i, S}(X_S)\bigr|^{4+2\eta}\right] <\infty.\\
\label{eq:gaussian-anm-regression-convergence}
&\mathbb{E}[\widehat{g}_{i, S}(X_S) - g_{i, S}(X_S)]^2=o(n^{-1/2}), \\
\label{eq:gaussian-anm-variance-convergence}
&\mathbb{E}(\widehat{\sigma}^2_{i,S}-{\sigma}^2_{i,S})^2=o(n^{-1/2}), 
\end{align}    
where $X_{S} \indp (\widehat{g}_{i,S}, \widehat{\sigma}_{i,S}^{2})$ and $ 0 < c < M < \infty $ are constants.
\end{assumption}

Assumptions \eqref{eq:gaussian-anm-finite-est-variance} and \eqref{eq:gaussian-anm-regression-moment-sup} are mild, only requiring boundedness of $\widehat{\sigma}_{i,S}^{2}$ and a finite moment of the MSE of $\widehat{g}_{i, S}$.
The $L_2$ convergence rates in \eqref{eq:gaussian-anm-regression-convergence} and \eqref{eq:gaussian-anm-variance-convergence} are the  key assumptions, which imply Condition (iii) of Theorem~\ref{theorem:np_lrt_clt} for Gaussian additive regression; see Appendix~\ref{sec:appendix-gaussian-additive-reg-model} for more details.
If each additive function in $g_{i,S}$ belongs to the H\"{o}lder class of smoothness $q$, the classical rate for the MSE
of $\widehat{g}_{i, S}$ is $n^{-2q/(2q+1)}$ \citep{Stone1985}, which is faster than $n^{-1/2}$ for all $q> 1/2$. The standard parametric rate $n^{-1/2}$ of $\widehat{\sigma}^2_{i,S}$ gives an MSE of $O(n^{-1})$.

We now list a few additional assumptions and formally state the consistency of the algorithm.

\begin{assumption}\label{asp:large-sample} Suppose $\{X_{i}\}_{i=1}^{p}$ follows a CAM~\eqref{eq:cam} with a faithful DAG $\calG_{0}$  and satisfies the following assumptions:
    \begin{itemize}
        \item[(A1)] For any $i,j\in[p]$ and any $S\subset [p]$, if $X_i \not\indp X_j \mid X_S$, then the partial correlation $|\rho_{ij \mid {S}}| > \tau$ for some $\tau>0$.
        \item[(A2)] For any $i\in[p]$, if $S \subseteq \pa_{\calG_0}(i)\cup\ch_{\calG_0}(i)$ and $S\cap \ch_{\calG_0}(i) \ne \varnothing$, then the mutual information $MI(\varepsilon_{i,S}, X_k) > \delta$ for some $k\in S$ and some constant $\delta>0$. 
        \item[(A3)] For any $i \in [p]$ and any $S \subseteq \pa_{\calG_0}(i)\cup \ch_{\calG_0}(i)$, the entropy measures $H(X_i),H(\varepsilon_{i, S})\in [c_{1}, c_{2}]$, where $c_{2} > c_{1} > 0$ are constants.
    \end{itemize}
 In (A2) and (A3),  mutual information and entropy measures are calculated through discretization with positive cell probability on every bin.
\end{assumption}

\begin{theorem}
    Let $\widehat{\calG}_n$ be the learned graph of the simplified Algorithm~\ref{alg:sneo-practical-version} applied to an i.i.d. sample of size $n$, in which all involved regression problems are estimated with $(\widehat{g}_{i,S},\widehat{\sigma}^2_{i,S})$ via Gaussian regression. If Assumptions~\ref{asp:approx-reg-fx} and~\ref{asp:large-sample} hold, then 
    \begin{equation}
        \displaystyle{\lim_{n \to \infty}} \mathbb{P}(\widehat{\calG}_n = \calG_{0}) \rightarrow 1
    \end{equation}
    for some choice of $\alpha_{1}, \alpha_{2} \rightarrow 0$.
\label{theorem:finite-sample-eo}
\end{theorem}

This result establishes the consistency of the algorithm in learning the true DAG. Assumption~\ref{asp:approx-reg-fx} is sufficient for the consistency of computing the normalized mutual information using estimated residuals and that of the likelihood ratio test for inferring the causal direction of undirected edges.
The consistency of $\widehat{\calG}_{n}$ also relies on the consistency of the statistical tests performed. Pertinent to skeleton learning in stage 1, Assumption~\ref{asp:large-sample} (A1) states that there exists a lower bound $\tau>0$ for the partial correlation when $X_i \not\indp X_j \mid S_{ij}$. We show that the probabilities of type I and type II errors converge to 0 for the CI tests in the large sample limit, by which our algorithm obtains a consistent CPDAG in stage 1. 
Assumptions~\ref{asp:large-sample} (A2) and (A3) are pertinent to identifying undirected edges that satisfy the PANM criterion. We assume the existence of a gap $\delta>0$ for $MI(\varepsilon_{i,S}, X_k), k \in S$ to precisely distinguish edges that do and do not follow the PANM. 
Last, we assume a mild boundedness assumption on the entropy of each $X_i$ and various residual variables, which guarantees that the normalized independence measure~\eqref{eq:normalized-mi} is well-defined. 

\begin{remark}
The proof of Theorem~\ref{theorem:finite-sample-eo} does not rely on the additive function assumption in \eqref{eq:cam} except for its identifiability. Thus, the structure learning consistency of our algorithm can be readily generalized to the larger class of identifiable ANMs with Gaussian noise.
\end{remark}

We analyze the computational complexity by counting the number of statistical tests performed and regression models fitted in the algorithm. For a $p-$node problem, the learned CPDAG can be a complete graph consisting of $p(p-1)/2$ edges in the worst-case scenario. The CPDAG $\mathcal{E}=(V, E)$ generally contains much fewer edges and its undirected edges $\{U\}_{i=1}^{m}, m < |E|$ generally account for only a fraction of all edges. To compare two causal directions in the edge orientation procedure, our method builds two models for each direction and conducts one test per direction. The edge ranking procedure performs $2m$ tests and fits $4m$ regression models, while the orientation procedure performs at most $|U|=m$ tests and fits $4m$ models. However, there are fewer tests and models required in practice because Meek's rules will orient additional edges. The computational complexity of procedure \textit{OrientEdges} is then of order $O(m)$. In the edge deletion step, the method performs one significance test per node on its covariates, amounting to $p$ tests and $p$ models, and has a complexity of order $O(p)$. While the PC algorithm only conducts conditional independence tests and is exponential with respect to $p$ in the worst case, it becomes polynomial when the underlying DAG is sparse. The empirical runtime comparisons are provided in Section~\ref{sec:experiments-compare-algs}.

\section{Numerical Experiments} \label{sec:experiments-simulated-data}

We conducted numerical experiments with synthetic data to benchmark the accuracy and effectiveness of our method. At a detailed level, we assess the performance of the ranking procedure in Section~\ref{sec:experiments-ordering}, as well as the type I error rate and statistical power of the likelihood ratio test in Section~\ref{sec:experiments-lrtest}. We then compare our method to competing causal discovery algorithms using simulated data sets in Sections~\ref{sec:experiments-compare-algs}. Intermediate results from each stage of our algorithm are provided in Section~\ref{sec:experiments-intermediate-res} to illustrate the effects of the individual components. Two real-world applications are presented in Section~\ref{sec:experiments-real-data}.

We develop two variations of the likelihood ratio test in our algorithm: the \textit{sample-splitting (SNOE-SS)} approach, which is delineated in Algorithm~\ref{alg:Likelihood Test}, and the \textit{cross-validation (SNOE-CV)} approach. The CV approach employs two-fold cross-validation in Algorithm~\ref{alg:Likelihood Test} to perform the likelihood ratio test twice by exchanging the training and test data sets and uses either the smaller or larger p-value for evaluation. The larger p-value is used in our experiments, but practitioners may specify either option. To learn the initial graph, we implemented our modified version of the PC-stable algorithm \citep{pc_stable} from the \textbf{bnlearn} package coupled with the partial correlation test \citep{scutari2010learning}. A more stringent threshold of $\alpha_1 = 0.05$ was applied for learning the CPDAG, while a relaxed threshold of \(\alpha_2 = 0.25\) was used for obtaining the additional candidate edges. The significance level for the likelihood ratio test was set at $\alpha=0.05$ and further reduced to $\alpha = 10^{-4}$ for edge pruning. The algorithm is implemented as an R package and can be accessed at \url{https://github.com/stehuang/snoe.git}.

\subsection{Accuracy of Ranking Procedure} \label{sec:experiments-ordering}

We performed several experiments to verify the precision of our edge ranking procedure. Specifically, we tested its ability to correctly rank undirected edges in a CPDAG. Three distinct DAG structures were considered, as depicted in Figure~\ref{fig:ordering-dags}, with $N=200$ data sets generated for each structure, each containing $n=2000$ samples. As the data was generated using nonlinear functions, the edge directions can be determined under all settings. Given the CPDAG of each DAG, we computed the edge-wise independence measure $\widetilde{I}(X, Y)$, defined in~\eqref{eq:indms}, for each undirected edge $X-Y$ and ranked the edges in ascending order. We assessed whether an edge satisfying the PANM was ranked first. 

\begin{figure}[h]
\centering
\begin{tikzpicture}[node distance={14mm}, thick, main/.style = {draw, circle}] 
\node(temp1) { }; 
\node[main] (A) [below left of =temp1]{$A$}; 
\node[main] (B) [below right of=temp1] {$B$};
\node[main] (Z) [below right of=A] {$Z$};
\node[main] (X) [below of=Z] {$X$};
\node[main] (Y) [below of=X] {$Y$};
\node[below of=Y, node distance = 7mm] {DAG 1};
\draw[->] (A) -- (Z); 
\draw[->] (B) -- (Z); 
\draw[->] (A) to [out=270,in=130, looseness=0.3](X); 
\draw[->] (B) to [out=270,in=50, looseness=0.3](X); 
\draw[->] (A) to [out=230,in=150, looseness=0.3](Y); 
\draw[->] (B) to [out=310,in=30, looseness=0.3](Y); 
\draw[->, color=red] (Z) -- (X);
\draw[->, color=red] (X) -- (Y); 
\end{tikzpicture} 
\hspace{0.2cm}
\begin{tikzpicture}[node distance={16mm}, thick, main/.style = {draw, circle}] 
\node(temp1) { }; 
\node[main] (A) [below left of =temp1]{$A$}; 
\node[main] (B) [below right of=temp1] {$B$};
\node[main] (Z) [below right of=A] {$Z$};
\node[main] (X) [below left of=Z] {$X$};
\node[main] (Y) [below right of=Z] {$Y$};
\node[below of=Z, node distance=30mm] {DAG 2};
\draw[->] (A) -- (Z); 
\draw[->] (B) -- (Z); 
\draw[->] (A) -- (X); 
\draw[->] (A) to [out=220,in=220,looseness=2](Y); 
\draw[->] (B) to [out=320,in=320,looseness=2](X); 
\draw[->] (B) -- (Y); 
\draw[->, color=red] (Z) -- (X);
\draw[->, color=red] (Z) -- (Y); 
\draw[->, color=red] (X) -- (Y); 
\end{tikzpicture} 
\hspace{0.2cm}
\begin{tikzpicture}[node distance={10mm}, thick, main/.style = {draw, circle}] 
\node(temp1) { }; 
\node[main] (A) [below left of =temp1]{$A$}; 
\node[main] (B) [below right of=temp1] {$B$};
\node[main] (Z) [below right of=A] {$Z$};
\node[main] (X) [below of=Z] {$X$};
\node[main] (Y) [below of=X] {$Y$};
\node[main] (W) [below of=Y] {$W$};
\node[main] (V) [below of=W] {$V$};
\node[below of=V, node distance=7mm] {DAG 3};
\draw[->] (A) -- (Z); 
\draw[->] (B) -- (Z); 
\draw[->] (A) -- (X); 
\draw[->] (B) -- (X); 
\draw[->] (A) to [out=270,in=150, looseness=0.2](Y); 
\draw[->] (B) to [out=270,in=30, looseness=0.2](Y); 
\draw[->] (A) to [out=250,in=150, looseness=0.2](W); 
\draw[->] (B) to [out=290,in=30, looseness=0.2](W); 
\draw[->] (A) to [out=230,in=150, looseness=0.2](V); 
\draw[->] (B) to [out=310,in=30, looseness=0.2](V); 
\draw[->, color=red] (Z) -- (X);
\draw[->, color=red] (X) -- (Y); 
\draw[->, color=red] (Y) -- (W); 
\draw[->, color=red] (W) -- (V); 
\end{tikzpicture} 
\caption{Example graphs used for testing the ranking procedure. Red edges indicate the undirected edges in the CPDAG that are evaluated and ranked by the procedure.}
\label{fig:ordering-dags}
\end{figure}

Results are presented in Table~\ref{tab:ordering-res} and are categorized by the data generating function used in the simulations. Note that the cubic, piecewise linear, and sigmoid functions are invertible. The values represent the proportion of data sets in which an undirected edge was ranked first for orientation. Uniformly across all graphs and functions, the edge following the PANM has the highest proportion of being ranked first. It is identified in the vast majority of data sets generated by the quadratic and sigmoid functions. Notably, our ranking procedure successfully determined the edge that satisfies the PANM in most cases under the cubic and piecewise functions, which is more challenging to distinguish since these invertible functions can be well-approximated by a linear function. These experiments verify both the ranking procedure and the use of the edge-wise independence measure to identify an edge fulfilling the orientation criterion.

\begin{table}
\centering
\begin{tabular}{ccccccc}
  \hline
 DAG Structure & Edge & Satisfies PANM & Cubic & Piecewise & Quadratic & Sigmoid \\ 
  \hline \\[-1em]
   1 & $Z - X$ & Yes & 0.64 & 0.62 & 0.99 & 0.72 \\ 
   1 & $X - Y$ & No & 0.36 & 0.38 & 0.01 & 0.28 \\ 
  \hline \\[-1em]
   2 & $Z - X$ & Yes & 0.43 & 0.52 & 0.46 & 0.93 \\ 
   2 & $Z - Y$ & No & 0.34 & 0.25 & 0.35 & 0.07 \\ 
   2 & $X - Y$ & No & 0.23 & 0.23 & 0.19 & 0\\ 
  \hline \\[-1em]
   3 & $Z - X$ & Yes & 0.38 & 0.52 & 0.95 & 0.41 \\ 
   3 & $X - Y$ & No & 0.14 & 0.24 & 0.01 & 0.19 \\ 
   3 & $Y - W$ & No & 0.26 & 0.14 & 0.01 & 0.21 \\ 
   3 & $W - V$ & No & 0.22 & 0.11 & 0.03 & 0.19 \\ 
  \hline
\end{tabular}
\caption{Frequency of an undirected edge ranked first under various settings.}
\label{tab:ordering-res}
\end{table}

\subsection{Evaluation of the Likelihood Ratio Test} \label{sec:experiments-lrtest}

In this subsection, we investigate the type I error and statistical power of the likelihood ratio test. We considered five distinct DAG structures, each comprising of 2 to 5 nodes and 1 to 7 edges, and generated data sets of sample sizes $n=250, 500, 1000, 1500, 2000$. In the CPDAG of each network, we applied the likelihood ratio test to a targeted undirected edge to determine its causal direction. A total of $N=400$ tests were performed per graphical structure and sample size setting. Under a linear Gaussian DAG, the true edge direction of the targeted undirected edge is not identifiable. Under a nonlinear DAG, the true edge direction is identifiable.

A type I error under the likelihood ratio test would be to declare one model more probable than the other when the two models are equivalent. In the context of structural learning, this occurs when an undirected edge is oriented, but should remain undirected. To quantify the type I error rate, we applied the likelihood ratio test to an undirected edge in the CPDAG of a linear, Gaussian DAG and recorded the errors made under significance levels $\alpha=0.01, 0.05$.

\begin{figure}
  \centering
  \includegraphics[scale=0.8]{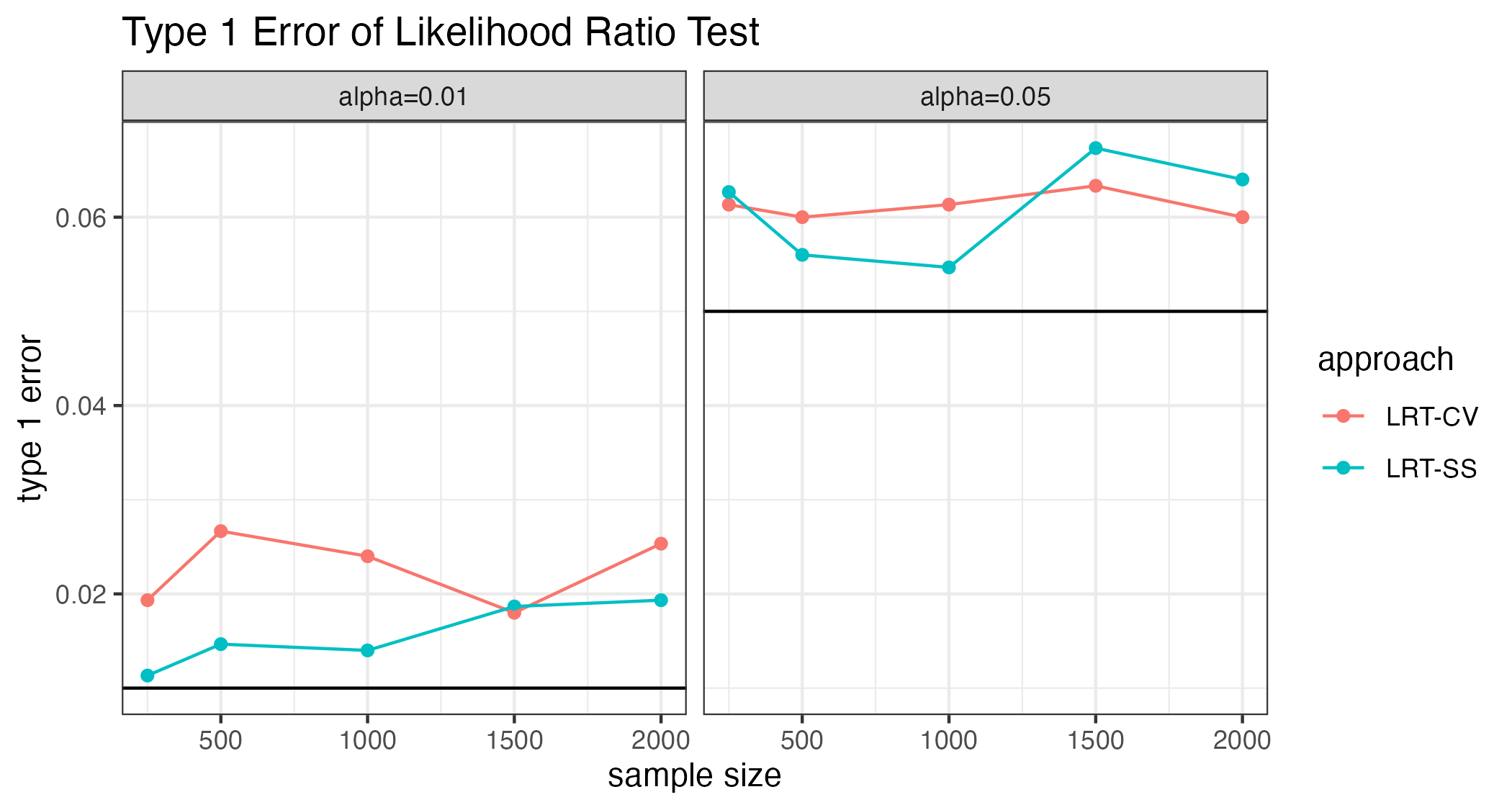}
  \caption{Type I error of likelihood ratio test applied to the targeted edge in various CPDAG structures. The black lines indicate the significance levels.}
\label{fig:lrt-type1-error}
\end{figure}

The type I error of the test, averaged across all DAG structures per sample size, is documented in Figure~\ref{fig:lrt-type1-error}. Overall, the type I error deviates minimally from the specified significance level and stabilizes as the sample size grows. Under both significance levels, the maximal difference between $\alpha$ and the type I error is within 0.015 for both the sample-splitting and cross-validation approaches.  This also signifies that the test is robust against type I errors at smaller sample sizes, which can be attributed to the sample splitting design that renders independence between the estimated model and test data. The difference between the two approaches is minimal as well, differing by 0.005 at $n=2000$, and indicates that both effectively control the false positive rate. 

A type II error under the likelihood ratio test would be to falsely declare the two models as equivalent when only one model is true. In regards to edge orientation, that is to fail at identifying the true causal direction of an edge or incorrectly orient an edge. For this experiment, we applied the likelihood ratio test to data generated from a singular non-linear function, where the orientation of the targeted edge is identifiable. We recorded the power of the test under significance level $\alpha=0.05$.

\begin{figure}
  \centering
  \includegraphics[scale=0.8]{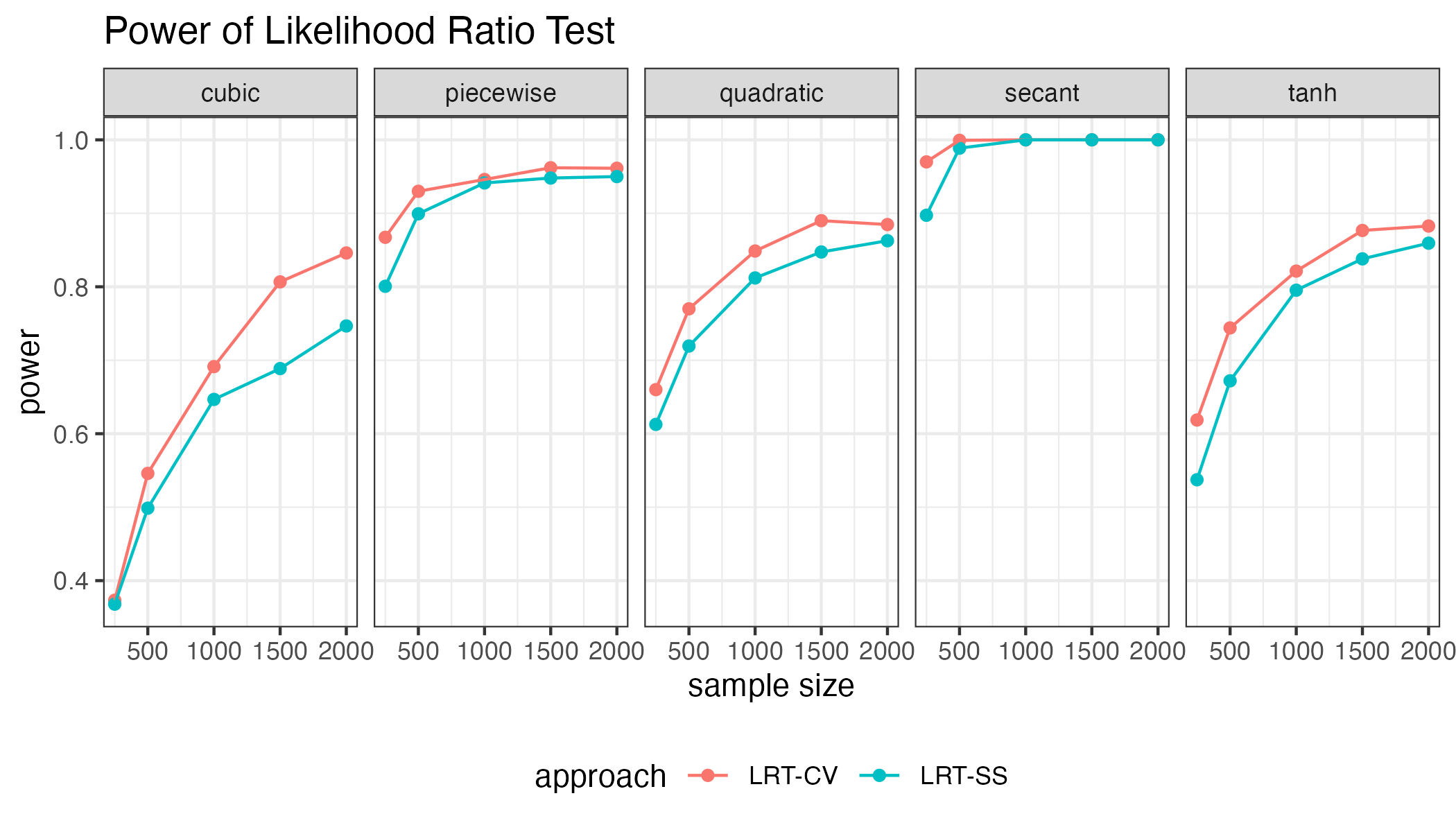}
\caption{The statistical power of the likelihood ratio test on an undirected edge in a CPDAG, with select nonlinear functions underlying the SEM. Results show that power increases as $n$ increases to at least 80\%.}
\label{fig:lrt-power}
\end{figure}

As seen in Figure~\ref{fig:lrt-power}, the statistical power of the likelihood ratio test increases with the sample size across all function types. For certain functions like the piecewise and secant functions, the power approaches one with $n\geq 1500$ samples. For other functions, the power improves more gradually but still strictly increases with the sample size, including the case of the cubic function. In particular, the power increases by at least 33\% between sample sizes $n=250$ and $n=1000$. The cross-validation based approach exhibits greater power, achieving at least 85\% by $n=2000$ across all function types. The results provide empirical evidence that the test can in practice identify the true causal direction of an undirected edge in a CPDAG, especially with sufficient data.

\subsection{Comparison of Algorithm Performances} \label{sec:experiments-compare-algs}

We evaluated the performance of SNOE against competing methods on synthetic data. We compared our method to CAM, NOTEARS, DAGMA, and SCORE, where each represents a different approach for learning nonlinear DAGs. CAM \citep{cam} is a score-based method that assumes a nonlinear causal additive model \eqref{eq:cam} with Gaussian noise and optimizes the log-likelihood function to learn a DAG. Utilizing deep neural networks to model SEMs, NOTEARS \citep{zheng2020learning} and DAGMA \citep{bello2023dagmalearningdagsmmatrices} formulate the structural learning problem as a continuous-optimization problem with an algebraic constraint to enforce acyclicity. In the following experiments, we employed the version tailored to learning from nonlinear data for both algorithms. SCORE \citep{rolland2022score} employs a bottoms-up approach to iteratively identify leaf nodes by computing the Jacobian of the score function under the assumption of a Gaussian error distribution. For all methods, we used their recommended parameter settings. While several constraint-based methods were tested as well, their performance fell short. A detailed analysis of their performances is provided in Appendix~\ref{sec:appendix-constraint-alg}.

The algorithms were applied to learn six DAG structures of varying sizes selected from the \textbf{bnlearn} network repository. For each network, we generated \(N = 75\) data sets, each with a sample size of \(n = 1000\), using the additive model with Gaussian noise in Equation~\ref{eq:cam}. The SEMs were created under three separate functional forms:  linear functions, invertible functions, and non-invertible functions. The function classes are denoted respectively as \textit{linear, inv,} and \textit{ninv} in the figures. Under the invertible functions setting, we randomly selected functions from a set consisting of the cubic, inverse sine, piece-wise linear, and exponential functions. The non-invertible functions were sampled from a Gaussian process using a squared exponential kernel and bandwidth $h \sim \text{Unif}(5, 5.25)$. For all cases, the Gaussian noise term was sampled with mean $\mu = 0$ and standard deviation $\sigma \sim \text{Unif}(0.5, 0.75)$.

The true DAG serves as the ground truth for evaluation, with the exception of the linear, Gaussian case for which only the MEC is identifiable and thus the true CPDAG is used as the ground truth. Our main evaluation metrics are the F1 score, the structural Hamming distance (SHD), and the computational complexity. The F1 score is a harmonic mean of the precision and recall scores. It is calculated as $F1 = \frac{2TP}{2TP+FP+FN+2IO}$,  where \textit{TP}, \textit{FP}, \textit{FN}, and \textit{IO} respectively denote the number of true positives, false positives, false negatives, and incorrectly oriented edges. The SHD measures the number of edge additions, deletions, and reversals required to convert the learned DAG into the true DAG. The computational complexity is measured by overall runtime in seconds.

\begin{figure}
  \centering
  \includegraphics[scale=0.575]{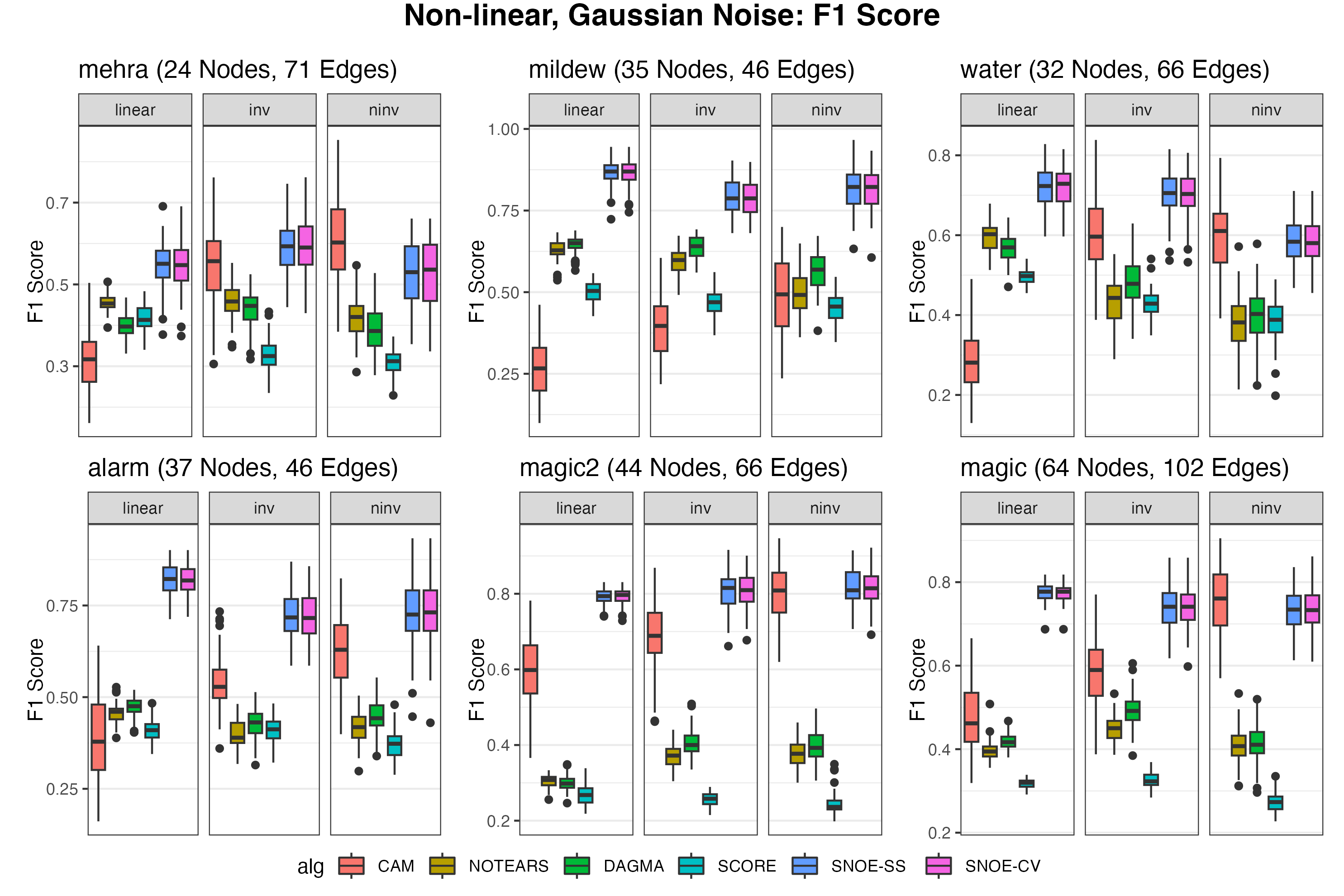}
  \caption{F1 score of learned graphs on simulated data generated under linear, invertible, and non-invertible functions with Gaussian errors.}
\label{fig:all-alg-gaussian-f1}
\end{figure}

The results presented in Figure~\ref{fig:all-alg-gaussian-f1} demonstrate that SNOE consistently outperforms competing methods. We observe that our algorithm achieved uniformly high F1 scores across all network structures and functional forms, with an average standard deviation of 0.06 in its performance across the three types of functions for both approaches. SNOE performed particularly well on invertible nonlinear DAGs, which presents a more challenging task due to the difficulty of detecting such nonlinear relations. 
On average across all function types, the F1 score of SNOE is respectively 67.6\%, 61.8\%, and 100.1\% higher than those of NOTEARS, DAGMA, and SCORE. A closer analysis reveals that NOTEARS and DAGMA produced sparser DAGs by missing considerable amounts of edges, while SCORE often predicted many extraneous edges without capturing the true edges. We also performed hyperparameter tuning on two learning rate parameters of SCORE for a wide range of values, $\{10^{-2}, 10^{-3}, 10^{-4}, 10^{-5}, 10^{-6}\}$, and the resulting F1 scores were very similar for all combinations. CAM performed similarly to our method under data generated from non-invertible functions, but showed considerable variability based on the data-generating function.
Its F1 scores for invertible and linear functions are lower than SNOE with sizable margins for most networks. Specifically, CAM produced denser DAGs with more false positives, especially under these settings.

\begin{figure}
  \centering
  \includegraphics[scale=0.75]{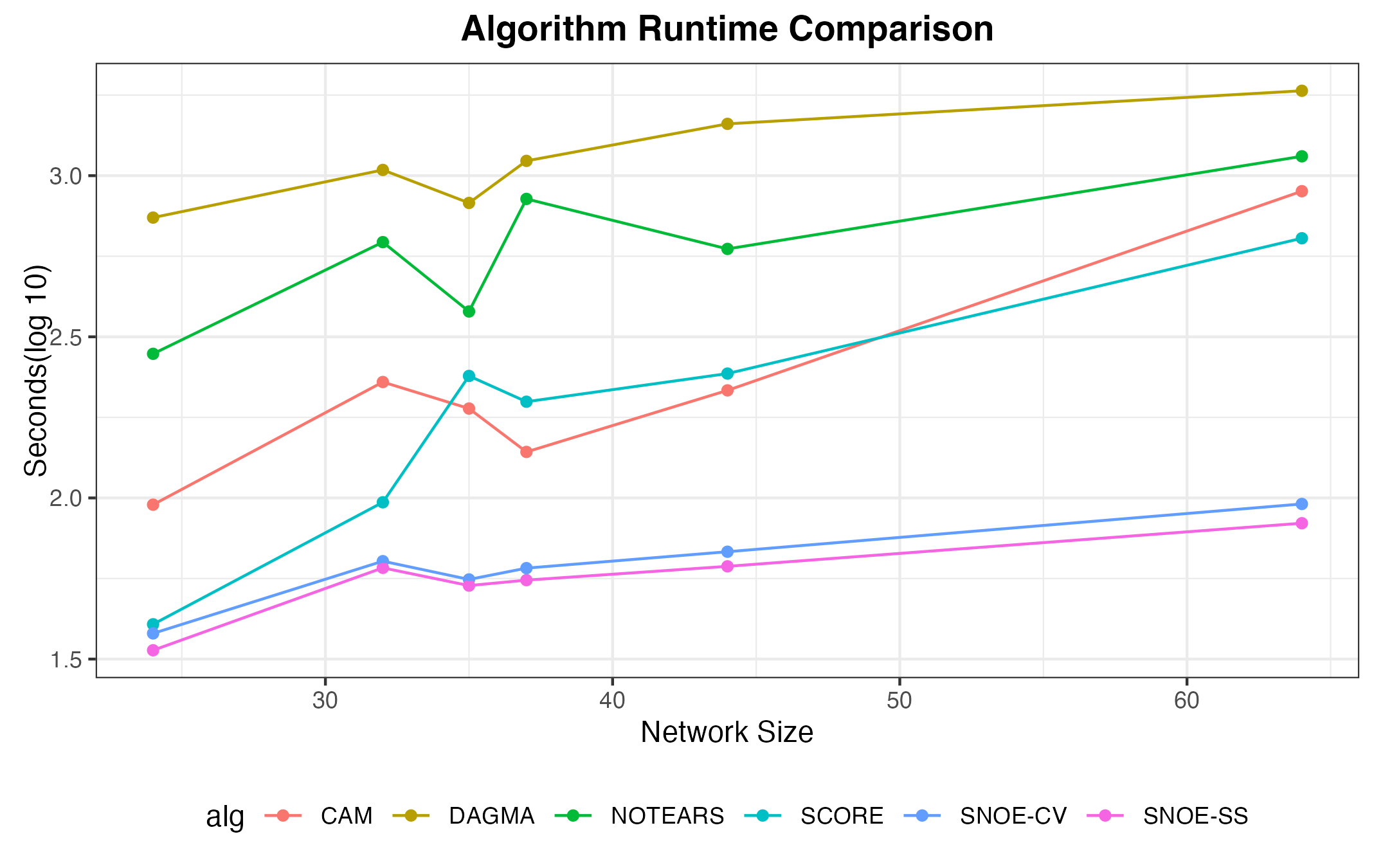}
  \caption{Average runtime in $\log_{10}(\text{seconds})$ of algorithms by network size.}
\label{fig:avg-runtime-comparison}
\end{figure}

Our method also ran significantly faster than the competing methods. We show the average runtime on the $\log_{10}$ scale against network size (number of nodes) in Figure~\ref{fig:avg-runtime-comparison}. The sample-splitting approach (SNOE-SS) has a slightly shorter runtime than the cross-validation approach (SNOE-CV), since the latter performs regression twice. 
For the largest network, our algorithm was at least 7.7 times faster than all competing methods. While CAM showed similar F1 scores in certain cases, SNOE-SS was between 2.8 to 10.7 times faster than CAM, with the difference magnified when learning larger networks. This efficiency can be attributed to its local learning approach of identifying edges satisfying the PANM, rather than optimizing over the entire DAG. Further reduction in runtime occurs as Meek's rules typically orient additional undirected edges after performing the orientation test. In contrast to score-based and optimization-based methods that search over a restricted DAG space, our method leverages properties inherent to the graphical structure and only evaluates a sub-DAG, containing just the relevant nodes, to determine the correct edge orientations. Consequently, this results in higher computational efficiency for structural learning.

\begin{figure}[t]
  \centering
  \includegraphics[scale=0.575]{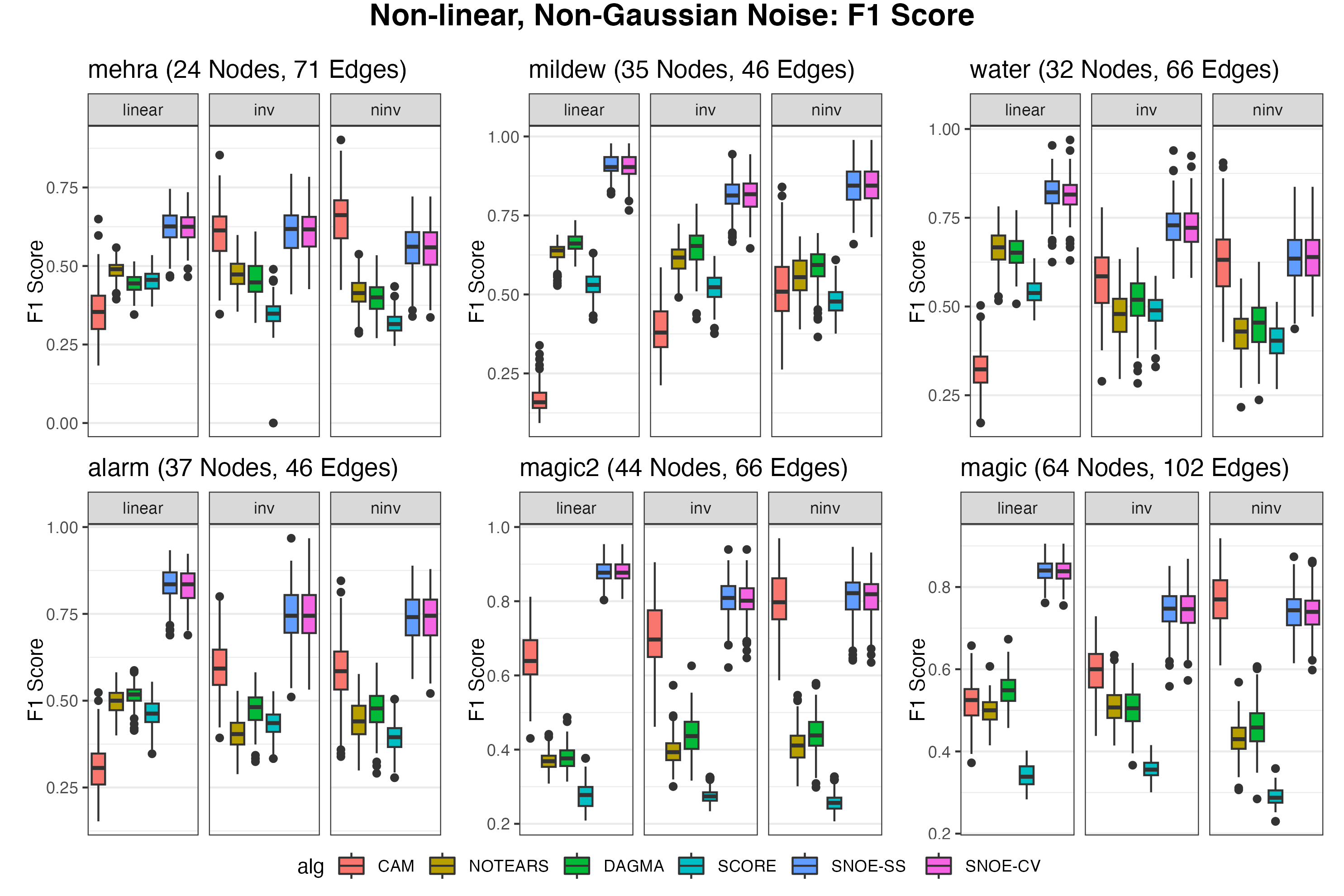}
  \caption{F1 score of learned graphs on simulated data generated under linear, invertible, and non-invertible functions with \textit{non-Gaussian errors}.}
\label{fig:nongaussianres}
\end{figure}

We further investigated the empirical performance of our algorithm under model mis-specification, specifically when the noise distribution is non-Gaussian, and present results in Figure~\ref{fig:nongaussianres}. Recall that the DAG is identifiable when the noise terms follow a non-Gaussian distribution. For this experiment, we simulated data under the same previous settings, but sampled the error variables from three different non-Gaussian distributions: the t-distribution with $df = 5$, the Laplace distribution, and the Gumbel distribution, all with $\mu = 0$ and $\sigma \sim \text{Unif}(0.5, 0.75)$. Since the learning accuracies for each error distribution are similar, we present the combined results. Both variations of SNOE achieved higher accuracy than competing methods across all settings again, while only CAM was comparable in a few cases. 
Similar to the Gaussian case, the F1 score of our method is consistent under different function types with non-Gaussian noise. The exact F1 scores are comparable to the Gaussian case as well, indicating that our method is robust to model mis-specification and is a versatile causal learning method. 

In addition, we examined the structure learning accuracy of our algorithm under data generating models that are not additive, such as  multi-layer perceptrons or nonlinear SEMs including interactions terms. Results in Appendix~\ref{sec:appendix-general-nonlinear-sem-f1} suggest that SNOE is quite robust and still outperformed the competing algorithms.

\subsection{Intermediate Results at Individual Stages} \label{sec:experiments-intermediate-res}

\begin{figure}[h]
  \centering
  \includegraphics[scale=0.575]{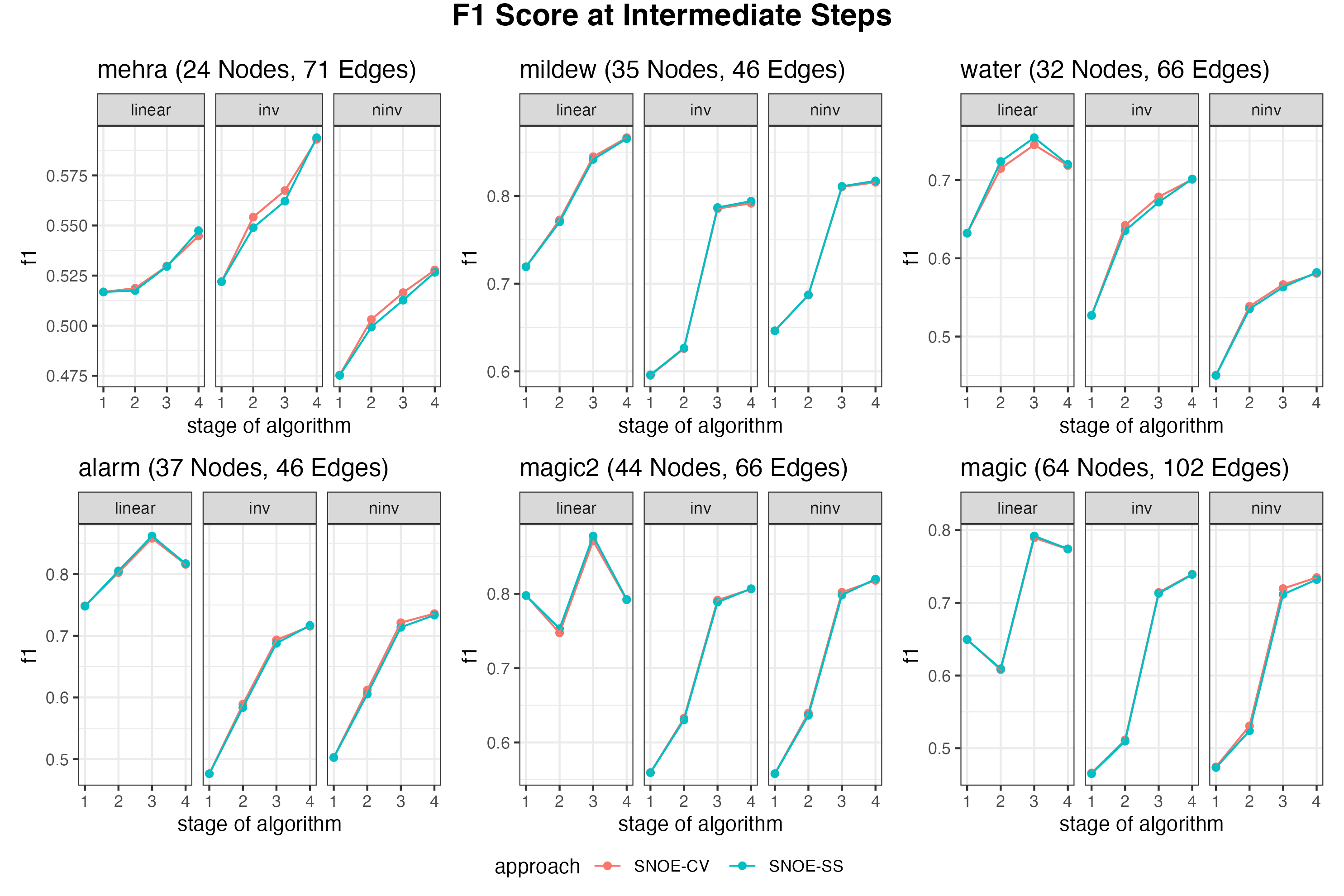}
  \caption{The F1 score after each stage of the framework: (1) initial CPDAG learning, (2) edge orientation, (3) edge deletion, and (4) graph refinement. The two curves overlap substantially in some panels.}
\label{fig:intermediate-f1-score}
\end{figure}

Having presented the performance of our algorithm against competing methods, we now closely analyze its accuracy after each of the following four stages in Algorithm~\ref{alg:sneo-practical-version}: (1)~\textit{initial learning} to learn the initial PDAG, (2)~\textit{edge orientation} to determine the causal direction of undirected edges in the PDAG, (3)~\textit{edge deletion} to remove superfluous edges, and (4)~\textit{graph refinement} to extend the PDAG to a DAG, if applicable, by applying the edge orientation step again. As mentioned in Section~\ref{sec:alg-overview}, while our framework produces a PDAG in general, the implementation incorporates a fourth step to produce a DAG given the non-linear ANM assumption. Figure~\ref{fig:intermediate-f1-score} shows the F1 score computed after each of these four stages. The algorithm's capabilities are best demonstrated in learning nonlinear DAGs, where the F1 score improves by 7.7\% to 53.3\% from the initial step to after the deletion step under nonlinear functions. 
See Appendix~\ref{sec:appendix-intermediate-step-runtime} for analysis and discussion on the runtime of the intermediate steps.

It is imperative to recall that the initial graph is dense because it incorporates an extra candidate edge set ($U_{\alpha_2}$ on Line~\ref{lst:line:getalpha1edges} in Algorithm~\ref{alg:sneo-practical-version}), which may include some false positives. Although the deletion step appears to yield the greatest increase in the F1 score, the edge orientation step actually first uncovers more true positives from the undirected edges (see Figure~\ref{fig:all-alg-gaussian-tp} in Appendix~\ref{sec:appendix-intermediate-step-tp}). A detailed analysis shows that the number of true positives increases by 3.3\% to 23.6\% after applying the orientation procedure. The extra candidate edges may contain true positive edges or edges in the true DAG that are crucial to correctly orienting undirected edges; these edges would otherwise not be recovered in the latter stages. Since the number of true positives remains unchanged after the edge deletion step, we can conclude that the deletion step correctly removes irrelevant edges. Therefore, the inclusion of additional candidate edges is essential and beneficial to our algorithm. Moreover, the significant increase of the F1 score from the initial stage evidences that SNOE can indeed enhance conventional algorithms that learn only an equivalence class.

Since in practice we do not know whether the nonlinear ANM is satisfied, we include the Gaussian linear case to show the performance when the identifiability result does not hold. The ground truth for evaluating this setting is the true CPDAG. In our algorithm design, we use the PC algorithm coupled with the partial correlation test to learn the CPDAG, yielding the relatively high F1 score after the first stage in Figure~\ref{fig:intermediate-f1-score}. For several networks, the F1 score also increases after the second stage of orienting edges. In most practical applications, the estimated CPDAG is not perfect and thus may not capture all v-structures. The observed increase results from recovering missing compelled edges in the true CPDAG by the likelihood ratio test, as confirmed by the increase in true positives after stage two in these networks (see Figure~\ref{fig:all-alg-gaussian-tp} in Appendix~\ref{sec:appendix-intermediate-step-tp}). Since the first stage estimates a dense graph, the third
stage removes false positives and thus improves the F1 score. The inclusion of candidate edges is helpful for the Gaussian linear case as well. Note that the final output is a DAG because we assume the nonlinear ANM; we then see a decrease in the F1 score after stage four which orients all remaining undirected edges. This is expected for Gaussian linear DAGs. Nevertheless, SNOE still exhibits high accuracy in this setting,  demonstrating its versatility for causal learning under various functional forms.

Furthermore, we conducted an experiment to study the effect of misranking edges in the orientation procedure. We evaluated the undirected edges for orientation separately in ranked and arbitrary orders on the same initial graph, and then recorded the F1 score after the orientation stage. The results in Table~\ref{tab:misranking-exp-results}\ in Appendix~\ref{sec:appendix-edge-misranking-exp} confirm the advantages of defining an order in our sequential method over a random search and the robustness against misranking. The analysis reveals that the ranking procedure indeed led to more true positives and higher F1 scores in the orientation stage. When edges are misranked, orienting an edge $X-Y$ that does not fulfill the PANM can lead to an erroneous orientation. However, the likelihood ratio test largely mitigates the impact of misranking since it may find models $X \rightarrow Y$ and $Y \rightarrow X$ indistinguishable or comparably fitted, thus leaving the edge undirected until evaluation at a later iteration. See Appendix~\ref{sec:appendix-edge-misranking-exp} for a more detailed discussion.

We have also examined the impact of the initial graph on the performance of our algorithm (Appendix~\ref{sec:appendix-initial-graph-exp}). The results show that our edge orientation procedure can consistently recover the true DAG from estimated CPDAGs with varying levels of accuracy. In Appendix~\ref{sec:appendix-lrt-downstream-error}, we provide insights on the impact of early errors during the orientation procedure; the experimental results suggest that the impact is minimal.

\section{Real Data Applications} \label{sec:experiments-real-data}

\subsection{Flow-Cytometry Data} \label{sec:experiments-sachs}

We applied all methods to the flow cytometry data set underlying the well-known Sachs network \citep{flow_cytometry}. The data set was collected in a study aiming to infer causal pathways amongst 11 phosphorylated proteins and phospholipids by measuring their expression levels after performing knockouts and spikings. Through a meta-analysis of both biological data and published literature, the researchers constructed a causal DAG that illustrates 17 causal relations among the 11 molecules. Given the high potential and broad applicability of causal learning in biological sciences, the Sachs network is one of the few verified causal graphs and is a popular means to benchmark causal learning methods. While the original data contains 7466 single-cell samples, we applied the algorithms to only the continuous version of the observational data set, which reduces the final data set to 2603 samples. It should be noted that the underlying skeleton is not fully connected; the graph consists of two disjoint clusters, one containing 8 nodes and the other containing 3. 

\begin{table}[h]
\centering
\begin{tabular}{lccccccc}
  \hline
 Algorithm & Edges & SHD & F1 & TP & FP & FN & Wrong Direction \\ 
  \hline
    CAM &  19 & 19 & 0.39 &   7 &   9 &   7 &   3 \\ 
    NOTEARS &  8 & 13 & 0.40 &   5 &   1 &  10 &   2 \\ 
    DAGMA &  6 & 15 & 0.26 &   3 &   1 &  12 &   2 \\ 
    SCORE &  17 & 20 & 0.29 &   5 &   8 &  8 &   4 \\ 
    \textbf{SNOE-CV} &  \textbf{10} & \textbf{12} & \textbf{0.52} & \textbf{7} & \textbf{2} &   \textbf{9} & \textbf{1} \\
    SNOE-SS &  11 & 13 & 0.50 &   7 &   3 &   9 &   1 \\ 
   \hline
\end{tabular}
\caption{Algorithm performance on flow-cytometry data.}
\label{tab:sachsres}
\end{table}

The performance summary of the estimated graphs in Table~\ref{tab:sachsres} shows the SNOE cross-validation approach produces a DAG closest to the ground truth, with a F1 score more than 30\% higher than those of competing methods. Although its learned DAG is sparser than the true DAG, SNOE-CV has the highest F1 score and ties with SNOE-SS and CAM for the number of true positives captured. Despite missing several edges, it predicted very few false positives and thus has a lower SHD. The sample-splitting approach ranks second and differs from the cross-validation approach by predicting just one more false positive. NOTEARS and DAGMA were applied to both the original and standardized data, with the better results reported. Nevertheless, both methods still suffer from relatively high numbers of false negatives. Although CAM and SCORE have the densest DAGs predicted, they also have the highest counts of false positives and SHD.

\subsection{Tübingen Cause-Effect Pairs} \label{sec:experiments-cause-effect}

The Cause-Effect database is a collection of 108 different cause-effect pairs sourced from various domains such as biology, climate science, economics, and sociology \citep{mooij2016distinguishing}. Each data set consists of two variables with a known causal relation. The database has emerged as a popular benchmark for evaluating bivariate causal discovery methods given its diversity in data sources and data types. In this experiment, we applied both variants of the likelihood ratio test, the sample-splitting (SS) and cross-validation (CV) approaches, to each pair using a significance level of $\alpha=0.05$. We consider 98 data sets since nine of them contain multi-dimensional variables.

\begin{table}[h]
\centering
\begin{tabular}{c p{4cm} p{4cm} p{4cm} p{4cm}}
\hline
Approach & \# of undetermined cases & Accuracy when causal direction is determined & Overall accuracy \\
\hline
LRT-SS & 63 & 68.6\% & 54.1\% \\
LRT-CV & 66 & 71.9\% & 66.3\% \\
\hline
\end{tabular}
\caption{Likelihood ratio test results on 98 cause-effect data sets.}
\label{tab:cause-effect-predictions}
\end{table}

The first two columns of Table~\ref{tab:cause-effect-predictions} show the results of applying the likelihood ratio test with $\alpha=0.05$ to determine the edge orientation. As shown in the first column, for more than 60 data sets, the causal direction cannot be determined by the likelihood ratio test at the significance level of 0.05. The causal models for $X \rightarrow Y$ and for $Y \rightarrow X$ were found to be equivalent for these data sets, suggesting that the SEMs may be approximated by linear models. Further analysis reveals that the average correlation between these undetermined pairs is 0.49, which signals a strong linear relation. 
For the other data sets, both approaches achieved a high accuracy in inferring the correct causal direction for about $70\%$ of the data sets.

\citet{mooij2016distinguishing} applied their causal discovery methods to these data sets, assuming nonlinear ANMs as well. 
Their methods predicted the causal relations for all 98 data sets by choosing the direction with less dependence between the residual and parent variables. To compare with their methods, we chose the direction having a larger log-likelihood value as the predicted causal direction for each data set, regardless of the test significance.
We focus on their results for six entropy-based approaches, since the entropy measures are closely related to the normalized MI used in our work. Only one of their six approaches reached an accuracy of around $70\%$ and the remaining all scored 40\%--60\%. 
As reported in the last column of Table~\ref{tab:cause-effect-predictions}, while LRT-SS exhibits a similar overall accuracy of 54.1\%, LRT-CV outperforms most of their approaches with an overall accuracy of 66.3\%. \citet{mooij2016distinguishing} attribute the large variation in accuracy to discretization effects when calculating the differential entropy. In our procedure, we normalize mutual information to avoid extreme values arising from distribution skewness or discretization. Furthermore, the MI measure is only used to identify edges for orientation by checking adherence to the PANM criterion, while the likelihood ratio test determines the causal relation in a robust way.

\section{Discussion} \label{sec:conclusion}

In this work, we present a novel algorithm for learning nonlinear causal DAGs through a sequential edge orientation framework. Specifically, we demonstrate that the edge orientation algorithm can learn the true DAG from the CPDAG by sequentially orienting the undirected edges. The framework is established on the pairwise additive noise model, a criterion we developed to ensure accurate inference of the causal direction for undirected edges from just the two nodes and their identified parent sets. The sequential orientation of edges is achieved through two key components: the likelihood ratio test, which provides a definitive decision on the causal direction of an undirected edge, and the edge ranking procedure, which recursively determines edges that follow the PANM to ensure the correctness of orientations made. These two procedures effectively address two fundamental questions for constraint-based approaches regarding how to determine the causal direction of edges and how to order edges for evaluation. We also propose two approaches to the likelihood ratio test, both of which demonstrate well-controlled type I error and high statistical power. SNOE provides a practical, yet still precise, alternative to kernel-based and regression-based learning of nonlinear causal relations. Compared to competing methods, SNOE exhibits robustness and high precision, which can be attributed to its reduced dependence on model assumptions. It also requires far less computational time and demonstrates strong generalizability to different data functions and distributions.

Potential extensions of this framework include learning nonlinear causal relations in the presence of hidden variables — a common challenge in constraint-based algorithms. The key is to adapt the edge ranking and the likelihood ratio test to take into account latent confounders.
In addition, this work can be refined and expanded in several ways. 
One direction is to expand our algorithm to the post-nonlinear (PNL) model, where the SEM is given as $X_{i} = g_{i}(f_{i}(\PA_{i}) + \varepsilon_{i})$ \ and the causal direction can be determined by testing the independence between $\varepsilon_{i}$ and $\PA_{i}$ \citep{zhang2009pnl_identifiability}. We can adapt the pairwise ANM criterion to the PNL model, as it also relies on the independence noise property for identification, and thereby produce an evaluation order of edges in learning the true DAG from its MEC. Deep learning methods \citep{uemura2020estimation_pnl} have been developed to better approximate $f_{i},g_{i}$ and the noise term, which would allow us to accurately compute the normalized MI, rank edges, and perform the likelihood ratio test under the PNL model.
Another promising direction is to integrate alternative independence measures and estimation methods into our algorithm. For general nonlinear models, a kernel-based test \citep{kci_test} may be more accurate and an alternative likelihood estimation, such as the method proposed in \citet{khemakhem2021causal}, can be used in place of the likelihood ratio test to determine the causal direction.

\acks{We thank Dr. Bingling Wang for helpful discussions. This work was supported by NSF grant DMS-2305631 and NIH grant R01GM163245. It used computational resources and storage capacities on the Hoffman2 Shared Cluster hosted by the UCLA Institute for Digital Research and Education’s Research Technology Group.}

\newpage

\appendix

\section{Restricted Additive Noise Model} \label{appendix:restricted-anm-def}

Suppose each variable $X_{j}$ is generated by an ANM $X_{j} = f_{j}(\PA_{j}) + N_{j}$, where $f_j$ is an arbitrary function of parent variables $\PA_{j}$ and $N_j$ is an additive noise. Moreover, let $ND_{j}$ denote the non-descendants of $X_j$ in the underlying DAG.
We denote by $L(Y)$ the distribution of a random variable $Y$.
In a bivariate additive noise model for variables $X_{i}, X_{j}$, \citet{hoyer2008nonlinear} have proven that if the triple $(f_{j}, L(X_{i}), L(N_{j}))$ satisfies the following condition, then the causal relation is identifiable. 

\begin{condition}\label{cond:anm-fx-ident}
Let $p_{X_i}$ and $p_{N_j}$ be strictly positive densities of $L(X_i)$ and $L(N_j)$, respectively.
    The triple $(f_j,L(X_i), L(N_j))$ does not solve the following differential equation for all $x_i, x_j$ with $\nu''(x_j - f(x_i))f'(x_i) \ne 0$:
    \begin{equation}
        \xi''' = \xi'' \left( -\frac{\nu'''}{\nu''}f' + \frac{f''}{f'} \right) - 2 \nu'' f'' f' + \nu' f''' + \frac{\nu' \nu''' f'' f'}{\nu''} - \frac{\nu'(f'')^2}{f'}, 
    \end{equation}
    where $f := f_j$,  $\xi := \log p_{X_i}$, and $\nu := \log p_{N_j}$. To improve readability, we have omitted the arguments $x_j - f(x_i), x_i$, and $x_i$ for $\nu, \xi$, and $f$ and their derivatives, respectively.
\label{cond:bivar-anm}
\end{condition}

\citet{resit_anm} then utilize this result to prove the identifiability of a DAG assuming a restricted additive noise model. 

\begin{definition}
We call the SEM~\eqref{eq:anm} a restricted additive noise model if for all $j \in V=[p]$, $i \in \PA_j$ and all sets $S \subseteq V$ with $\PA_j \setminus \{i\} \subseteq S \subseteq ND_j \setminus \{i,j\}$, there is an $x_S$ with $p_{S}(x_S) > 0$, s.t.
\[
\left( f_j(x_{\PA_j \setminus \{i\}}, \underbrace{\cdot}_{X_i}), L(X_i \mid X_{S} = x_S), L(N_j) \right)
\]
satisfies Condition~\ref{cond:bivar-anm}. In particular, we require the noise variables to have non-vanishing densities and the functions $f_j$ to be continuous and three times continuously differentiable.
\label{eq:restricted-anm}
\end{definition}

Under causal minimality, if $L(X)=L(X_{1}, \ldots, X_{p})$ is generated by a restricted ANM with DAG $\calG_{0}$, then $\calG_{0}$ is identifiable from the joint distribution.

\section{Likelihood Ratio under Gaussian Regression Models} \label{sec:appendix-gaussian-additive-reg-model}

In this section, we verify conditions (i) - (iii) of Theorem~\ref{theorem:np_lrt_clt} under the Gaussian Regression model:
\begin{equation}
X = m(Z) + \varepsilon,\qquad \varepsilon \sim N(0,\sigma^2), 
\label{eq:gaussian-regression-model}
\end{equation}
where $Z=(Z_1,\dots,Z_d)$, the noise $\varepsilon$ is independent of $Z$, and $0 < \sigma^{2} < \infty$.
Suppose we compare two Gaussian regression models, $F(X \mid Z)$ and $G(X \mid Z)$. Without loss of generality, assume $F:X=m_{F}(Z)+\varepsilon_{F}$ is the true model, while $G:X=m_{G}(Z)+\varepsilon_{G}$ is not necessarily true. In particular, properties of the error variable, $\varepsilon_{F} \sim N(0, \sigma_{F}^{2})$ and $\varepsilon_{F} \indp Z$, hold under model $F$, whereas they may not hold for 
$\varepsilon_{G}$ if $m_G\ne m_F$.

Under the Gaussian regression model~\eqref{eq:gaussian-regression-model}, the conditional densities $F(\cdot\mid Z)$ and $G(\cdot\mid Z)$ are of
the form
\[
F(x\mid z)
=
\frac{1}{\sqrt{2\pi\sigma_F^2}}
\exp\!\left\{-\frac{(x-m_F(z))^2}{2\sigma_F^2}\right\},
\qquad
G(x\mid z)
=
\frac{1}{\sqrt{2\pi\sigma_G^2}}
\exp\!\left\{-\frac{(x-m_G(z))^2}{2\sigma_G^2}\right\},
\]
where the noise variances $\sigma_F^2,\sigma_G^2$ are finite and positive.
Let $\widehat m_F,\widehat m_G$ and $\widehat\sigma_F^2,\widehat\sigma_G^2$
be estimators constructed from a training sample of size $n$. Denote by $\widehat F$ (and $\widehat G$)
the corresponding estimator of the conditional density $F$ (and $G$) after plugging in 
$\widehat m_F,\widehat\sigma_F^2$ (and $\widehat m_G,\widehat\sigma_G^2$).
Let $(X,Z)$ be a test sample independent of the training sample. 

\subsection{Verification of Condition (i)}

We start with 
\begin{align*}
\log\frac{\widehat F(X\mid Z)}{\widehat G(X\mid Z)}
&=
\frac12\log\frac{\widehat\sigma_G^2}{\widehat\sigma_F^2}
-\frac{(X-\widehat m_F(Z))^2}{2\widehat\sigma_F^2}
+\frac{(X-\widehat m_G(Z))^2}{2\widehat\sigma_G^2}.
\end{align*}
For any $r\ge 1$, there exists a constant $C_1<\infty$ such that
\begin{align} 
\left|
\log\frac{\widehat F(X\mid Z)}{\widehat G(X\mid Z)}
\right|^r
\;\le\;
C_1\Biggl[1+
\frac{(X-\widehat m_F(Z))^{2r}}{(\widehat\sigma_F^2)^{r}}+
\frac{(X-\widehat m_G(Z))^{2r}}{(\widehat\sigma_G^2)^{r}}
\Bigg],
\label{eq:llr_bound}
\end{align}
assuming $\widehat{\sigma}_{F}^{2}, \widehat{\sigma}_{G}^{2}$ are finite and bounded away from 0.
Note that we can decompose the regression error $X-\widehat{m}_{G}(Z)$ as 
\[
X-\widehat{m}_{G}(Z)  = \varepsilon_{G} + (m_{G}(Z)-\widehat{m}_{G}(Z))
\]
to obtain a bound
\[
|X - m_{G}(Z)|^{2r} \le C_{2}\left[|\varepsilon_{G}|^{2r} + |m_{G}(Z)-\widehat{m}_{G}(Z)|^{2r}\right],
\]
where $C_{2} > 0$, and a similar bound for $|X - m_{F}(Z)|^{2r}$.
Let $r = 2 + \eta$ for some $\eta>0$ and  assume the following conditions:
\begin{enumerate} \label{asp:gaussian-model-condition1-assumptions}
\item[(i-1)]
The error variables $\varepsilon_{F}$ and $\varepsilon_{G}$ have finite $(4+2\eta)$-th moments. Note that the Gaussian error $\varepsilon_{F}$ satisfies this automatically.

\item[(i-2)]
The regression estimators satisfy the uniform higher-order moment bound
\[
\sup_{n}
\E\!\left[
\bigl|\widehat m_F(Z)-m_F(Z)\bigr|^{4+2\eta}
+
\bigl|\widehat m_G(Z)-m_G(Z)\bigr|^{4+2\eta}
\right]
<\infty.
\]

\item[(i-3)]
For all $n$, the variance estimators $\widehat{\sigma}_{F}^{2}, \widehat{\sigma}_{G}^{2} \in[c, M] $ for some $0 < c < M < \infty$.

\end{enumerate}
Then, the expectation of the right-hand side of \eqref{eq:llr_bound} is bounded uniformly in $n$. Consequently,
\[
\sup_n
\E\!\left[
\left|
\log\frac{\widehat F(X\mid Z)}{\widehat G(X\mid Z)}
\right|^{2+\eta}
\right]
<\infty,
\]
which verifies Condition (i) of Theorem~\ref{theorem:np_lrt_clt}. We summarize this result as a lemma:

\begin{lemma}\label{lemma:gaussian-anm-cond1}
    Under assumptions (i-1) to (i-3) in Appendix \ref{asp:gaussian-model-condition1-assumptions}, Condition (i) of Theorem~\ref{theorem:np_lrt_clt} holds for the Gaussian regression models $F$ and $G$.
\end{lemma}

\begin{remark}
Assumptions (i-1) and (i-2) only require finite $(4+2\eta)$-th moments for $\varepsilon_G$ and for the error of the regression function estimators.
Assumption (i-3) is satisfied if the variance estimators are finite and bounded away from 0. These are all mild and easily satisfied
for Gaussian regression models.
\end{remark}

\subsection{Verification of Condition (ii)}

Define the population and estimated log-likelihood ratios as

\begin{align*}
R
&=\log\frac{F(X\mid Z)}{G(X\mid Z)} =
\frac12\log\frac{\widehat\sigma_G^2}{\widehat\sigma_F^2}
-\frac{(X-\widehat m_F(Z))^2}{2\widehat\sigma_F^2}
+\frac{(X-\widehat m_G(Z))^2}{2\widehat\sigma_G^2},
\\
R_0
&= \log\frac{\widehat F(X\mid Z)}{\widehat G(X\mid Z)} =
\frac12\log\frac{\sigma_G^2}{\sigma_F^2}
-\frac{(X-m_F(Z))^2}{2\sigma_F^2}
+\frac{(X-m_G(Z))^2}{2\sigma_G^2}.
\end{align*}
We show that the following assumptions are sufficient for condition (ii):
\begin{enumerate} \label{asp:gaussian-model-condition2-assumptions}
    \item[(ii-1)] $\E(\varepsilon_{F}^{4}), \E(\varepsilon_{G}^{4}) < \infty$,
    \item[(ii-2)] $\widehat m_F(Z)\xrightarrow{L^2} m_F(Z), \widehat m_G(Z)\xrightarrow{L^2} m_G(Z)$,
    \item[(ii-3)]$\widehat\sigma_F^2\xrightarrow{L^2}\sigma_F^2, \widehat\sigma_G^2\xrightarrow{L^2}\sigma_G^2$.
\end{enumerate}
Assumptions (ii-2) and (ii-3) imply that each term in $R-R_0$ converges to zero in $L^2$, and consequently,
\begin{equation}
\E\bigl[(R-R_0)^2\bigr]\to 0.
\label{eq:l2_convergence_llr}
\end{equation}
Since $\varepsilon_{F}$ is Gaussian and $\mathbb{E}(\varepsilon_{G}^{4}) < \infty$ by (ii-1),
$
\sigma_0^2 := \Var(R_0) < \infty,
$
and \eqref{eq:l2_convergence_llr} implies
$\E[R]\to\E[R_0]$ and $\E[R^2]\to\E[R_0^2]$,
it follows that $\Var(R)\to \Var(R_0)=\sigma_0^2$.
Let
\[
U_n^2 := \E\!\left[(R-R_0)^2 \mid \widehat F,\widehat G\right].
\]
By the tower property, $\E[U_n^2]=\E\!\left[(R-R_0)^2\right] \to 0$,
which implies
\begin{equation}\label{eq:conditional_L2}
\E\!\left[(R-R_0)^2 \mid \widehat F,\widehat G\right] = U_{n}^{2} 
\overset{p}\longrightarrow 0.
\end{equation}
Consider the decomposition
\begin{align*}
\Var(R\mid \widehat F,\widehat G)
&=
\Var(R_0\mid \widehat F,\widehat G)
+
\Var(R-R_0\mid \widehat F,\widehat G)
+
2\,\Cov(R_0, R-R_0\mid \widehat F,\widehat G),
\end{align*}
where $\Var(R_0\mid \widehat F,\widehat G)=\sigma_0^2$ as $R_0$ does not depend on $(\widehat F,\widehat G)$. 
Note that 
\[\Var(R-R_{0} \mid \widehat{F}, \widehat{G}) \leq \mathbb{E}[(R-R_{0})^{2} \mid \widehat{F}, \widehat{G}]\] 
and by the Cauchy--Schwarz inequality,
\[
\left|
\Cov(R_0,R-R_0\mid \widehat F,\widehat G)
\right|
\le
\sqrt{\Var(R_0)}\,
\sqrt{\E[(R-R_0)^2\mid \widehat F,\widehat G]}.
\]
Combining these bounds with \eqref{eq:conditional_L2}, we obtain
$\Var(R\mid \widehat F,\widehat G)\xrightarrow{p}\sigma_0^2$,
which establishes Condition (ii). The result is summarized as a lemma:

\begin{lemma}\label{lemma:gaussian-anm-cond2}
    Under assumptions (ii-1) to (ii-3) in Appendix~\ref{asp:gaussian-model-condition2-assumptions}, Condition (ii) of Theorem~\ref{theorem:np_lrt_clt} holds for the Gaussian regression models $F$ and $G$.
\end{lemma}

\begin{remark}
In Gaussian additive models, Condition (ii) is implied by the $L^{2}$ consistency of the
regression function estimators and the variance estimators, together with a mild 4-th moment condition for the error distributions. These conditions are standard
for spline or series estimators under sample splitting.
\end{remark}

\subsection{Verification of Condition (iii)}

Since $F$ and $G$ both define a Gaussian distribution for $[X\mid Z]$, we calculate $D(p\|\widehat{p})$ between 
\[
p(x\mid z)=\frac{1}{\sqrt{2\pi\sigma^2}}
\exp\!\left\{-\frac{(x-m(z))^2}{2\sigma^2}\right\}
\quad\text{and}\quad
\widehat p(x\mid z)=\frac{1}{\sqrt{2\pi\widehat\sigma^2}}
\exp\!\left\{-\frac{(x-\widehat m(z))^2}{2\widehat\sigma^2}\right\}.
\]
Let $\varepsilon:=X-m(Z)$. Given ($\widehat{m},\widehat{\sigma}^2$), taking expectation with respect to $(X,Z)$, we have
\begin{align*} 
    D(p\|\widehat{p}) = \frac{1}{2} \left[\log\left(\frac{\widehat{\sigma}^{2}}{\sigma^{2}} \right)  -1 \right] +
\frac{\E[(X-\widehat{m}(Z))^{2}\mid \widehat{m}]}{2\widehat{\sigma}^{2}},
\end{align*}
using that $\E(X-m(Z))^2=\Var(\varepsilon)=\sigma^2$ for both $F$ and $G$. Note that
\begin{align}\label{eq:compsq}
(X-\widehat{m}(Z))^{2}=\varepsilon^2 + (m(Z)-\widehat{m}(Z))^2 + 2\varepsilon (m(Z)-\widehat{m}(Z)).
\end{align}
Since $F$ is the true model, $\varepsilon_F \indp Z$ and thus $\E[\varepsilon_F (m_F(Z)-\widehat{m}_F(Z))\mid \widehat{m}_F]=0$. Taking conditional expectation of~\eqref{eq:compsq} given $\widehat{m}_F$, we have 
\begin{align}\label{eq:expected-kl-gaussian-anm}
D(F\|\widehat{F}) =
\frac{1}{2} \left[\log\left(\frac{\widehat{\sigma}_F^{2}}{\sigma^{2}} \right) + \frac{\sigma^{2}}{\widehat{\sigma}_F^{2}} -1 \right] +
\frac{\E[(\widehat{m}_F(Z)-m_F(Z))^{2}\mid \widehat{m}_F]}{2\widehat{\sigma}_F^{2}}.
\end{align}
To achieve the required rates in condition (iii), we assume the following:
\begin{enumerate} \label{asp:gaussian-model-condition3-assumptions}
    \item[(iii-1)] $\E(\widehat{\sigma}^{2} - \sigma^{2})^2=o(n^{-1/2})$ for both $\widehat F$ and $\widehat G$. 
    \item[(iii-2)] $\mathbb{E}(\widehat{m}(Z) - m(Z))^{2} = o(n^{-1/2})$ for both $\widehat F$ and $\widehat G$.
\end{enumerate}

The function
$g(t)=\log(t/\sigma^2)+\sigma^2/t-1$
satisfies $g(\sigma^2)=g'(\sigma^2)=0$ and $g''(\sigma^2)=1/\sigma^4$, hence
a Taylor expansion shows that
$g(\widehat\sigma^2)=O\bigl((\widehat\sigma^2-\sigma^2)^2\bigr).$
Consequently, if $\mathbb E(\widehat\sigma^2-\sigma^2)^2=o(n^{-1/2})$,  then the first
term, $\frac{1}{2}g(\widehat{\sigma}_F^2)$, on the right-hand side of \eqref{eq:expected-kl-gaussian-anm} is
$o_p(n^{-1/2})$.
By Assumption (iii-1), $\widehat\sigma^2\to\sigma^2$ in probability and is bounded away from zero. 
The second term in \eqref{eq:expected-kl-gaussian-anm} is then $o_p(n^{-1/2})$ due to the $L_2$ rate in (iii-2).
This shows that condition (iii) with $b=1/2$ is satisfied for $\widehat{F}$.

For $D(G\|\widehat{G})$, the only difference is that we also need to
consider the expectation of the last term in~\eqref{eq:compsq} since $\varepsilon_G$ is not necessarily independent of $Z$. 
By the Cauchy-Schwarz inequality,
\[
\left|\E[\varepsilon_G (m_G(Z)-\widehat{m}_G(Z))\mid \widehat{m}_G]\right| \leq [\Var(\varepsilon_G)]^{1/2}\left\{ \E[(\widehat{m}_G(Z)-m_G(Z))^2\mid \widehat{m}_G]\right\}^{1/2}.
\]
Assumption (iii-2) implies that $\E[(\widehat{m}_G(Z)-m_G(Z))^2\mid \widehat{m}_G]=o_p(n^{-1/2})$
and thus 
\[
\E[\varepsilon_G (m_G(Z)-\widehat{m}_G(Z))\mid \widehat{m}_G]=o_p(n^{-1/4}).
\]
Together with $\widehat{\sigma}_G^2 \xrightarrow{p} \sigma_G^2>0$ due to (iii-1), this shows that $D(G\|\widehat{G})=o_p(1)$, i.e., condition (iii) holds with $b=0$ for $\widehat{G}$.

We summarize these results into a lemma:
\begin{lemma}\label{lemma:gaussian-anm-cond3}
    Under assumptions (iii-1) and (iii-2) in Appendix~\ref{asp:gaussian-model-condition3-assumptions}, Condition (iii) of Theorem~\ref{theorem:np_lrt_clt} holds for $\widehat{F}$ with $b=1/2$ and holds for $\widehat{G}$ with $b=0$.
\end{lemma}

\section{Proofs}

\subsection{Proof of Theorem~\ref{theorem:population-eo}} \label{app:proof-theorem-1}

Before proving Theorem~\ref{theorem:population-eo}, we first state a relevant result:

\begin{lemma}
Suppose the variables (nodes) of the DAG $\calG_0$ follow a restricted ANM. Let $(X, Y)$ be an undirected edge in $\calG$ on Line~\ref{lst:line:snoe-population-edge-to-orient} at any stage of Algorithm~\ref{alg:sneo-population-version} with the input being the CPDAG of $\calG_0$. Let $Z_{1}=\pa_{\calG}(X)$ and $Z_{2}=\pa_{\calG}(Y)$. If $\pa_{\calG_0}(X) = Z_1, \pa_{\calG_{0}}(Y) = Z_2 \cup \{X\}$ or $\pa_{\calG_0}(Y) = Z_2, \pa_{\calG_{0}}(X) = Z_1 \cup \{Y\}$, then $[X,Y|Z_{1}, Z_{2}]$ follows a pairwise additive noise model. 
\label{lem:full-pa-panm}
\end{lemma}

Lemma~\ref{lem:full-pa-panm} identifies a type of undirected edge, amongst all, that satisfy the PANM. We will show that the PDAG $\calG$ contains at least one such undirected edge on Line~\ref{lst:line:snoe-population-edge-to-orient} at any iteration of the algorithm. Now we prove the theorem:

\begin{proof}
    To prove that the sequential edge orientation procedure can recover the true DAG $\calG_{0}$ by correctly orienting all undirected edges in the CPDAG $\mathcal{E}$, we demonstrate that the following holds true for $\calG$ at every iteration at Line~\ref{lst:line:snoe-population-edge-to-orient} in Algorithm~\ref{alg:sneo-population-version}: If $\calG$ is not a DAG, then there exists an undirected edge $X-Y$ such that $[X, Y \mid Z_{1}, Z_{2}]$ satisfies the PANM, where $Z_{1} = \pa_{\calG}(X)$ and $Z_{2} = \pa_{\calG}(Y)$. It is easy to see that every orientation step in Algorithm~\ref{alg:sneo-population-version} will only lead to correct orientation that is consistent with the DAG $\calG_0$ if the input $\calG$ is the CPDAG. 
    
    Let $T$ be an undirected component of the PDAG $\calG$ of size $|T|\geq 2$. Let $v_{1}\in T$ be a node that precedes all other nodes in $T$ according to some topological ordering $\prec$ of $\calG_{0}$. Because $|T|\geq 2$, the neighbor set $\nbr_\calG(v_{1})$ is not empty. If there exists $v_{2} \in \nbr_\calG(v_{1})$ such that $\{v_{1}, v_{2}\}$ satisfies the conditions in Lemma~\ref{lem:full-pa-panm}, then $(v_1,v_2)$ satisfies the PANM and the proof is complete. By construction all parents of $v_1$ have been identified in $\calG$, i.e. $\pa_\calG(v_1)=\pa_{\calG_0}(v_1)$. Let $v_2$ be a node that precedes all other nodes in $\nbr_\calG(v_{1})$ according to some sort of $\calG_0$. Then, $v_1\in\pa_{\calG_0}(v_2)$ and none of the nodes in $\nbr_\calG(v_{1})$ is a parent of $v_2$ in $\calG_0$. It remains to show that $\pa_{\calG_0}(v_2)=\pa_{\calG}(v_2) \cup \{v_1\}$.    
    If this is not the case, then there must exist another node $v_{j} \in \nbr_\calG(v_2)\setminus\{v_1\}$ that is a parent of $v_{2}$ in $\calG_0$ and $v_j\not\in \nbr_\calG(v_{1})$, i.e. there is no undirected edge between $v_1$ and $v_j$ in the PDAG $\calG$. There are two possibilities with respect to the connectivity between $v_1$ and $v_j$ in $\calG_0$. 
    The first possibility is that there is no edge between $v_1$ and $v_j$ in $\calG_0$. This would form a new v-structure $v_{1} \rightarrow v_{2} \leftarrow v_{j}$ in $\calG_{0}$, which is a contradiction to that the input $\calG$ is the CPDAG of $\calG_0$. 
    The second possibility is that $v_1\to v_j$ in $\calG_0$ since $v_1\prec v_j\in T$ and this edge has been oriented so in $\calG$. Since neither $v_1-v_2$ nor $v_2-v_j$ has been oriented in $\calG$, the edge $v_1 \to v_j$ must have been oriented either by Line~\ref{lst:line:snoe-population-orient-nc} or by Meek's rule 1 on Line~\ref{lst:line:snoe-population-meeks-rules}. In what follows, we show that both scenarios would lead to contradictions, and thus such $v_j$ does not exist.

    If $v_{1} \rightarrow v_{j}$ is oriented by Line~\ref{lst:line:snoe-population-orient-nc}, then there must be another node $v_{i}$ such that $v_{1} - v_{i}$ was oriented in either direction on Line~\ref{lst:line:snoe-population-orient-found-edge} first and $v_i$ is adjacent to $v_j$. The algorithm would also orient $v_{i} \rightarrow v_{j}$ by Line~\ref{lst:line:snoe-population-orient-nc} or from previous actions. Shown in Figure~\ref{fig:thm1-orient-by-nc} (a) and (b), there are two cases regarding the adjacency of $v_2$ and $v_i$ in $\calG$ assuming $v_{1} \rightarrow v_{i}$ has been oriented. (a)~If $v_{i}, v_{2}$ are adjacent, $v_i$ must be a parent of $v_2$ because otherwise there would be a directed cycle $v_2\to v_i\to v_j \to v_2$ in $\calG_0$. Then, Line~\ref{lst:line:snoe-population-orient-nc} would orient $v_{1} \rightarrow v_{2}$ too, contradicting to that $v_1-v_2$ is undirected in $\calG$. (b)~If $v_{i}, v_{2}$ are not adjacent, then the algorithm would orient $v_{j} \rightarrow v_{2}$ by Meek's rule 1 in the following step, contradicting to that
    $v_j\in\nbr_\calG(v_2)$ (i.e. $v_j-v_2$ in $\calG$). Similar arguments under the orientation $v_{i} \rightarrow v_{1}$ result in contradictions that supposedly undirected edges in $\calG$ would have been oriented.

    If $v_{1} \rightarrow v_{j}$ is oriented by Meek's rule 1 on Line~\ref{lst:line:snoe-population-meeks-rules}, then there must exist a node $Z \in V$ in the configuration $Z \rightarrow v_{1}- v_j$ and $Z$ is not adjacent to $v_j$.
There are four possible cases, depicted in Figure~\ref{fig:thm1-orient-by-meeks-rule}, with respect to the connection between $Z$ and $v_2$ in the PDAG $\calG$. Case 1: there is no edge between $Z$ and $v_{2}$. The undirected edge $v_{1}- v_{2}$ would then be oriented by rule 1 as $v_1\to v_2$, which leads to a contradiction. 
Case 2: The two nodes are connected by an undirected edge $Z - v_{2}$. Then, $Z\in T$ and is a parent node of $v_1$, contradicting the fact that $v_1$ precedes all other nodes in $T$. Case 3: $Z \rightarrow v_{2}$ in $\calG$. Meek's rule 1 would then orient $v_{2} \rightarrow v_{j}$, which would result in an incorrect orientation as $v_j$ is assumed to be a parent of $v_2$ in $\calG_0$, again a contradiction. Case 4: $v_{2} \rightarrow Z$ in $\calG$. Then the edge $v_{1} - v_{2}$ must be $v_2\to v_1$ in $\calG_0$, which contradicts the assumed ordering $v_{1} \prec v_{2}$.
\end{proof}

\begin{figure}
\centering
\begin{tikzpicture}[node distance={16mm}, thick, main/.style = {draw, circle}] 
\node(temp1) { }; 
\node[main] (v1) [below left of =temp1]{$v_{1}$}; 
\node[main] (v2) [below right of=temp1] {$v_{2}$};
\node[main] (vi) [below of=v1] {$v_{i}$};
\node[main] (vj) [below of=v2] {$v_{j}$};
\node[below of=temp1, node distance = 34mm]{(a) $v_{i}, v_{2}$ are adjacent};
\draw[->] (v1) -- (vi);
\draw[->] (v1) -- (vj); 
\draw[->, color=red] (v1) -- (v2); 
\draw[->] (vi) -- (vj); 
\draw[->] (vi) -- (v2); 
\draw (vj) -- (v2); 
\end{tikzpicture} 
\hspace{1.25cm}
\begin{tikzpicture}[node distance={16mm}, thick, main/.style = {draw, circle}] 
\node(temp1) { }; 
\node[main] (v1) [below left of =temp1]{$v_{1}$}; 
\node[main] (v2) [below right of=temp1] {$v_{2}$};
\node[main] (vi) [below of=v1] {$v_{i}$};
\node[main] (vj) [below of=v2] {$v_{j}$};
\node[below of=temp1, node distance = 34mm]{(b) $v_{i}, v_{2}$ are not adjacent};
\draw[->] (v1) -- (vi);
\draw[->] (v1) -- (vj); 
\draw (v1) -- (v2); 
\draw[->] (vi) -- (vj); 
\draw[->, color=red] (vj) -- (v2); 
\end{tikzpicture} 
\caption{First possibility: orienting $v_{1} \rightarrow v_{j}$ by Line~\ref{lst:line:snoe-population-orient-nc} in Algorithm~\ref{alg:sneo-population-version} results in contradictions where $v_{1} \rightarrow v_{2}$ is oriented in (a) or $v_{j} \rightarrow v_{2}$ in (b).}
\label{fig:thm1-orient-by-nc}
\end{figure}

\begin{figure}
\centering
\begin{tikzpicture}[node distance={16mm}, thick, main/.style = {draw, circle}] 
\node(temp1) { }; 
\node[main] (Z) [below left of =temp1]{Z}; 
\node[main] (v1) [below right of=temp1] {$v_{1}$};
\node[main] (v2) [below of=Z] {$v_{2}$};
\node[main] (vj) [below of=v1] {$v_{j}$};
\node[below of=temp1, node distance = 34mm] {Case 1};
\draw[->] (Z) -- (v1);
\draw[->] (v1) -- (vj); 
\draw[->, color=red] (v1) -- (v2); 
\draw (vj) -- (v2); 
\end{tikzpicture} 
\hspace{0.6cm}
\begin{tikzpicture}[node distance={16mm}, thick, main/.style = {draw, circle}] 
\node(temp1) { }; 
\node[main] (Z) [below left of =temp1]{Z}; 
\node[main] (v1) [below right of=temp1] {$v_{1}$};
\node[main] (v2) [below of=Z] {$v_{2}$};
\node[main] (vj) [below of=v1] {$v_{j}$};
\node[below of=temp1, node distance = 34mm] {Case 2};
\draw[->] (Z) -- (v1);
\draw[->] (v1) -- (vj); 
\draw (Z) -- (v2); 
\draw (v1) -- (v2); 
\draw (v2) -- (vj); 
\end{tikzpicture} 
\hspace{0.6cm}
\begin{tikzpicture}[node distance={16mm}, thick, main/.style = {draw, circle}] 
\node(temp1) { }; 
\node[main] (Z) [below left of =temp1]{Z}; 
\node[main] (v1) [below right of=temp1] {$v_{1}$};
\node[main] (v2) [below of=Z] {$v_{2}$};
\node[main] (vj) [below of=v1] {$v_{j}$};
\node[below of=temp1, node distance = 34mm] {Case 3};
\draw[->] (Z) -- (v1);
\draw[->] (v1) -- (vj); 
\draw[->] (Z) -- (v2); 
\draw (v1) -- (v2); 
\draw[->, color=red] (v2) -- (vj);
\end{tikzpicture} 
\hspace{0.6cm}
\begin{tikzpicture}[node distance={16mm}, thick, main/.style = {draw, circle}] 
\node(temp1) { }; 
\node[main] (Z) [below left of =temp1]{Z}; 
\node[main] (v1) [below right of=temp1] {$v_{1}$};
\node[main] (v2) [below of=Z] {$v_{2}$};
\node[main] (vj) [below of=v1] {$v_{j}$};
\node[below of=temp1, node distance = 34mm] {Case 4};
\draw[->] (Z) -- (v1);
\draw[->] (v1) -- (vj); 
\draw[->] (v2) -- (Z); 
\draw[->, color=red] (v2) -- (v1); 
\draw (v2) -- (vj);
\end{tikzpicture} 
\label{fig:population-eo-proof}
\caption{Second possibility: edge $v_{1} \rightarrow v_{j}$ can be oriented by Meek's rule 1 under four cases, all of which lead to a contradiction.}
\label{fig:thm1-orient-by-meeks-rule}
\end{figure}

\subsection{Proof of Theorem~\ref{theorem:np_lrt_clt}}\label{sec:proof-np-lrt-clt}

\begin{proof}
Conditioning on $(\widehat F,\widehat G)$, the variables
$\{R_i\}_{i=1}^n$ are i.i.d.\ with conditional mean
\[
m_n=\E\!\left[
\log\frac{\widehat F(X\mid Z)}{\widehat G(X\mid Z)}\,\left|\,\widehat F,\widehat G\right.\right]
\]
and conditional variance $v_n$.
By Lyapunov central limit theorem, applicable because of condition (i), 
\begin{equation}\label{eq:condCLT}
\Big[\frac{\sqrt{n}(\bar R-m_n)}{\sqrt{v_n}}\,\left|\,
(\widehat F,\widehat G)\right.\Big]
\overset{D}{\longrightarrow}
N(0,1).
\end{equation}
It is easy to see that
\begin{align}\label{eq:absm_n}
\mu_0-m_n=D(F\,\|\,\widehat{F})+D(G\,\|\,\widehat{G})=o_p(n^{-b})
\end{align}
by condition (iii).
If $b=1/2$, this implies $\sqrt{n}\,(m_n-\mu_0) \xrightarrow{p} 0$.
Condition (ii) ensures $v_n \xrightarrow{p} \sigma^2_0$. 
Together with~\eqref{eq:condCLT}, we have
\[
\frac{\sqrt{n}(\bar R-\mu_0)}{\sigma_0} \overset{D}{\longrightarrow} N(0,1).
\]
The result then follows by Slutsky's theorem as $s_R^2\xrightarrow{p}v_n \xrightarrow{p} \sigma^2_0$, the first convergence due to condition (i).

If $b=0$, inequality~\eqref{eq:absm_n} implies that $m_n\xrightarrow{p} \mu_0$. Applying the weak law of large numbers to $\bar R$ conditioning on $(\widehat{F},\widehat{G})$, we have $\bar R-m_n \xrightarrow{p} 0$ and thus $\bar R\xrightarrow{p} \mu_0$.

\end{proof}

\subsection{Proof of Theorem~\ref{theorem:finite-sample-eo}}

\begin{proof}
The intersection of the following three events imply that Algorithm~\ref{alg:sneo-practical-version} will recover the true DAG $\calG_0$: 
\begin{enumerate}
    \item The algorithm recovers the true CPDAG of $\calG_0$.
    \item The independence measure ranks undirected edges that satisfy the PANM ahead of those that do not.
    \item The likelihood ratio test returns the true orientation of an undirected edge that satisfying the PANM.
\end{enumerate}
Note that sample splitting is used in the second and third events.

Together, the three events imply that the simplified Algorithm~\ref{alg:sneo-practical-version} coincides with its population version Algorithm~\ref{alg:sneo-population-version}, and thus, by Theorem~\ref{theorem:population-eo}, it outputs the true DAG $\calG_0$.
In the following three subsections, we prove each of the three events occurs with probability approaching one in the large sample limit,
which completes the proof.
\end{proof}

\subsubsection{Consistency of Initial CPDAG Learning} 

Our method uses the PC algorithm to learn the initial graph, which is achieved by iteratively finding a separating set $S_{ij}$ to test the conditional independence between $(X_i, X_j)$. Our implementation uses the partial correlation test. The consistency of this step is given by
the following result.

\begin{lemma}[Consistency of Initial Learning Algorithm]
    Suppose that the joint distribution of $(X_{1},\ldots,X_p)$ is faithful to a DAG $\calG_{0}$ and satisfies (A1) of Assumption~\ref{asp:large-sample}. Let $\mathcal{E}$ be the CPDAG of $\calG_{0}$ and let $\widehat\calG$ be the graph estimated by the initial learning phase (Lines~\ref{lst:line:alpha2skelstart} $\textendash$~\ref{lst:line:stage1pdag} of Algorithm~\ref{alg:sneo-practical-version}). For some choice of $\alpha_{1}, \alpha_{2} \rightarrow 0, P(\widehat{\calG} = \mathcal{E}) \rightarrow 1$ as $n \rightarrow \infty$.
\label{lem:consistent-pc}
\end{lemma}

\begin{proof} 
To prove this lemma, we show that the conditional independence tests are pointwise consistent. In the partial correlation test, let $\rho_{ij \mid S}^{*}$ and $\widehat{\rho}_{ij \mid S}$ respectively denote the true and estimated partial correlation values of $[X_{i}, X_{j} \mid S]$. Let 
$$Z_{n}(i, j, S) = \frac{1}{2} \log\frac{1 + \widehat{\rho}_{ij \mid S}}{1 - \widehat{\rho}_{ij \mid S}}$$ 
be the Fisher z-transformation applied to the estimated partial correlation, and $Z^{*}$ be defined similarly for $\rho_{ij \mid S}^{*}$. Under $H_{0}: X_{i} \indp X_{j} \mid S$, the quantity $\sqrt{n-|S|-3}\ Z_{n}(i, j, S) \overset{D}{\longrightarrow} N(0, 1)$ as $n \rightarrow \infty$. Let $Z_{1-\gamma_{n}/2}$ denote the critical value corresponding to the percentile $1-\gamma_{n}/2$ under the standard normal distribution. Under this asymptotic distribution, the probability of a type I error $P(|Z_{n}| > Z_{1-\gamma_{n}/2} \mid H_{0}) \rightarrow \gamma$ if the sequence $\gamma_{n} \rightarrow \gamma$ as $n \rightarrow \infty$.

By (A1) of Assumption~\ref{asp:large-sample}, $|\rho_{ij \mid S}^{*}| > \tau > 0$ if $X_i\not\indp X_j\mid S$. The power of the test is
\begin{align*}
&P\left(|Z_{n}(i, j, S)|\sqrt{n-|S|-3} > Z_{1-\gamma_{n}/2}\right)\\
> &P\left[\Big(\frac{1}{2}\log\frac{1+ \tau}{1-\tau} + O_{p}(n^{-1/2})\Big)\sqrt{n-|S|-3} > Z_{1-\gamma_{n}/2}\right],
\end{align*}

since $|Z_n(i, j, S)|>\frac{1}{2} \log\frac{1 + \tau}{1 - \tau}+O_p(n^{-1/2})$. Observe that 
$$\Big(\frac{1}{2}\log\frac{1+ \tau}{1-\tau} + O_{p}(n^{-1/2})\Big)\sqrt{n-|S|-3 } = C_{\tau}\sqrt{n}+O_p(1) \overset{p}\longrightarrow \infty, $$
where $C_{\tau} >0$ is some constant dependent on $\tau$. 
Thus, we can choose $\gamma_{n} \rightarrow 0$ such that $Z_{1-\gamma_{n}/2} = o(\sqrt{n})$ and 
the power converges to 1 as $n\to\infty$.  

We can specify the input significance levels as $\alpha_1=\gamma_n$ and $\alpha_2=(1+\delta)\gamma_n$ for some constant $\delta \geq 0$. By obtaining the correct skeleton and applying Meek's rules, the algorithm recovers the true CPDAG in the large-sample limit.

\end{proof}

\subsubsection{Consistency of Edge Ranking}

In the edge orientation procedure, the algorithm ranks edges by the normalized, edge-wise mutual information in~\eqref{eq:normalized-mi} to identify edges satisfying the PANM criterion. The calculation involves first discretizing variables $\varepsilon_{i,S}, X_{k}$, then computing the mutual information $\widehat{MI}(\varepsilon_{i, S}, X_{k})$ for any $k \in S \subseteq \pa_{\calG}(i) \cup \ch_{\calG}(i)$ in PDAG $\calG$. We first show the consistency of the mutual information estimator $\widehat{MI}$. 

Suppose $(X,Z)$ satisfy an additive regression model
\[
X = m(Z) + \varepsilon,
\]
where $Z=(Z_1,\ldots,Z_d)$ and $\varepsilon$ could be dependent. Note that $m$ and $\varepsilon$ can be defined by~\eqref{eq:population-regression-form} and~\eqref{eq:population-error-form}. Fix an index $j\in\{1,\ldots,d\}$.
Let $\{(x_i,z_i)\}_{i=1}^n$ be an i.i.d.\ test sample, and let
\[
\widehat{\varepsilon}_i := x_i-\widehat m(z_i),\qquad i=1,\ldots,n,
\]
where $\widehat m$ is constructed from an independent training sample, so that
$\widehat m \indep \{(x_i,z_i)\}_{i=1}^n$.

Let $\{I_a\}_{a=1}^A$ be a fixed finite collection of bins partitioning
$\mathbb{R}$ for $\varepsilon$, and let $\{J_b\}_{b=1}^B$ be a fixed finite
collection of bins partitioning the support of $Z_j$.
Define the (true) cell probabilities
\[
p_{ab} := P(\varepsilon\in I_a,\ Z_j\in J_b),\qquad
p_{a\cdot}:=\sum_{b=1}^B p_{ab},\qquad p_{\cdot b}:=\sum_{a=1}^A p_{ab}.
\]
Let $MI(\varepsilon,Z_j)$ denote the discretized mutual information
computed from the true cell probabilities $\{p_{ab}\}$:
\[
MI(\varepsilon,Z_j)
=
\sum_{a=1}^A\sum_{b=1}^B
p_{ab}\log\frac{p_{ab}}{p_{a\cdot}p_{\cdot b}}.
\]
Let $\widehat p_{ab}$ be the empirical frequency of
$\{(\widehat{\varepsilon}_i,z_{ij})\}_{i=1}^n$ in the bin $I_a\times J_b$.
Denote by $\widehat{MI}$ the plug-in mutual information estimator using the empirical frequencies $\{\widehat p_{ab}\}$.

We make two assumptions before presenting a result on the consistency of the mutual information estimator:

\begin{enumerate}\label{asp:mi-consistency-assumptions}
\item[(C1)] 
Let $\partial I:=\cup_{a=1}^A \partial I_a$ be the set of $\varepsilon$-bin
boundaries. Then
\[
P(\varepsilon\in \partial I)=0,
\quad\text{and}\quad
P\bigl(\dist(\varepsilon,\partial I)\le t\bigr)\to 0 \ \text{as } t\downarrow 0.
\]

\item[(C2)]
There exists $c_0>0$ such that $p_{ab}\ge c_0$ for all $a,b$.
\end{enumerate}
Note that for Gaussian $\varepsilon$, (C1) holds for any finite binning.

\begin{theorem}[Consistency of discretized MI based on test residuals]
\label{theorem:MI_residual_consistency}
Assume (C1) and (C2) hold true.
If $\widehat m$ is $L^2$ consistent in the sense that, for an independent
draw $Z\indep \widehat m$,
\begin{align} \label{eq:est-reg-model-l2-convergence}
    \E\!\left[(\widehat m(Z)-m(Z))^2\right]\to 0,
\end{align}
then
$\widehat{MI}\xrightarrow{p} MI(\varepsilon,Z_j)$ as $n\to \infty$.
\end{theorem}

\begin{proof}
Write $\Delta(z):=\widehat m(z)-m(z)$ and note that on the test sample
\[
\widehat{\varepsilon}_i
=
x_i-\widehat m(z_i)
=
m(z_i)+\varepsilon_i-\widehat m(z_i)
=
\varepsilon_i-\Delta(z_i).
\]

Fix a bin index $a$ and define the indicator
\[
U_{n,i}^{(a)} := \mathbf 1\{\widehat{\varepsilon}_i\in I_a\},
\qquad
U_i^{(a)} := \mathbf 1\{\varepsilon_i\in I_a\}.
\]
If $|\Delta(z_i)|\le t$ and $\dist(\varepsilon_i,\partial I)>t$, then
$\widehat{\varepsilon}_i=\varepsilon_i-\Delta(z_i)$ lies in the same $\varepsilon$-bin
as $\varepsilon_i$, hence $U_{n,i}^{(a)}=U_i^{(a)}$. Therefore,
\[
P\bigl(U_{n,i}^{(a)}\neq U_i^{(a)}\bigr)
\le
P\bigl(|\Delta(Z)|>t\bigr)
+
P\bigl(\dist(\varepsilon,\partial I)\le t\bigr),
\]
where $(\varepsilon,Z)$ is an independent draw (and $Z\indep \widehat m$).
Since $\E[\Delta(Z)^2]\to 0$, we have $\Delta(Z)\to 0$ in probability, so for
any fixed $t>0$, $P(|\Delta(Z)|>t)\to 0$. By (C1),
$P(\dist(\varepsilon,\partial I)\le t)\to 0$ as $t\downarrow 0$.
Choosing $t=t_n\downarrow 0$ slowly, we obtain
$P\bigl(U_{n,i}^{(a)}\neq U_i^{(a)}\bigr)\to 0.$
Because there are finitely many bins, the same conclusion holds jointly over
all $a=1,\ldots,A$.

For each cell $(a,b)$ define
\[
V_{n,i}^{(ab)} := \mathbf 1\{\widehat{\varepsilon}_i\in I_a,\ Z_{ij}\in J_b\},
\qquad
V_i^{(ab)} := \mathbf 1\{\varepsilon_i\in I_a,\ Z_{ij}\in J_b\}.
\]
Then $|V_{n,i}^{(ab)}-V_i^{(ab)}|\le \mathbf 1\{U_{n,i}^{(a)}\neq U_i^{(a)}\}$, so
\[
\E\bigl|V_{n,i}^{(ab)}-V_i^{(ab)}\bigr|
\le
P(U_{n,i}^{(a)}\neq U_i^{(a)})\to 0.
\]
By Markov's inequality,
\[
\left|\widehat p_{ab}-\frac{1}{n}\sum_{i=1}^n V_i^{(ab)}\right|
=
\left|\frac{1}{n}\sum_{i=1}^n (V_{n,i}^{(ab)}-V_i^{(ab)})\right|
\xrightarrow{p} 0,
\]
and hence for each fixed $(a,b)$,
\[
\widehat p_{ab}\xrightarrow{p} \frac{1}{n}\sum_{i=1}^n V_i^{(ab)} \xrightarrow{p} p_{ab}.
\]
Because $A,B$ are fixed and finite, we have joint convergence of the entire
finite vector $(\widehat p_{ab})_{a,b}$ to $(p_{ab})_{a,b}$ in probability, and
likewise for the marginals $\widehat p_{a\cdot}\to p_{a\cdot}$ and
$\widehat p_{\cdot b}\to p_{\cdot b}$.

Under (C2), the mutual information function
\[
\Phi\bigl((q_{ab})\bigr)
=
\sum_{a,b} q_{ab}\log\frac{q_{ab}}{q_{a\cdot}q_{\cdot b}}
\]
is continuous on the compact set $\{q_{ab}\ge c_0,\ \sum_{a,b}q_{ab}=1\}$.
Therefore, by the continuous mapping theorem,
\[
\widehat{MI}=\Phi\bigl((\widehat p_{ab})\bigr)\xrightarrow{p}\Phi\bigl((p_{ab})\bigr)
= MI(\varepsilon,Z_j).
\]
This concludes the proof.
\end{proof}

Theorem~\ref{theorem:MI_residual_consistency} shows that the estimated mutual information of the discretized variables $(\varepsilon, Z_j)$ converges in probability to the true mutual information as $n \rightarrow \infty$.
Assumption (C1) holds for any absolutely continuous error distribution and any fixed finite discretization, since bin boundaries have Lebesgue measure zero. Since $X_{S}$ and $\varepsilon_{i, S}$ in~\eqref{eq:population-error-form} follow absolute continuous distributions, (C1) holds. 
Assumption (C2) holds if there are no empty bins as stated in Assumption~\ref{asp:large-sample}. Equation~\eqref{eq:gaussian-anm-regression-convergence} in Assumption~\ref{asp:approx-reg-fx} implies~\eqref{eq:est-reg-model-l2-convergence}. Thus, we have $\widehat{MI}(\varepsilon_{i, S}, X_{k}) \overset{p}\longrightarrow MI(\varepsilon_{i, S}, X_{k})$ for all $k \in S$ and all $i\in[p]$.

The ranking procedure sorts undirected edges in the PDAG $\calG$ by an independence measure. We demonstrate that the ranking of undirected edges by Algorithm~\ref{alg:OrientEdges} ranks those satisfying the PANM before other edges in the large-sample limit.

Recall from Section 4.2 that the edge-wise independence measure $\widetilde{I}(X, Y)$ for random variables $X, Y$ is calculated by Equation~\eqref{eq:indms} (through discretization). Let $\widehat{\widetilde{I}}(X_{i}, X_{j})$ be the estimator of $\widetilde{I}(X_{i}, X_{j})$, which is calculated from the estimated mutual information $\widehat{MI}(\varepsilon_{v, S}, X_{k})$ for $k \in S \subseteq \pa_{\calG}(v) \cup \ch_{\calG}(v)$ and $v\in\{i,j\}$ and the entropy measures $\widehat{H}(\cdot)$ of such terms. By Theorem~\ref{theorem:MI_residual_consistency}, ${\widehat{MI}(\widehat{\varepsilon}_{v,S},X_k)} \overset{p}\longrightarrow MI({\varepsilon}_{v,S},X_k)$ for $k \in S$ and $v\in[p]$. The consistency of the entropy, $\widehat{H}(X) \overset{p}\longrightarrow H(X)$, follows from a similar result, as $H(X) = MI(X, X)$ for a random variable $X$. 

Therefore, if random variables $(X_{i}, X_{j})$ follow the PANM, then $\widehat{MI}(\cdot, \cdot) \overset{p}\longrightarrow  0$ under the true causal direction and thus $\widehat{\widetilde{I}}(X_{i}, X_{j}) \overset{p}\longrightarrow 0$ as $n \rightarrow \infty$. If $(X_{i}, X_{j})$ do not follow the PANM, then (A2) and (A3) of Assumption~\ref{asp:large-sample} imply that 
$\widehat{\widetilde{I}}(X_{i}, X_{j}) \overset{p}\longrightarrow \widetilde{I}(X_{i}, X_{j}) >{\delta}/{c_{2}} > 0$
as $n \rightarrow \infty$, where $\delta$ is the mutual information lower bound in (A2) and $c_{2}$ is the upper bound of entropy measures in (A3). 
This reasoning shows that our algorithm would rank all pairs of $(X,Y)$ that satisfy the PANM ahead of those that do not satisfy.

\subsubsection{Consistency of Likelihood Ratio Test}

Suppose the true causal direction of an undirected edge $X-Y$ is $X \rightarrow Y$ and $[X, Y \mid Z_{1}, Z_{2}]$ satisfies the PANM criterion, where $Z_{1}=PA_{X}$ and $Z_{2}=PA_{Y}$. Denote the models for the two orientations by $F: X\rightarrow Y$ and $G: Y\rightarrow X$ and their estimators by $\widehat{F}, \widehat{G}$ which are independent of the test samples $\{(X_{i}, Y_{i}, Z_{1, i}, Z_{2, i})\}_{i=1}^{n}$. The LR test statistic $LR_{n}(\widehat{F}, \widehat{G})$~\eqref{eq:likelihood-ratio-stat} can be written as
\[
    \frac{1}{n}LR_{n}(\widehat{F}, \widehat{G})
    = \frac{1}{n}\sum_{i=1}^{n} \log \frac{\widehat{F}(X_i, Y_i \mid Z_{1, i}, Z_{2, i})}{\widehat{G}{(X_i, Y_i \mid Z_{1, i}, Z_{2, i})}} := \widehat{R}_{X} + \widehat{R}_{Y},
\]
a sum of two log-likelihood ratios with
\[
\widehat{R}_{X} = \frac{1}{n} \sum_{i=1}^{n}\log\frac{\widehat{F}_{X}(X_{i} \mid Z_{1, i})} {\widehat{G}_{X}(X_{i} \mid Y_{i}, Z_{1, i})}, \quad \widehat{R}_{Y} = \frac{1}{n} \sum_{i=1}^{n} \log\frac{\widehat{F}_{Y}(Y_{i} \mid X_{i}, Z_{2, i})} {\widehat{G}_{Y}(Y_{i} \mid Z_{2, i})}.
\]
Furthermore, let $\mu_X$ and $\mu_Y$ denote the population log-likelihood ratios. Note that $F_{X}, F_{Y}, G_{X}, G_{Y}$ are all Gaussian regression models~\eqref{eq:gaussian-regression-model}, where $F_X$ and $F_Y$ are true models.
We first show that all three conditions of Theorem~\ref{theorem:np_lrt_clt} are satisfied for both $\widehat{R}_X$ and $\widehat{R}_Y$
by verifying the assumptions of Lemmas~\ref{lemma:gaussian-anm-cond1},~\ref{lemma:gaussian-anm-cond2}, and~\ref{lemma:gaussian-anm-cond3}.

\paragraph{Verification of Condition (i)}
Assumption (i-1) in Appendix~\ref{asp:gaussian-model-condition1-assumptions} is implied by the finite $(4+2\eta)$-th moment requirement on the error variables $\varepsilon_{i,S}$ in Assumption~\ref{asp:approx-reg-fx}. Assumption (i-2) is implied by \eqref{eq:gaussian-anm-regression-moment-sup} of Assumption~\ref{asp:approx-reg-fx}, and (i-3) by \eqref{eq:gaussian-anm-finite-est-variance}. By Lemma~\ref{lemma:gaussian-anm-cond1}, Condition (i) of Theorem~\ref{theorem:np_lrt_clt} holds for both $\widehat{R}_X$ and $\widehat{R}_Y$.

\paragraph{Verification of Condition (ii)}
For both $(\widehat{R}_{X}, R_{X})$ and $(\widehat{R}_{Y}, R_{Y})$, assumptions (ii-1) to (ii-3) of Appendix~\ref{asp:gaussian-model-condition2-assumptions} are implied by the finite $(4+2\eta)$-th moments of the error variables, \eqref{eq:gaussian-anm-regression-convergence}, and \eqref{eq:gaussian-anm-variance-convergence} in Assumption~\ref{asp:approx-reg-fx}, respectively. By Lemma~\ref{lemma:gaussian-anm-cond2}, Condition (ii) of Theorem~\ref{theorem:np_lrt_clt} holds true for $\widehat{R}_X$ and $\widehat{R}_Y$.

\paragraph{Verification of Condition (iii)}

It is straightforward to see that \eqref{eq:gaussian-anm-variance-convergence} and \eqref{eq:gaussian-anm-regression-convergence} in Assumption~\ref{asp:approx-reg-fx} imply Assumptions (iii-1) and (iii-2) in Appendix~\ref{asp:gaussian-model-condition3-assumptions}
for both $\widehat{R}_X$ and $\widehat{R}_Y$. Thus, by Lemma~\ref{lemma:gaussian-anm-cond3}, Condition (iii) with $b=0$ holds for both
likelihood ratios.

Then, as an immediate consequence of Theorem~\ref{theorem:np_lrt_clt},
\[
\frac{1}{n}LR_{n}(\widehat{F}, \widehat{G}) = \widehat{R}_{X}+ \widehat{R}_{Y} \overset{p}\longrightarrow {\mu}_{X} + {\mu}_{Y}.
\]
By Lemma~\ref{lem:panm-identifiability}, the causal direction $X\to Y$ is identifiable, which implies that ${\mu}_{X} + {\mu}_{Y} >0$
and $\omega_*^2>0$. Therefore, together with $\widehat{\omega}_n^2 \xrightarrow[]{p} \omega^2_*$, we have
\[
LR_{n}(\widehat{F}, \widehat{G})/{(\sqrt{n}\widehat{\omega}_n)} = a \sqrt{n} +o_p(\sqrt{n}) \overset{p}\longrightarrow + \infty,
\] 
where $a>0$ is a constant. This establishes the consistency of the likelihood ratio test for determining the true causal direction
with the same $\alpha_1$ as in the proof of Lemma~\ref{lem:consistent-pc}.

\section{Supplementary Numerical Results}

\subsection{Comparison with Nonlinear Constraint-based Algorithms}\label{sec:appendix-constraint-alg}

We compared SNOE with three constraint-based methods for nonlinear DAG learning, kPC \citep{gretton2009nonlinear}, NNCL \citep{wang_zhou_nonlinear}, and RESIT \citep{resit_anm}. Note that, in general, kPC and NNCL produce a PDAG instead of a DAG. A significance level of $\alpha=0.01$ is used for both algorithms as recommended in their paper or software tutorials. RESIT does not scale well to $p > 20$ nodes, as mentioned in their paper and in several other works that compare with this algorithm. Thus, we did not include RESIT in this comparison since all networks in our experiment are of size $p \geq 24$. 

\begin{figure}[h]
  \centering
  \includegraphics[scale=0.6]{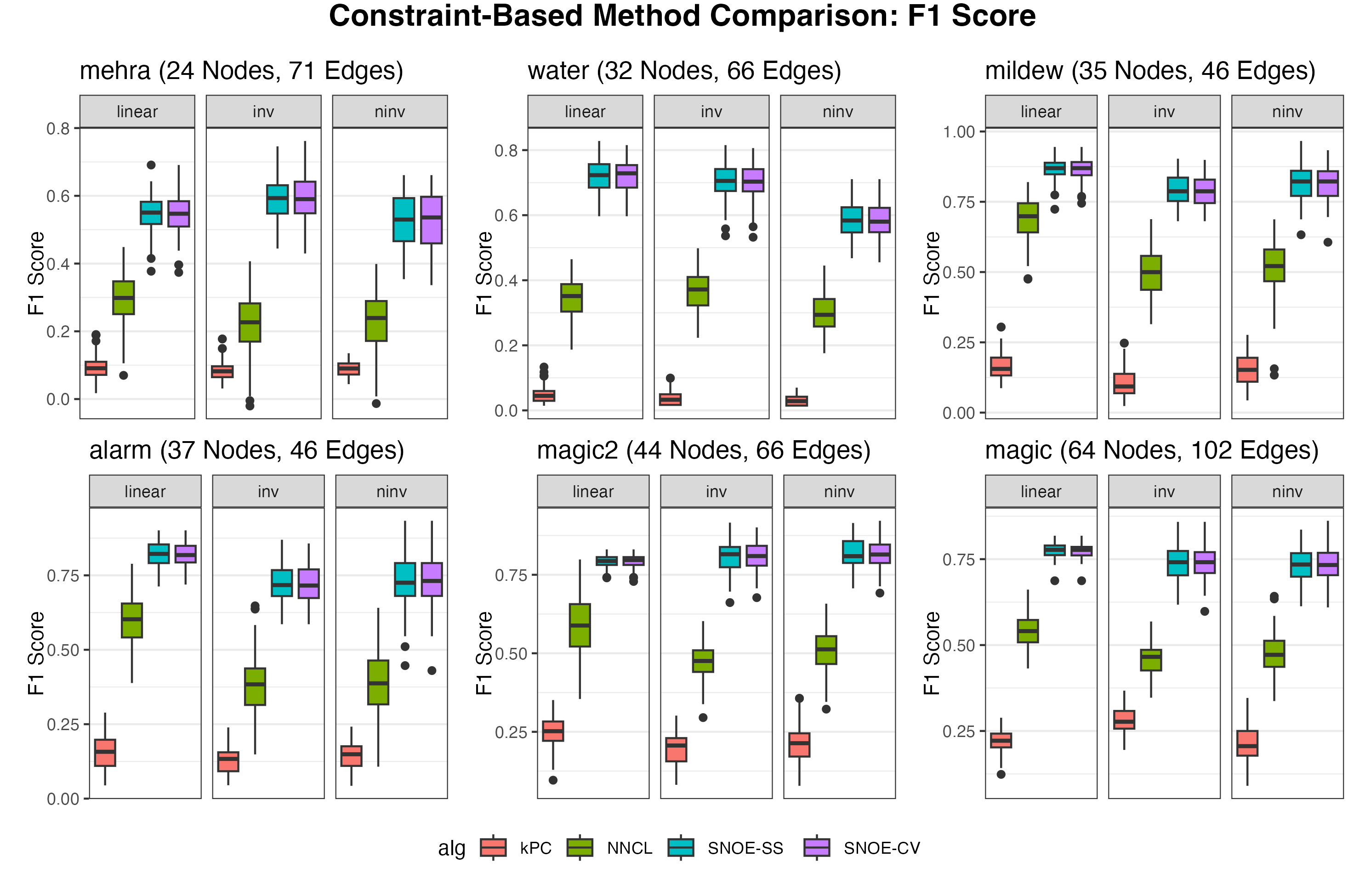}
  \caption{F1 scores of various constraint-based methods.}
\label{fig:constraint-based-methods-f1}
\end{figure}

Figure~\ref{fig:constraint-based-methods-f1} shows the F1 scores with respect to the true DAG. The results show that SNOE outperforms both kPC and NNCL under all settings. A detailed analysis reveals that kPC and NNCL have higher counts of extraneous and missing edges than SNOE. In particular, kPC has a greater amount of missing edges, likely due to sensitivity issues that stem from the HSIC test. The NNCL algorithm also uses the PC algorithm to estimate the CPDAG, yet its performance is most likely limited by the use of a piecewise linear function to approximate the underlying SEM. Both algorithms also contain a large number of incorrectly oriented and undirected edges in the estimated PDAG. As SNOE outputs a PDAG without applying the final stage, we also analyzed the PDAG output from SNOE (not shown). We see that the PDAGs learned by SNOE still have a higher F1 score than the two competing methods. The F1 score is just slightly lower than that of the DAG output, which is also reflected in Figure~\ref{fig:intermediate-f1-score} showing intermediate results. More detailed results show that there are still fewer undirected edges in the PDAG produced by SNOE than the other two  methods. This reflects the advantageous design of our method to order edges satisfying the PANM criterion first, enabling correct identification of the causal direction with more oriented edges.

\subsection{Performance Under General Nonlinear SEMs}\label{sec:appendix-general-nonlinear-sem-f1}

We conducted additional simulations to assess the performance under more general functional forms. The data was simulated separately under invertible functions, non-invertible functions (Gaussian processes), and a multi-layer perceptron (MLP). The SEM under the invertible and non-invertible functions is 
\[
X_{i} = \sum_{j \in \pa{(i)}}f_{j}(X_{j}) + \sum_{k,m \in \pa{(i)}} X_{k}X_{m} + \varepsilon_{i}.
\]
The function $f_{j}$ is either an invertible or non-invertible function and up to five pairwise interaction terms are introduced in the SEM. For the MLP, the data generating function can be expressed as $X_{i} = W_2\sigma(W_1 \PA_{i})+\varepsilon_{i}$, where $\PA_{i}$ is regarded as a column vector corresponding to the $k$ parent variables, $W_{1} \in \mathbb{R}^{h \times k}$ and $W_{2} \in \mathbb{R}^{1 \times h}$ for $h=200$, and $\sigma(\cdot)$ is the sigmoid function applied component-wise. Other settings remain the same as those detailed in Section~\ref{sec:experiments-simulated-data}. 

\begin{figure}
  \centering
  \includegraphics[scale=0.6]{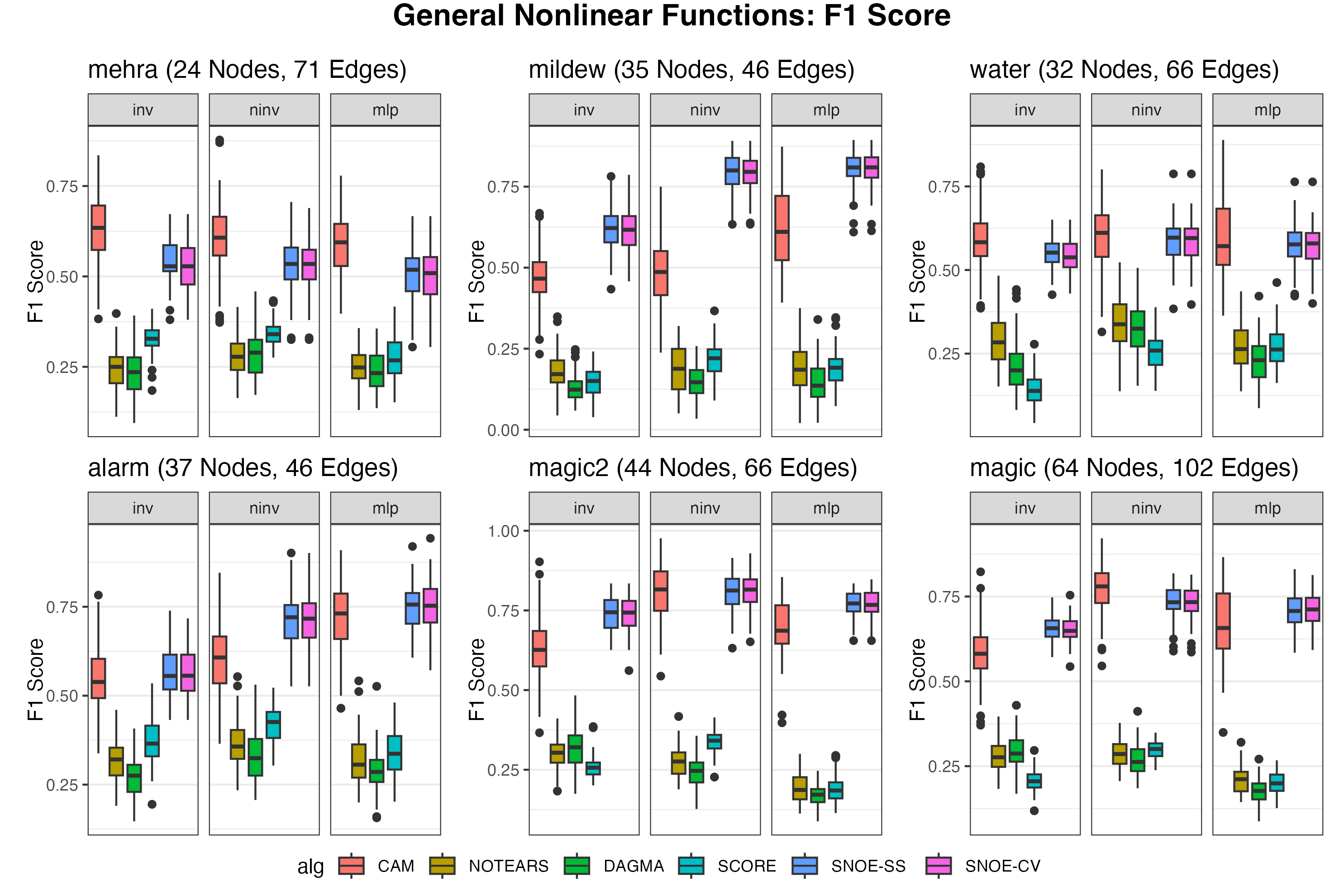}
  \caption{Performance comparison under violations of the CAM assumption, where the data generating functions are not purely additive.}
\label{fig:f1-general-nonlinear-fx}
\end{figure}

In Figure~\ref{fig:f1-general-nonlinear-fx}, we present a comparison of algorithm performances on data generated by general nonlinear functions. We observe an overall decrease in the F1 score compared to results on data generated under purely additive functions for all algorithms, as expected. A close analysis shows that SNOE has higher counts of false negatives, likely because the statistical power of the partial correlation test, utilized for learning the CPDAG, decreases when testing for more complex relations. While the difference in structural learning accuracies between SNOE and CAM is now smaller, SNOE still outperformed CAM in some cases. The F1 score for SNOE is more consistent and has a smaller interquartile range across all function types. CAM has fewer missing edges in comparison, but captures far more false positives. SNOE outperformed SCORE, NOTEARS, and DAGMA again in all cases. SCORE, NOTEARS, and DAGMA have higher skeleton learning accuracies under these settings, as neural networks can better capture nonlinearities, but also learned many incorrectly oriented edges. Moreover, SCORE captured many false positives and NOTEARS and DAGMA contain many false negatives. Under these more generalized nonlinear functions, the small change in F1 scores and consistent performance across different functions indicate that SNOE is robust to violations of the functional form.

\subsection{Accuracy of Initial Graph on Algorithm Performance}\label{sec:appendix-initial-graph-exp}

\begin{figure}
  \centering
  \includegraphics[scale=0.65]{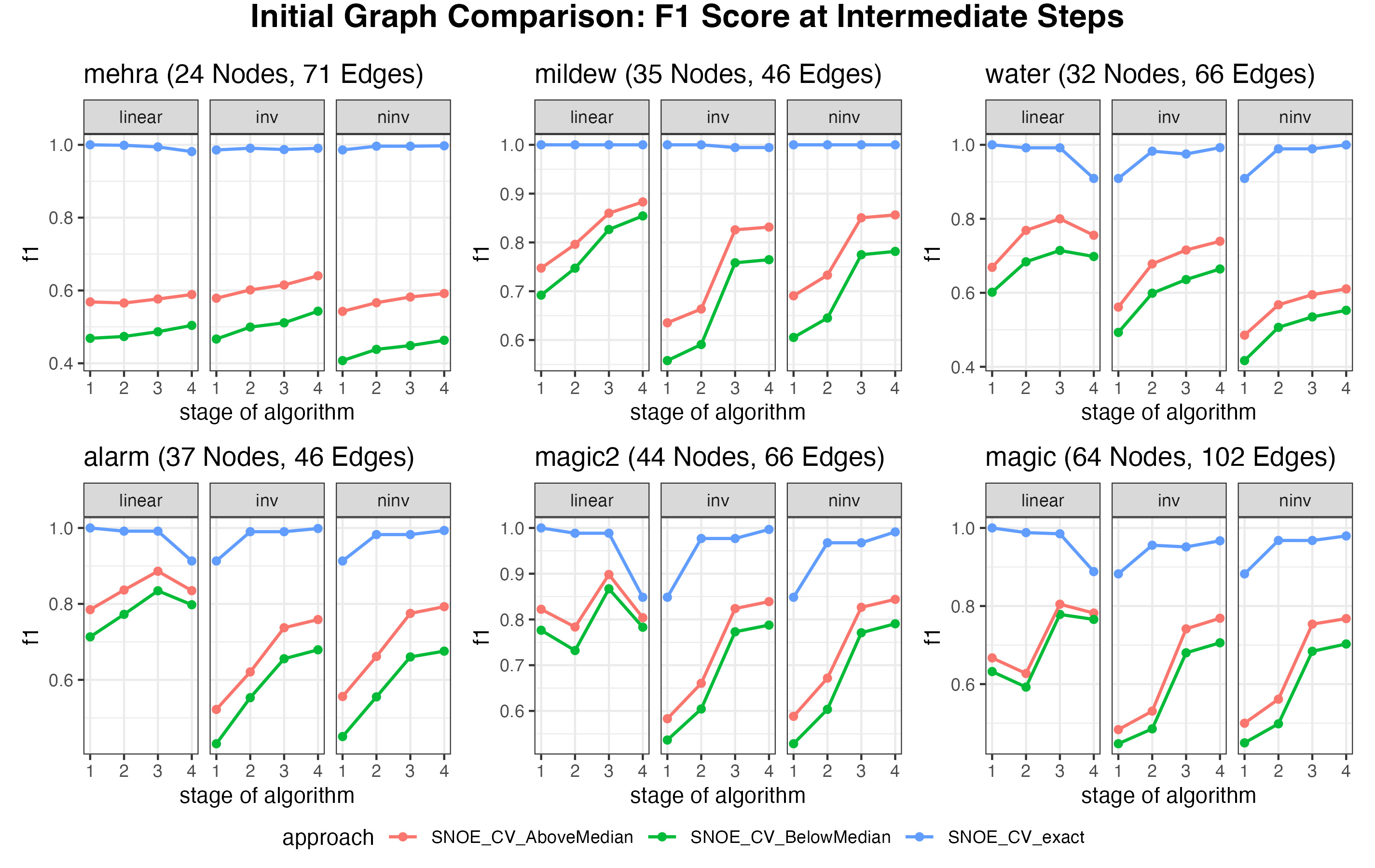}
  \caption{F1 score of SNOE using different initial graphs: the true CPDAG (`exact'), a learned CPDAG with above median F1 score (`AboveMedian'), and a learned CPDAG with below median F1 score (`BelowMedian').}
\label{fig:diff-cpdag-comparison}
\end{figure}

As SNOE improves upon a learned CPDAG, we investigate how the correctness of the initial graph affects the performance by testing our method separately on the exact, true CPDAG and the learned CPDAG. For the learned CPDAG, we split results by whether the F1 score, calculated with respect to the true CPDAG, is above the median across multiple data sets. The performances of the cross-validation (CV) and sample-splitting (SS) approaches are almost identical, so we only show that of the former. The results are shown in Figure~\ref{fig:diff-cpdag-comparison}, with the F1 score of each approach calculated according to the true DAG for nonlinear functions and the true CPDAG for linear functions. 

We first discuss the linear case when given the exact CPDAG, as one may notice that the F1 score decreases at the final stage. This is because we assume a nonlinear ANM and under this prior assumption, our algorithm extends the PDAG learned in step 3 to a DAG in the final step. Without this assumption, the algorithm would stop after stage 3, where the F1 score generally remains unchanged.
Under the other nonlinear settings, we see indeed the initial CPDAG is imperative to learning the true DAG. Yet, the parallel F1 curves between the ``AboveMedian'' and ``BelowMedian'' approaches indicate that an inaccurate estimation of a CPDAG does not exacerbate the difference in the final DAG learning accuracy. In fact, the results show that our subsequent steps are useful in learning the true DAG, whether using the true CPDAG or an estimated CPDAG as the starting point.

\subsection{True Positives Captured at Intermediate Steps}\label{sec:appendix-intermediate-step-tp}

\begin{figure}
  \centering
  \includegraphics[scale=0.575]{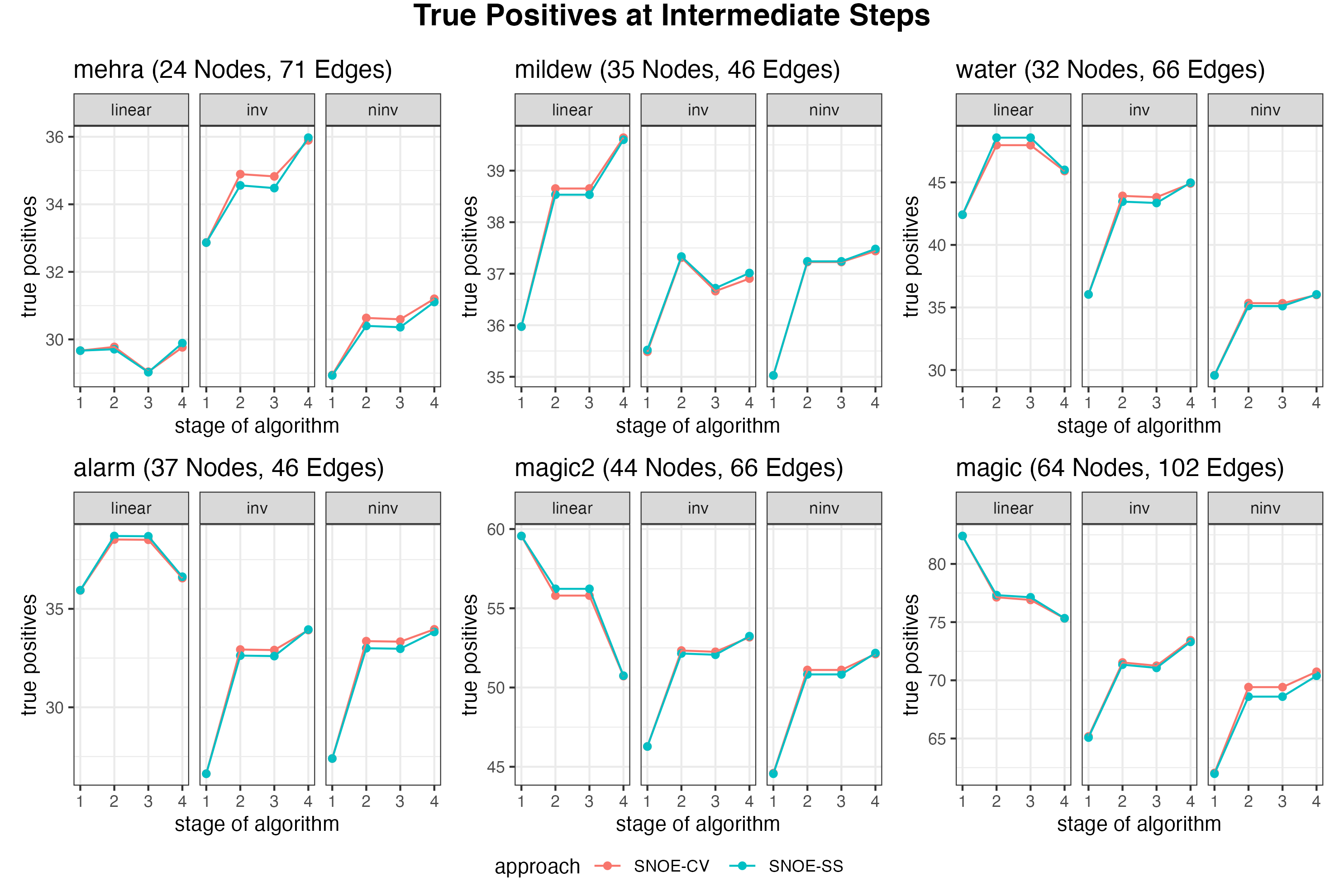}
  \caption{Number of true positives captured. The edge orientation step (2) uncovers more edges and lifts the F1 score despite having extra candidate edges in the graph.}
  \label{fig:all-alg-gaussian-tp}
\end{figure}

The number of true positives captured at each stage of the SNOE algorithm is shown in Figure~\ref{fig:all-alg-gaussian-tp} to complement and explain the intermediate F1 scores (see Figure~\ref{fig:intermediate-f1-score}). For data generated from nonlinear functions, the edge orientation procedure (step 2) uncovers a great number of true, directed edges from the learned CPDAG. Although the increase in the F1 score is relatively small at this stage since there are additional edges from the candidate set that may be false positives, the F1 score increases greatly once these edges are removed in the third stage. As for the linear, Gaussian DAG, the number of true positives remains unchanged, signaling high precision in the initial graph and correct inference on edge directions in subsequent stages. The results reflect the effectiveness of the edge orientation step in correctly orienting undirected edges. Moreover, the edge pruning step (step 3) deletes few if none true positives.

\subsection{Runtime of Intermediate Steps}\label{sec:appendix-intermediate-step-runtime}

\begin{figure}
  \centering
  \includegraphics[scale=0.7]{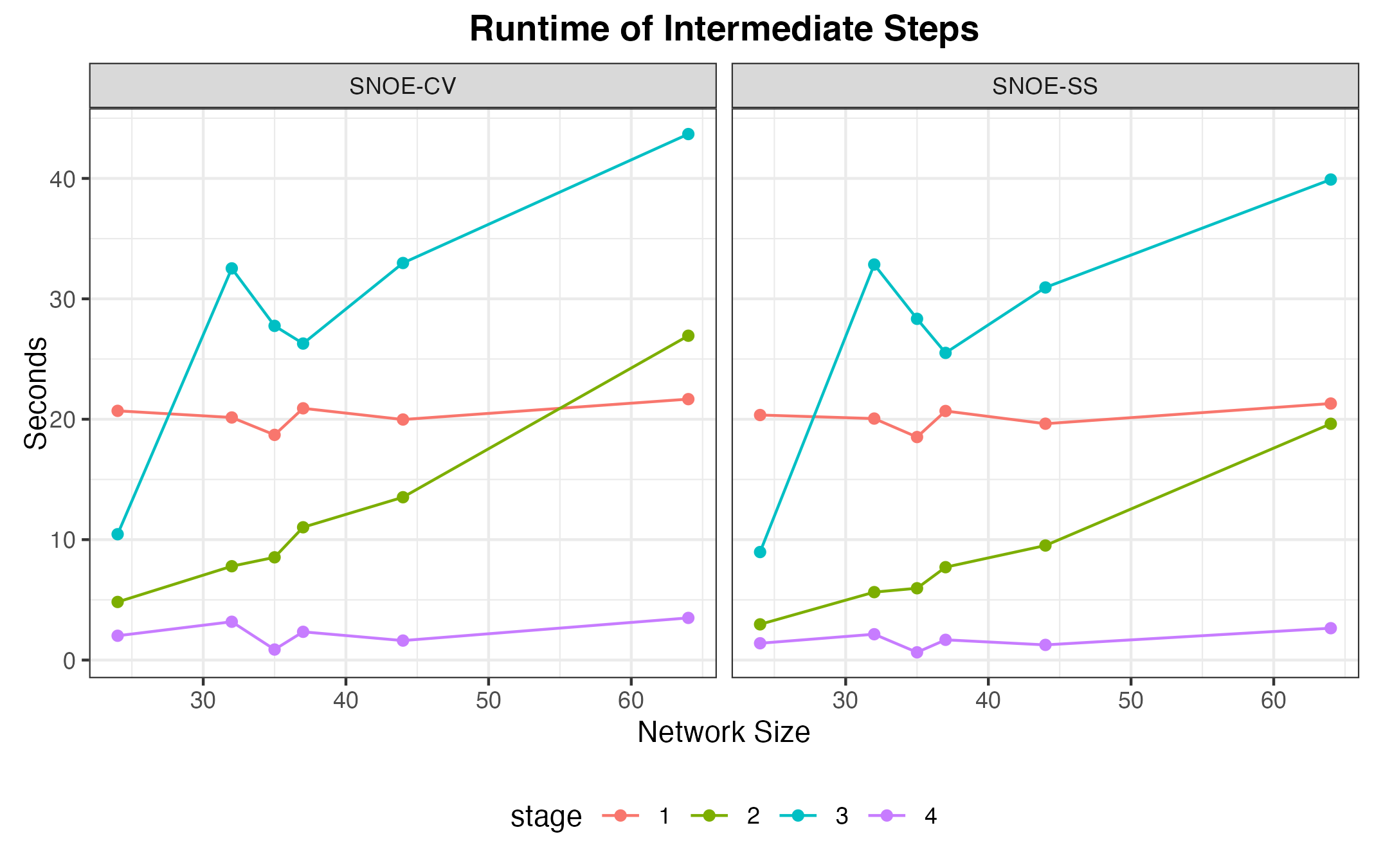}
  \caption{Average runtime of intermediate steps by network size. The steps are (1) initial CPDAG learning, (2) edge orientation, (3) edge deletion, and (4) graph refinement.}
\label{fig:runtime-intermediate-steps}
\end{figure}

To breakdown the computational cost, we report the average runtime of each intermediate step in Figure~\ref{fig:runtime-intermediate-steps}. Stages 1 (learning CPDAG by PC algorithm) and 4 (orienting graph into DAG per ANM assumption) exhibit similar runtimes across all networks. We utilized the PC-stable algorithm from the bnlearn package, which implements the iterative testing and orientation process in C++, and chose the partial correlation test for learning the skeleton, hence resulting in a lower runtime. The last stage orients remaining undirected edges in the PDAG, whose runtime is negligible relative to the other three stages. We observe that both stage 2 (edge orientation) and stage 3 (edge deletion) increase with network size, but at an approximately linear rate, indicating that our algorithm is scalable with graph size. 
SNOE-CV has a slightly higher runtime in stage 2 since we employ cross validation in the likelihood test. Moreover, the use of Meek's rules assists in further detecting more directed edges, hence reducing the actual number of edges to evaluate.

\subsection{Effect of Early Errors in Orientation Procedure}\label{sec:appendix-lrt-downstream-error}

\begin{table}[h]
\centering
\begin{tabular}{llllrrrr}
\hline
Network & Approach & Data & Graph & F1 & TP & Wrong Dir & Undir \\
\hline
mehra  & LRT-CV & inv  & before\_error & 0.446 & 26.528 & 6.696 & 4.857 \\
       &        &      & with\_error   & 0.460 & 27.366 & 8.335 & 2.379 \\
       &        &      & fixed\_error  & 0.483 & 28.745 & 7.932 & 1.404 \\
\hline
mehra  & LRT-CV & ninv & before\_error & 0.316 & 20.688 & 4.875 & 22.271 \\
       &        &      & with\_error   & 0.388 & 25.458 & 11.573 & 10.802 \\
       &        &      & fixed\_error  & 0.409 & 26.802 & 12.385 & 8.646 \\
\hline
mehra  & LRT-SS & inv  & before\_error & 0.447 & 26.632 & 6.712 & 4.908 \\
       &        &      & with\_error   & 0.462 & 27.491 & 8.362 & 2.399 \\
       &        &      & fixed\_error  & 0.483 & 28.742 & 7.951 & 1.558 \\
\hline
mehra  & LRT-SS & ninv & before\_error & 0.312 & 20.405 & 4.216 & 23.252 \\
       &        &      & with\_error   & 0.387 & 25.315 & 11.766 & 10.793 \\
       &        &      & fixed\_error  & 0.407 & 26.631 & 12.108 & 9.135 \\
\hline
magic2 & LRT-CV & inv  & before\_error & 0.510 & 42.612 & 2.225 & 10.969 \\
       &        &      & with\_error   & 0.543 & 45.325 & 6.444 & 4.037 \\
       &        &      & fixed\_error  & 0.587 & 49.025 & 5.631 & 1.150 \\
\hline
magic2 & LRT-CV & ninv & before\_error & 0.222 & 23.413 & 6.717 & 22.457 \\
       &        &      & with\_error   & 0.250 & 26.435 & 15.370 & 10.783 \\
       &        &      & fixed\_error  & 0.265 & 28.022 & 15.522 & 9.043 \\
\hline
magic2 & LRT-SS & inv  & before\_error & 0.510 & 42.649 & 2.030 & 11.250 \\
       &        &      & with\_error   & 0.542 & 45.375 & 6.494 & 4.060 \\
       &        &      & fixed\_error  & 0.584 & 48.839 & 5.702 & 1.387 \\
\hline
magic2 & LRT-SS & ninv & before\_error & 0.226 & 23.829 & 7.158 & 20.974 \\
       &        &      & with\_error   & 0.248 & 26.237 & 15.289 & 10.434 \\
       &        &      & fixed\_error  & 0.260 & 27.487 & 15.079 & 9.395 \\
\hline
\end{tabular}

\caption{F1 score and edge breakdown of the orientation procedure results: before making the first incorrect orientation (before\_error), continued with the first incorrect orientation (with\_error), and continued with the first incorrect orientation fixed (fixed\_error).}
\label{table:lrt-downstream-error}
\end{table}

To study the downstream effect of an incorrect orientation by the likelihood ratio test in Algorithm~\ref{alg:OrientEdges}, we compare the performance of the orientation procedure with early errors with that of when such errors are corrected. When applying the orientation procedure, the algorithm takes note of the first incorrect orientation made (on Line~\ref{lst:line:orientation-procedure-lrt}) during the procedure. Simultaneously in a separate process, the algorithm corrects the edge direction and proceeds with the remaining steps (Line~\ref{lst:line:orientation-procedure-post-lrt} and onward). There are no interventions thereafter. We then compare the learning accuracies of the graph before making the first incorrect orientation, the output graph of the procedure without intervention (original procedure), and the output graph of the procedure with the first error corrected. This comparison quantifies the negative effect of the first incorrect edge orientation. These graphs are labeled as \textit{before\_error}, \textit{with\_error}, and \textit{fixed\_error} in Table~\ref{table:lrt-downstream-error}. We used the same simulation specifications as Section~\ref{sec:experiments-compare-algs} and tested the orientation procedure on $N=300$ datasets per setting.

The downstream effect of the error is well-controlled, as evidenced by the small differences in the true positive (TP) and wrong orientation (Wrong Dir) counts in the table. If no downstream effects arose from the error, then the graph with the error fixed should have $\Delta \text{TP}=1$ more true positive and $\Delta \text{Wrong Dir}=1$ less incorrect orientation than the graph containing the error. We see that only one instance, network magic2 with invertible SEMs, has a difference of more than 2 true positives. 
The number of edges oriented by the procedure is slightly reduced in the presence of an error, as evidenced by a higher count of undirected edges (Undir). 
It is possible that some orientation rules on Lines~\ref{lst:line:orientation-procedure-orient-shared-nbr} and~\ref{lst:line:meeksrule} become inapplicable as a consequence of an earlier incorrect orientation. 
Nevertheless, the increase in undirected edges is less than 3 edges for the invertible setting and less than 2 edges under the non-invertible setting.
Observing $\Delta \text{TP} \leq 2$ mostly and $\Delta \text{Wrong Dir} \leq 2$ for all cases, we can conclude that the downstream effect is minimal.

This reflects the key feature in our algorithm design to evaluate pairs of nodes sharing no neighbors first, as described on on Lines~\ref{lst:line:calculate-nbr} --\ref{lst:line:filterbynbr} of Algorithm~\ref{alg:OrientEdges}, which minimizes the propagation of an error, should there be one. For instance, suppose the algorithm is applied to learn a DAG from its equivalence class, shown in Figure~\ref{fig:early-error-example} (a) and (b). The orientation procedure would first evaluate the edge $X_{1} - X_{2}$, per the ranking procedure, and apply the likelihood ratio test. If the edge is oriented as $X_{1} \rightarrow X_{2}$, then the procedure will apply Meek's rules to orient $X_{2} \rightarrow X_{3}$ and $X_{2} \rightarrow X_{4}$, depicted in (c). If the edge were incorrectly oriented as $X_{2} \rightarrow X_{1}$ as in (d), neither Meek's rules nor the orientation on Line~\ref{lst:line:orientation-procedure-orient-shared-nbr} of Algorithm~\ref{alg:OrientEdges} would apply since $X_{1}$ and $X_{2}$ share no neighbors. By this design, an error arising from an incorrect orientation does not easily propagate in the graph.

\begin{figure}
\centering
\begin{tikzpicture}[node distance={14mm}, thick, main/.style = {draw, circle}] 
\node(temp1) { }; 
\node[main] (X1) [below left of =temp1]{$X_{1}$}; 
\node[main] (X2) [below right of=temp1] {$X_{2}$};
\node[main] (X3) [below of=X1] {$X_{3}$};
\node[main] (X4) [below of=X2] {$X_{4}$};
\node[below of=temp1, node distance = 37mm]{(a) DAG $\calG$};
\draw[->, color=red] (X1) -- (X2);
\draw[->] (X2) -- (X3); 
\draw[->] (X2) -- (X4); 
\draw[->] (X3) -- (X4); 
\end{tikzpicture} 
\hspace{0.01cm}
\begin{tikzpicture}[node distance={14mm}, thick, main/.style = {draw, circle}] 
\node(temp1) { }; 
\node[main] (X1) [below left of =temp1]{$X_{1}$}; 
\node[main] (X2) [below right of=temp1] {$X_{2}$};
\node[main] (X3) [below of=X1] {$X_{3}$};
\node[main] (X4) [below of=X2] {$X_{4}$};
\node[below of=temp1, node distance = 37mm]{(b) CPDAG of $\calG$};
\draw[color=red] (X1) -- (X2);
\draw (X2) -- (X3); 
\draw (X2) -- (X4); 
\draw (X3) -- (X4); 
\end{tikzpicture} 
\hspace{0.01cm}
\begin{tikzpicture}[node distance={14mm}, thick, main/.style = {draw, circle}] 
\node(temp1) { }; 
\node[main] (X1) [below left of =temp1]{$X_{1}$}; 
\node[main] (X2) [below right of=temp1] {$X_{2}$};
\node[main] (X3) [below of=X1] {$X_{3}$};
\node[main] (X4) [below of=X2] {$X_{4}$};
\node[below of=temp1, node distance = 37mm]{(c) With correct orient.};
\draw[->, color=red] (X1) -- (X2);
\draw[->] (X2) -- (X3); 
\draw[->] (X2) -- (X4); 
\draw (X3) -- (X4); 
\end{tikzpicture} 
\hspace{0.01cm}
\begin{tikzpicture}[node distance={14mm}, thick, main/.style = {draw, circle}] 
\node(temp1) { }; 
\node[main] (X1) [below left of =temp1]{$X_{1}$}; 
\node[main] (X2) [below right of=temp1] {$X_{2}$};
\node[main] (X3) [below of=X1] {$X_{3}$};
\node[main] (X4) [below of=X2] {$X_{4}$};
\node[below of=temp1, node distance = 37mm]{(d) With wrong orient.};
\draw[->, color=red] (X2) -- (X1);
\draw (X2) -- (X3); 
\draw (X2) -- (X4); 
\draw (X3) -- (X4); 
\end{tikzpicture} 
\caption{The design of Algorithm~\ref{alg:OrientEdges} can minimize further errors from an incorrect orientation. Under the correct orientation $X_1\to X_2$, orientation rules are correctly applied and result in (c). On the converse shown in (d), the rules are not applicable and thus do not create any further errors.}
\label{fig:early-error-example}
\end{figure}

\subsection{Sensitivity to the Misranking of Edges}\label{sec:appendix-edge-misranking-exp}

Given that the learned CPDAG is usually not perfect, it is difficult to supply a correct ranking of the undirected edges to the algorithm. Instead, we compare the results of orienting edges by our ranking as in SNOE-SS and SNOE-CV to those of orienting edges in an arbitrary order (SNOE-SS-Rand, SNOE-CV-Rand). We apply the approaches to data sets generated under linear, invertible(inv), and non-invertible(ninv) functions with Gaussian noise. The simulated data sets use the exact settings described in Section 6.3. Then, we examine F1 scores after the orientation stage of two select networks (mehra and magic2) in Figure~\ref{fig:all-alg-gaussian-f1} under the different functions. Table~\ref{tab:misranking-exp-results} shows the F1 score after orienting edges in a random or ranked order, i.e. after stage 2 in Figure~\ref{fig:intermediate-f1-score}.

\begin{table}[h]
\centering
\begin{tabular}{rllrrr}
  \hline
network & approach & F1\_linear & F1\_inv & F1\_ninv \\ 
  \hline
mehra & SNOE-CV & 0.516 & 0.551 & 0.504 \\ 
mehra & SNOE-CV-Rand & 0.514 & 0.536 & 0.504 \\ 
mehra & SNOE-SS & 0.524 & 0.554 & 0.510 \\ 
mehra & SNOE-SS-Rand & 0.512 & 0.533 & 0.499 \\ 
magic2 & SNOE-CV & 0.761 & 0.631 & 0.639 \\ 
magic2 & SNOE-CV-Rand & 0.744 & 0.628 & 0.620 \\ 
magic2 & SNOE-SS & 0.766 & 0.632 & 0.631 \\ 
magic2 & SNOE-SS-Rand & 0.753 & 0.618 & 0.612 \\ 
   \hline
\end{tabular}
\caption{F1 scores of the graphs after evaluating edges in a ranked or random manner in the orientation stage.}
\label{tab:misranking-exp-results}
\end{table}

All approaches are applied to the same initial graph. The edge ranking procedure indeed led to more true positives in the orientation stage, as evidenced by the higher F1 scores. One may notice that the difference between the ranked and random procedures is not large. Under a random order, a greater proportion of undirected edges not satisfying the PANM criterion, due to missing parent variables, are evaluated by the likelihood test. As the algorithm cannot accurately estimate the true regression function due to missing parent variables, the likelihood test would likely find models for $X \rightarrow Y$ and $Y \rightarrow X$ equivalent, and thus keep the edge undirected. These edges will be correctly oriented later, once they meet the PANM condition following the orientation of other edges incident on the two nodes. In summary, the ranking procedure offers an improvement over a random ordering of edges. In the case where edges are misranked, the likelihood test prevents incorrect orientation and mitigates the impact of misranking.

\vskip 0.2in

\newpage
\bibliography{sample}

@article{cam,
author = {Peter B{\"u}hlmann and Jonas Peters and Jan Ernest},
title = {{CAM: Causal additive models, high-dimensional order search and penalized regression}},
volume = {42},
journal = {The Annals of Statistics},
number = {6},
publisher = {Institute of Mathematical Statistics},
pages = {2526 -- 2556},
year = {2014},
doi = {10.1214/14-AOS1260}
}

@article{resit_anm,
  author  = {Jonas Peters and Joris M. Mooij and Dominik Janzing and Bernhard Sch{{\"o}}lkopf},
  title   = {Causal Discovery with Continuous Additive Noise Models},
  journal = {Journal of Machine Learning Research},
  year    = {2014},
  volume  = {15},
  number  = {58},
  pages   = {2009--2053}
}

@article{hoyer2008nonlinear,
  title={Nonlinear causal discovery with additive noise models},
  author={Hoyer, Patrik and Janzing, Dominik and Mooij, Joris M and Peters, Jonas and Sch{\"o}lkopf, Bernhard},
  journal={Advances in neural information processing systems},
  volume={21},
  year={2008}
}

@article{vuong_test,
 ISSN = {00129682, 14680262},
 author = {Quang H. Vuong},
 journal = {Econometrica},
 number = {2},
 pages = {307--333},
 publisher = {[Wiley, Econometric Society]},
 title = {Likelihood Ratio Tests for Model Selection and Non-Nested Hypotheses},
 volume = {57},
 year = {1989}
}

@book{causation_search_prediction,
  author = {Spirtes, P. and Glymour, C. and Scheines, R.},
  edition = {2nd},
  publisher = {MIT press},
  review = {PC algorithm},
  title = {Causation, Prediction, and Search},
  year = 2000
}

@book{pearl@2000,
author = {Pearl, Judea},
title = {Causality: models, reasoning, and inference},
year = {2000},
isbn = {0521773628},
publisher = {Cambridge University Press},
address = {USA}
}

@article{shimizu2006linear,
  title={A linear non-Gaussian acyclic model for causal discovery.},
  author={Shimizu, Shohei and Hoyer, Patrik O and Hyv{\"a}rinen, Aapo and Kerminen, Antti and Jordan, Michael},
  journal={Journal of Machine Learning Research},
  volume={7},
  number={10},
  year={2006}
}

@article{wang_zhou_nonlinear,
  title={Causal network learning with non-invertible functional relationships},
  author={Wang, Bingling and Zhou, Qing},
  journal={Computational Statistics \& Data Analysis},
  volume={156},
  pages={107141},
  year={2021},
  publisher={Elsevier}
}

@inproceedings{meek1995causal_background_knowledge,
author = {Meek, Christopher},
title = {Causal inference and causal explanation with background knowledge},
year = {1995},
isbn = {1558603859},
publisher = {Morgan Kaufmann Publishers Inc.},
address = {San Francisco, CA, USA},
booktitle = {Proceedings of the Eleventh Conference on Uncertainty in Artificial Intelligence},
pages = {403–410},
numpages = {8},
location = {Montr\'{e}al, Qu\'{e}, Canada},
series = {UAI'95}
}

@inproceedings{cd_benchmarks_neurips2021,
 author = {Reisach, Alexander and Seiler, Christof and Weichwald, Sebastian},
 booktitle = {Advances in Neural Information Processing Systems},
 editor = {M. Ranzato and A. Beygelzimer and Y. Dauphin and P.S. Liang and J. Wortman Vaughan},
 pages = {27772--27784},
 publisher = {Curran Associates, Inc.},
 title = {Beware of the Simulated {DAG}! Causal Discovery Benchmarks May Be Easy to Game},
 volume = {34},
 year = {2021}
}

@article{pc_stable,
  title={Order-independent constraint-based causal structure learning.},
  author={Colombo, Diego and Maathuis, Marloes H and others},
  journal={J. Mach. Learn. Res.},
  volume={15},
  number={1},
  pages={3741--3782},
  year={2014}
}

@inproceedings{kci_test,
author = {Zhang, Kun and Peters, Jonas and Janzing, Dominik and Sch\"{o}lkopf, Bernhard},
title = {Kernel-based conditional independence test and application in causal discovery},
year = {2011},
isbn = {9780974903972},
publisher = {AUAI Press},
address = {Arlington, Virginia, USA},
booktitle = {Proceedings of the Twenty-Seventh Conference on Uncertainty in Artificial Intelligence},
pages = {804–813},
numpages = {10},
location = {Barcelona, Spain},
series = {UAI'11}
}

@article{chickering2002learning,
  title={Learning equivalence classes of Bayesian-network structures},
  author={Chickering, David Maxwell},
  journal={The Journal of Machine Learning Research},
  volume={2},
  pages={445--498},
  year={2002},
  publisher={JMLR. org}
}

@article{shah2020hardness,
   title={The hardness of conditional independence testing and the generalised covariance measure},
   volume={48},
   ISSN={0090-5364},
   DOI={10.1214/19-aos1857},
   number={3},
   journal={The Annals of Statistics},
   publisher={Institute of Mathematical Statistics},
   author={Shah, Rajen D. and Peters, Jonas},
   year={2020},
   month=jun }

@article{strobl2019kci_rcot,
  title={Approximate kernel-based conditional independence tests for fast non-parametric causal discovery},
  author={Strobl, Eric V and Zhang, Kun and Visweswaran, Shyam},
  journal={Journal of Causal Inference},
  volume={7},
  number={1},
  pages={20180017},
  year={2019},
  publisher={De Gruyter}
}

@article{flow_cytometry,
author = {Karen Sachs  and Omar Perez  and Dana Pe'er  and Douglas A. Lauffenburger  and Garry P. Nolan },
title = {Causal Protein-Signaling Networks Derived from Multiparameter Single-Cell Data},
journal = {Science},
volume = {308},
number = {5721},
pages = {523-529},
year = {2005},
doi = {10.1126/science.1105809},
eprint = {https://www.science.org/doi/pdf/10.1126/science.1105809}}

@inproceedings{Dor1992ASA,
  title={A simple algorithm to construct a consistent extension of a partially oriented graph},
  author={Dor, Dorit and Tarsi, Michael},
  booktitle={Technicial Report R-185, Cognitive Systems Laboratory, UCLA},
  pages={45},
  year={1992},
  publisher={Citeseer}
}

@article{notears,
  title={{DAG}s with no tears: Continuous optimization for structure learning},
  author={Zheng, Xun and Aragam, Bryon and Ravikumar, Pradeep K and Xing, Eric P},
  journal={Advances in neural information processing systems},
  volume={31},
  year={2018}
}

@article{mgcv_package,
  title={Package ‘mgcv’},
  author={Wood, Simon and Wood, Maintainer Simon},
  journal={R package version},
  volume={1},
  number={29},
  pages={729},
  year={2015}
}

@inproceedings{rolland2022score,
  title={Score matching enables causal discovery of nonlinear additive noise models},
  author={Rolland, Paul and Cevher, Volkan and Kleindessner, Matth{\"a}us and Russell, Chris and Janzing, Dominik and Sch{\"o}lkopf, Bernhard and Locatello, Francesco},
  booktitle={International Conference on Machine Learning},
  pages={18741--18753},
  year={2022},
  organization={PMLR}
}

@article{bello2023dagmalearningdagsmmatrices,
  title={{DAGMA}: Learning {DAG}s via m-matrices and a log-determinant acyclicity characterization},
  author={Bello, Kevin and Aragam, Bryon and Ravikumar, Pradeep},
  journal={Advances in Neural Information Processing Systems},
  volume={35},
  pages={8226--8239},
  year={2022}
}

@InProceedings{pmlr-v161-wienobst21a-extendability,
  title = 	 {Extendability of causal graphical models: Algorithms and computational complexity},
  author =       {Wien\"{o}bst, Marcel and Bannach, Max and Li\'{s}kiewicz, Maciej},
  booktitle = 	 {Proceedings of the Thirty-Seventh Conference on Uncertainty in Artificial Intelligence},
  pages = 	 {1248--1257},
  year = 	 {2021},
  editor = 	 {de Campos, Cassio and Maathuis, Marloes H.},
  volume = 	 {161},
  series = 	 {Proceedings of Machine Learning Research},
  month = 	 {27--30 Jul},
  publisher =    {PMLR}
}

@article{heinze2018invariant,
  title={Invariant causal prediction for nonlinear models},
  author={Heinze-Deml, Christina and Peters, Jonas and Meinshausen, Nicolai},
  journal={Journal of Causal Inference},
  volume={6},
  number={2},
  pages={20170016},
  year={2018},
  publisher={De Gruyter}
}

@inproceedings{rosenfeld2021risks,
  title={The Risks of Invariant Risk Minimization},
  author={Rosenfeld, Elan and Ravikumar, Pradeep and Risteski, Andrej},
  booktitle={International Conference on Learning Representations},
  volume={9},
  year={2021}
}

@inproceedings{yu2019dag,
  title={{DAG-GNN}: {DAG} structure learning with graph neural networks},
  author={Yu, Yue and Chen, Jie and Gao, Tian and Yu, Mo},
  booktitle={International conference on machine learning},
  pages={7154--7163},
  year={2019},
  organization={PMLR}
}

@article{scutari2010learning,
  title={Learning Bayesian networks with the bnlearn R package},
  author={Scutari, Marco},
  journal={Journal of statistical software},
  volume={35},
  pages={1--22},
  year={2010}
}

@article{glymour2019review,
  title={Review of causal discovery methods based on graphical models},
  author={Glymour, Clark and Zhang, Kun and Spirtes, Peter},
  journal={Frontiers in genetics},
  volume={10},
  pages={524},
  year={2019},
  publisher={Frontiers Media SA}
}

@article{vowels2022d,
  title={D’ya like {DAG}s? a survey on structure learning and causal discovery},
  author={Vowels, Matthew J and Camgoz, Necati Cihan and Bowden, Richard},
  journal={ACM Computing Surveys},
  volume={55},
  number={4},
  pages={1--36},
  year={2022},
  publisher={ACM New York, NY}
}

@article{zhang2016nonlinear,
  title={Nonlinear functional causal models for distinguishing cause from effect},
  author={Zhang, Kun and Hyv{\"a}rinen, Aapo},
  journal={Statistics and causality: Methods for applied empirical research},
  pages={185--201},
  year={2016},
  publisher={Wiley Online Library}
}

@article{li2020nonparametric,
  title={On nonparametric conditional independence tests for continuous variables},
  author={Li, Chun and Fan, Xiaodan},
  journal={Wiley Interdisciplinary Reviews: Computational Statistics},
  volume={12},
  number={3},
  pages={e1489},
  year={2020},
  publisher={Wiley Online Library}
}

@inproceedings{zheng2020learning,
  title={Learning sparse nonparametric {DAG}s},
  author={Zheng, Xun and Dan, Chen and Aragam, Bryon and Ravikumar, Pradeep and Xing, Eric},
  booktitle={International Conference on Artificial Intelligence and Statistics},
  pages={3414--3425},
  year={2020},
  organization={Pmlr}
}

@article{andersson1997characterization,
  title={A characterization of Markov equivalence classes for acyclic digraphs},
  author={Andersson, Steen A and Madigan, David and Perlman, Michael D},
  journal={The Annals of Statistics},
  volume={25},
  number={2},
  pages={505--541},
  year={1997},
  publisher={Institute of Mathematical Statistics}
}

@inproceedings{verma1990,
author = {Verma, Thomas and Pearl, Judea},
title = {Equivalence and synthesis of causal models},
year = {1990},
isbn = {0444892648},
publisher = {Elsevier Science Inc.},
address = {USA},
booktitle = {Proceedings of the Sixth Annual Conference on Uncertainty in Artificial Intelligence},
pages = {255–270},
numpages = {16},
series = {UAI '90}
}

@article{spirtes1991algorithm,
  title={An algorithm for fast recovery of sparse causal graphs},
  author={Spirtes, Peter and Glymour, Clark},
  journal={Social science computer review},
  volume={9},
  number={1},
  pages={62--72},
  year={1991},
  publisher={Sage Publications Sage CA: Thousand Oaks, CA}
}

@article{hasan2023survey,
  title={A survey on causal discovery methods for iid and time series data},
  author={Hasan, Uzma and Hossain, Emam and Gani, Md Osman},
  journal={arXiv preprint arXiv:2303.15027},
  year={2023}
}

@inproceedings{peters2011IFMOC,
author = {Peters, Jonas and Mooij, Joris M. and Janzing, Dominik and Sch\"{o}lkopf, Bernhard},
title = {Identifiability of causal graphs using functional Models},
year = {2011},
isbn = {9780974903972},
publisher = {AUAI Press},
address = {Arlington, Virginia, USA},
booktitle = {Proceedings of the Twenty-Seventh Conference on Uncertainty in Artificial Intelligence},
pages = {589–598},
numpages = {10},
location = {Barcelona, Spain},
series = {UAI'11}
}

@article{gretton2009nonlinear,
  title={Nonlinear directed acyclic structure learning with weakly additive noise models},
  author={Gretton, Arthur and Spirtes, Peter and Tillman, Robert},
  journal={Advances in neural information processing systems},
  volume={22},
  year={2009}
}

@inproceedings{khemakhem2021causal,
  title={Causal autoregressive flows},
  author={Khemakhem, Ilyes and Monti, Ricardo and Leech, Robert and Hyvarinen, Aapo},
  booktitle={International conference on artificial intelligence and statistics},
  pages={3520--3528},
  year={2021},
  organization={PMLR}
}

@inproceedings{zhang2009pnl_identifiability,
author = {Zhang, Kun and Hyv\"{a}rinen, Aapo},
title = {On the identifiability of the post-nonlinear causal model},
year = {2009},
isbn = {9780974903958},
publisher = {AUAI Press},
booktitle = {Proceedings of the Twenty-Fifth Conference on Uncertainty in Artificial Intelligence},
pages = {647–655},
numpages = {9},
series = {UAI '09}
}

@inproceedings{uemura2020estimation_pnl,
  title={Estimation of post-nonlinear causal models using autoencoding structure},
  author={Uemura, Kento and Shimizu, Shohei},
  booktitle={ICASSP 2020-2020 IEEE International Conference on Acoustics, Speech and Signal Processing (ICASSP)},
  pages={3312--3316},
  year={2020},
  organization={IEEE}
}

@article{mooij2016distinguishing,
  title={Distinguishing cause from effect using observational data: methods and benchmarks},
  author={Mooij, Joris M and Peters, Jonas and Janzing, Dominik and Zscheischler, Jakob and Sch{\"o}lkopf, Bernhard},
  journal={Journal of Machine Learning Research},
  volume={17},
  number={32},
  pages={1--102},
  year={2016}
}

@inproceedings{monti2020causal,
  title={Causal discovery with general non-linear relationships using non-linear ICA},
  author={Monti, Ricardo Pio and Zhang, Kun and Hyv{\"a}rinen, Aapo},
  booktitle={Uncertainty in artificial intelligence},
  pages={186--195},
  year={2020},
  organization={PMLR}
}

@inproceedings{huang2018generalized,
  title={Generalized score functions for causal discovery},
  author={Huang, Biwei and Zhang, Kun and Lin, Yizhu and Sch{\"o}lkopf, Bernhard and Glymour, Clark},
  booktitle={Proceedings of the 24th ACM SIGKDD international conference on knowledge discovery \& data mining},
  pages={1551--1560},
  year={2018}
}

@article{Stone1985,
  author = {Stone, Charles J.},
  title = {Additive regression and other nonparametric models},
  journal = {Annals of Statistics},
  year = {1985},
  volume = {13},
  pages = {689--705}
}

@article{ramsey2025scalable,
  title={Scalable Causal Discovery from Recursive Nonlinear Data via Truncated Basis Function Scores and Tests},
  author={Ramsey, Joseph and Andrews, Bryan and Spirtes, Peter},
  journal={arXiv preprint arXiv:2510.04276},
  year={2025}
}

\end{document}